\documentclass{article} 
\usepackage{iclr2026_conference,times}

\usepackage{amsmath,amsfonts,bm}

\def\eqref#1{equation~\ref{#1}}

\def\1{\bm{1}}

\DeclareMathAlphabet{\mathsfit}{\encodingdefault}{\sfdefault}{m}{sl}
\SetMathAlphabet{\mathsfit}{bold}{\encodingdefault}{\sfdefault}{bx}{n}

\usepackage{hyperref}
\usepackage{url}
\usepackage{xurl}
\usepackage{microtype}
\usepackage{bm}
\usepackage{amsmath, amssymb, amsfonts}
   
\usepackage{needspace}
\usepackage{booktabs}
\usepackage{tabularx}
\usepackage{fvextra}
\usepackage{placeins}
\usepackage{array}
\usepackage{subcaption}
\usepackage{rotating}
\usepackage{multirow}
\usepackage{algorithm}
\usepackage{algpseudocode}
\usepackage{wrapfig}
\usepackage{times}
\usepackage{latexsym}
\usepackage{makecell}
\usepackage{tabularray}
\usepackage{listings}
\usepackage{upquote}
\usepackage{amsfonts}
\usepackage{colortbl} 
\usepackage{graphicx}
\usepackage{pifont}
\newcommand{\tfyes}{\ding{51}}
\newcommand{\tfno}{\ding{55}}

\newcolumntype{Y}{>{\centering\arraybackslash}X}

\definecolor{chrxcolor}{RGB}{1, 140, 116}

\definecolor{bestbg}{rgb}{0.78,0.90,0.79}
\definecolor{secondbg}{rgb}{0.87,0.94,1.00}
\definecolor{directrow}{RGB}{228,225,244}

\definecolor{gazerow}{RGB}{243,212,219}
\definecolor{overviewrow}{RGB}{232,242,250}
\definecolor{relativerow}{RGB}{245,228,196}
\newcommand{\bestcell}[1]{%
  \cellcolor{bestbg}\textbf{#1}%
}

\newcommand{\secondcell}[1]{%
  \cellcolor{secondbg}#1%
}

\title{The Earth in One Gaze: Training-Free Active Focus for UHR Remote Sensing Understanding}

\newcommand{\authorsep}{\hspace{1.1em}}
\author{%
    \textbf{Yao Zhang}$^{1,*}$\authorsep
    \textbf{Pengyu Dai}$^{2,3,*}$\authorsep
    \textbf{Wei Guo}$^{1,\dagger}$\authorsep
    \textbf{Jian Liang}$^{1}$\\[0.25em]
    \textbf{Jian Song}$^{3}$\authorsep
    \textbf{Yafei Ou}$^{3}$\authorsep
    \textbf{Hongruixuan Chen}$^{3,\dagger}$\authorsep
    \textbf{Naoto Yokoya}$^{2,3}$\\[0.7em]
    $^{1}$Wuhan University\authorsep
    $^{2}$The University of Tokyo\authorsep
    $^{3}$RIKEN AIP\\[0.35em]
    {\small\texttt{\{zhangyyy.ai, guowei-lmars\}@whu.edu.cn}\authorsep
    \texttt{pengyu.dai@riken.jp}\authorsep
    \texttt{qschrx@gmail.com}}%
}

\iclrfinalcopy 
\makeatletter
\newcommand{\wrapfigureplaceholder}{%
  \par
  \ifvoid\WF@box\else
    \begin{figure}[htbp]
      \hfill\box\WF@box
    \end{figure}%
  \fi
  \WFclear
}
\makeatother

\begin{document}
\raggedbottom
\setlength{\textfloatsep}{14pt plus 4pt minus 4pt}
\setlength{\floatsep}{12pt plus 2pt minus 2pt}
\setlength{\intextsep}{12pt plus 2pt minus 2pt}
\setlength{\abovecaptionskip}{6pt}

\maketitle
\begingroup
\renewcommand{\thefootnote}{\fnsymbol{footnote}}
\footnotetext[1]{Equal contribution.}
\footnotetext[2]{Corresponding authors.}
\footnotetext[3]{\href{https://github.com/zhangyyy-ai-rs/GazeEarth}{\textcolor{magenta}{https://github.com/zhangyyy-ai-rs/GazeEarth}}}
\endgroup

\begin{figure*}[htbp]
  \centerline{\includegraphics[scale=0.44]{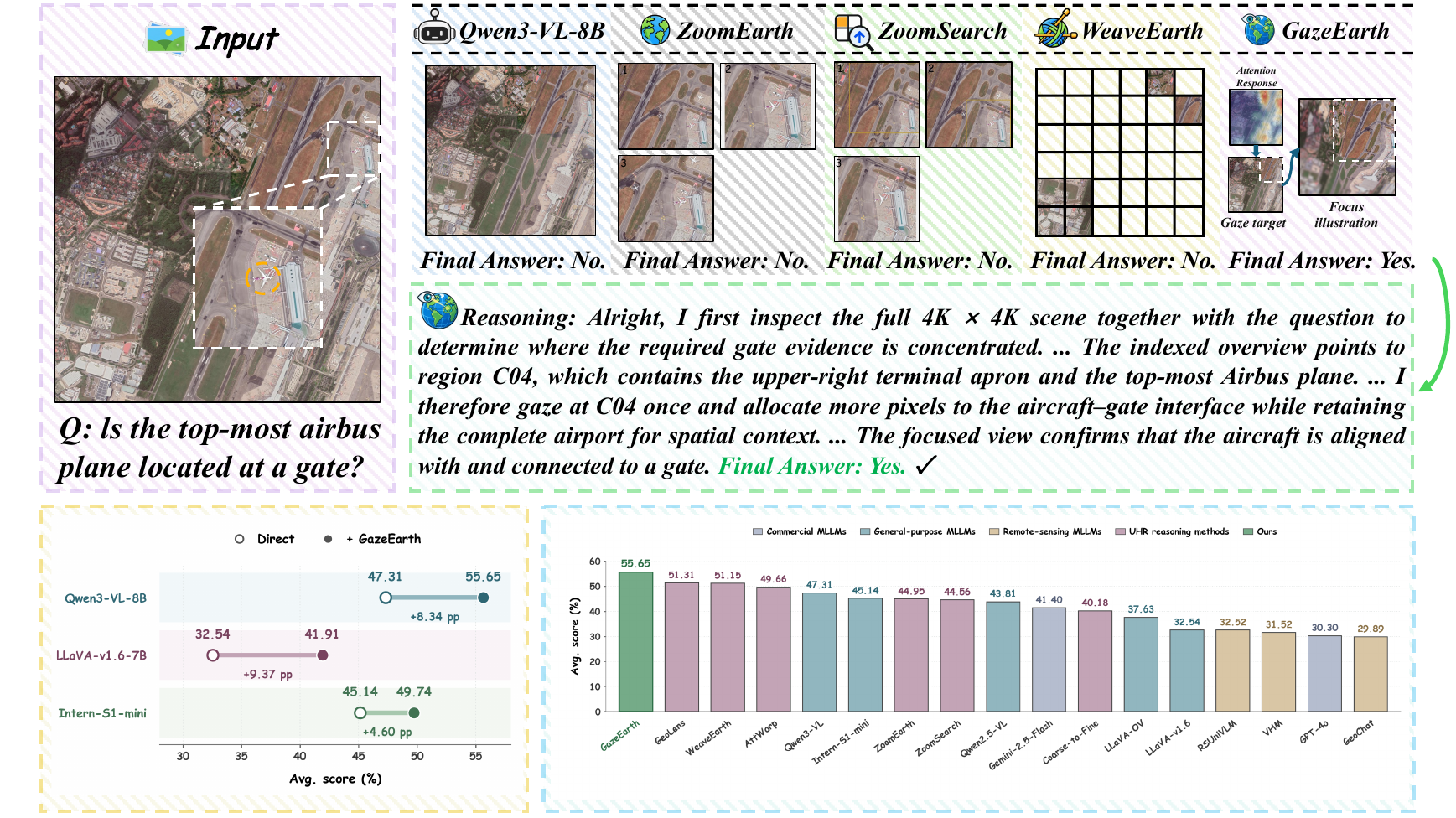}}
  \vspace{-0.05cm}
    \caption{
    Overview of GazeEarth.
    (i) Different observation strategies can lead to different answers.
    (ii) GazeEarth combines question-guided region selection with foveated observation in a continuous full-scene view.
    (iii) The same strategy benefits different frozen backbones.
    (iv) Mean-score comparison with other methods on three remote-sensing benchmarks.
    }
  \label{fig-teaser}
\end{figure*}

\begin{abstract}
Multimodal large language models (MLLMs) must balance local detail against scene context when interpreting ultra-high-resolution (UHR) remote sensing (RS) imagery within a limited visual-input budget.
Existing selection-based methods either prune tokens and
select patches through relevance scoring, or crop actively
through repeated inspection.
Neither strategy directly redistributes pixels within
a continuous full-scene view: the first retains selected
tokens or patches, and the second re-encodes a crop
detached from its surroundings.
Our pilot study finds that a frozen MLLM already produces
useful question-guided spatial requests, yet crop-based
inspection of the selected regions does not consistently
improve its answers.
We therefore formulate UHR understanding as a question of where to spend a fixed pixel budget.
Based on this, we introduce GazeEarth, a simple-yet-effective training-free framework that couples question-guided region selection with full-scene foveated observation.
The MLLM selects evidence cells from an indexed overview;
a deterministic, topology-preserving warp resamples the original
image onto a fixed-size canvas, enlarging their shared neighborhood
while compressing the periphery; the same frozen model answers
from this focused view, using at most two MLLM calls
and no external selector or iterative search.
Across three UHR remote sensing benchmarks and four frozen backbones, GazeEarth improves benchmark-averaged accuracy by 4.6 to 9.4 percentage points over direct answering and 3.4 to 4.3 over overview answering, outperforming task-trained methods.
Our analyses show that existing MLLMs can guide where
to look in UHR images on their own, and that what they
can infer from the selected evidence depends on how
that evidence is presented.

\end{abstract}

\section{Introduction}
\label{sec:introduction}

\par Understanding ultra-high-resolution (UHR) remote sensing imagery requires balancing local fine-grained details against the broader scene context~\citep{li2025describeearth,xin2026segearthr2}. For instance, determining whether a specific aircraft is parked at a terminal necessitates not only identifying the aircraft instance but also analyzing its spatial relationship to the surrounding airport infrastructure. However, a typical $10,000 \times 10,000$ scene contains roughly a hundred million pixels~\citep{wang2025xlrsbench, luo2025coarsetofine}, far exceeding the visual-token limits of current multimodal large language models (MLLMs). Standard uniform downsampling inevitably obscures critical small-scale objects~\citep{guo2025llavauhd,ding2025hilmd,li2026minigemini,liu2026adapatch}.

\par To circumvent this resolution bottleneck, recent methodologies have increasingly adopted active vision strategies. Methods such as targeted zooming~\citep{wu2024vstar,shen2025zoomeye,zheng2026deepeyes}, learned crop-and-zoom policies~\citep{liu2026zoomearth}, or patch retrieval~\citep{zhou2025zoomsearch} selectively allocate the limited pixel budget to specific regions of interest. Nevertheless, these approaches inherently suffer from contextual fragmentation: while extracting a high-resolution crop exposes local visual features, it detaches the target from its surrounding environment~\citep{zhang2025falcon,zhang2026beyondllavahd}. In remote sensing, semantic meaning is heavily contingent on spatial relationships~\citep{wang2024earthvqa}. Providing a global thumbnail alongside isolated crops
requires the model to implicitly reconstruct spatial
alignments, adding a cross-view association step
during reasoning.

\par This raises a critical question: in UHR image understanding, is the primary bottleneck the localization of evidence, or its effective spatial presentation? Our pilot study (Section~\ref{pilot}) reveals a counterintuitive insight. We demonstrate that off-the-shelf, frozen MLLMs can
generate spatial requests aligned with question-specified
regions, achieving higher alignment than dedicated
external retrievers such as SigLIP2~\citep{tschannen2025siglip2}. However, accurate localization does not guarantee successful task completion. When these question-selected regions are extracted as isolated crops and provided for reasoning, downstream performance remains inconsistent—even degrading by over 9 percentage points on MME-RealWorld-RS compared to the downsampled global overview. This indicates that identifying a relevant neighborhood is insufficient if inspecting it requires discarding its spatial surroundings.

\par Biological vision systems naturally resolve this via foveation: fixating on a region of interest to inspect fine-grained details
while maintaining a compressed peripheral representation
for global spatial awareness.~\citep{schwartz1977spatial,recasens2018learningtozoom,thavamani2021fovea,thavamani2023unzoom}. Inspired by this mechanism, we propose GazeEarth, a simple-yet-effective training-free framework that reframes UHR understanding as a continuous spatial reallocation of the pixel budget rather than discrete image cropping. In GazeEarth, a frozen MLLM first selects an evidence region from an indexed overview.
Subsequently, a deterministic, topology-preserving warp
resamples the original image directly onto a fixed-size canvas,
allocating more output pixels to the selected neighborhood
while compressing the periphery. This ensures the focal region remains physically connected to its broader geographical context.
The identical frozen model then performs reasoning over this single, foveated view. Without relying on iterative search algorithms, external selection modules, or parameter updates, this unified perception-action loop proves highly effective. Evaluated across three UHR remote sensing benchmarks
utilizing four distinct frozen backbones, GazeEarth
consistently improves over both direct and overview
answering, delivering benchmark-averaged gains of
4.6--9.4 and 3.4--4.3 percentage points, respectively.

\par Our primary contributions are summarized as follows:

\begin{enumerate}
    \setlength{\itemsep}{0.25em}
    \setlength{\parsep}{0pt}

    \item We distinguish coarse evidence localization from effective evidence presentation in UHR remote sensing, showing that a frozen MLLM provides useful spatial requests, but selected-region cropping does not consistently improve downstream answering.

    \item We introduce GazeEarth, which couples a frozen MLLM's explicit region selection with deterministic, topology-preserving focus within a fixed-size answer-stage canvas, using at most two MLLM calls and no external selector or iterative search.

    \item We demonstrate gains across three benchmarks and four frozen backbones, and characterize the roles of gaze source, evidence presentation, and selection depth through grounding, accuracy, and inference-cost analyses.
\end{enumerate}

\section{Motivation: Rethinking Evidence Presentation}
\label{pilot}

\par To identify the true bottleneck in ultra-high-resolution (UHR) image reasoning, we design two preliminary probes. Specifically, we investigate whether the primary challenge lies in the model's inability to locate the evidence or in the way the extracted evidence is presented for downstream reasoning.

\begin{table}[t]
\centering
\begin{minipage}[t]{0.50\textwidth}
\centering
\caption{\textbf{Who selects the region.}
Benchmark-averaged gaze grounding of different observers, together with the answer accuracy
of their aligned and off-target requests.}
\label{tab:pilot-grounding}
\scriptsize
\renewcommand{\arraystretch}{1.0}
\setlength{\tabcolsep}{1.5pt}
\setlength{\aboverulesep}{0.25ex}
\setlength{\belowrulesep}{0.25ex}
\begin{tabularx}{\linewidth}{@{}l
>{\hsize=0.85\hsize\linewidth=\hsize}Y
>{\hsize=0.85\hsize\linewidth=\hsize}Y
>{\hsize=1.05\hsize\linewidth=\hsize}Y
>{\hsize=1.25\hsize\linewidth=\hsize}Y@{}}
\toprule
\raisebox{1.0\baselineskip}[0pt][0pt]{\textbf{Observer}}
& \raisebox{0.5\baselineskip}[0pt][0pt]{\shortstack[c]{\textbf{Align.}\\\textbf{(\%)} $\uparrow$}}
& \raisebox{0.5\baselineskip}[0pt][0pt]{\shortstack[c]{\textbf{Grid}\\\textbf{Dist.} $\downarrow$}}
& \shortstack[c]{\textbf{Acc.@}\\\textbf{Aligned}\\\textbf{(\%)} $\uparrow$}
& \shortstack[c]{\textbf{Acc.@}\\\textbf{Off-target}\\\textbf{(\%)} $\uparrow$}
\\
\midrule
Random Focus & 16.35 & 1.879 & 44.91 & \bestcell{42.93} \\
Question-only MLLM & 42.82 & 1.450 & - & - \\
SigLIP2 & \secondcell{43.09} & \secondcell{1.233} & \secondcell{48.95} & 41.66 \\
RemoteCLIP & 37.20 & 1.343 & 48.80 & 41.47 \\
\midrule
Frozen MLLM & \bestcell{77.74} & \bestcell{0.334} & \bestcell{50.49} & \secondcell{42.00} \\
\bottomrule
\end{tabularx}
\end{minipage}\hfill
\begin{minipage}[t]{0.47\textwidth}
\centering
\caption{\textbf{How the selected evidence is presented.}
Answer accuracy (\%) of the frozen MLLM when the same
question-selected region is shown in different ways.}
\label{tab:pilot-presentation}
\scriptsize
\renewcommand{\arraystretch}{1.0}
\setlength{\tabcolsep}{1.5pt}
\setlength{\aboverulesep}{0.25ex}
\setlength{\belowrulesep}{0.25ex}
\begin{tabularx}{\linewidth}{@{}>{\raggedright\arraybackslash}p{0.33\linewidth}
>{\hsize=1.00\hsize\linewidth=\hsize}Y
>{\hsize=0.90\hsize\linewidth=\hsize}Y
>{\hsize=1.30\hsize\linewidth=\hsize}Y
>{\hsize=0.80\hsize\linewidth=\hsize}Y@{}}
\toprule
\raisebox{1.0\baselineskip}[0pt][0pt]{\textbf{Presentation}}
& \raisebox{0.5\baselineskip}[0pt][0pt]{\shortstack[c]{\textbf{XLRS-}\\\textbf{Bench}}}
& \raisebox{0.5\baselineskip}[0pt][0pt]{\shortstack[c]{\textbf{LRS-}\\\textbf{GRO}}}
& \shortstack[c]{\textbf{MME-}\\\textbf{Real-}\\\textbf{World-RS}}
& \raisebox{1.0\baselineskip}[0pt][0pt]{\textbf{Avg.}}
\\
\midrule
Overview & 45.13 & 59.99 & \bestcell{49.01} & \secondcell{51.38} \\
Gaze + Crop & \secondcell{50.55} & 58.93 & 39.81 & 49.76 \\
Think with image & \bestcell{51.43} & \bestcell{60.54} & \secondcell{42.24} & \bestcell{51.40} \\
\bottomrule
\end{tabularx}
\end{minipage}
\end{table}

\paragraph{Finding 1: the answerer can already guide where to look.}
\par We compare the MLLM's spatial requests with
question-derived reference cells and with requests from
external observers and a question-only MLLM.
As reported in Table~\ref{tab:pilot-grounding},
the image-conditioned MLLM achieves 77.74\%
benchmark-averaged alignment, compared with 42.82\%
for Question-only MLLM, 43.09\% for SigLIP2,
and 37.20\% for RemoteCLIP.
Its reported alignment exceeds question-only selection
on all three benchmarks
(Table~\ref{tab:gaze-grounding}).
Figure~\ref{fig:gaze-tolerance} further shows that
most off-target requests lie one grid step from
a question-derived reference cell.
These results characterize spatial-request alignment
with question-specified regions.
Further evaluation against dataset-provided target
boxes on 537 LRS-GRO questions without explicit
location cues yields a center-hit rate of 64.99\%,
compared with 52.33\% for SigLIP2 and 37.99\%
for RemoteCLIP
(Table~\ref{tab:independent-target-localization}).
The next probe examines whether presenting
the selected regions as crops improves answering.

\begin{wrapfigure}[14]{r}{0.42\textwidth}
    \vspace{-\baselineskip}
    \centering
    \setlength{\abovecaptionskip}{2pt}
    \includegraphics[width=\linewidth]
    {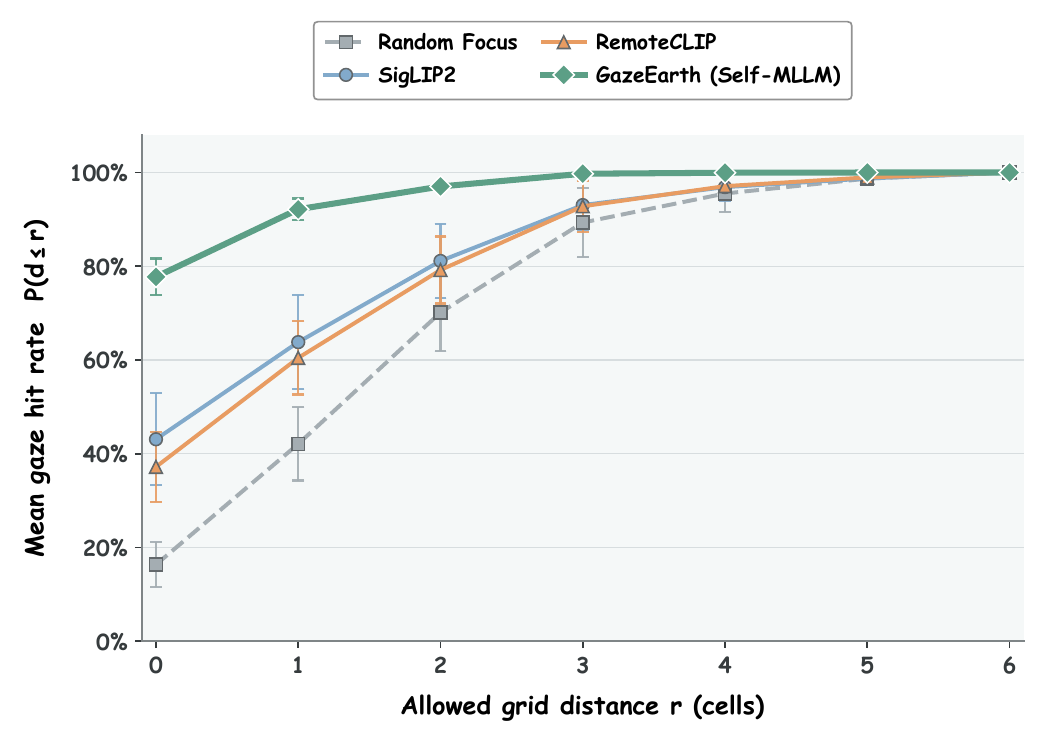}
    \caption{Gaze hit rates across three benchmarks
    (mean $\pm$ standard deviation).}
    \label{fig:gaze-tolerance-avg}
\end{wrapfigure}

\paragraph{Finding 2: showing the model the crop does not reliably help.}

\par If the model can already generate a useful spatial
request, does extracting and zooming into the selected
region solve the UHR reasoning problem? To answer this, we conduct an evidence-presentation probe comparing reasoning on the downsampled full-scene view (Overview) against reasoning on the isolated high-resolution crop of the
question-selected region (Gaze + Crop). As reported in Table~\ref{tab:pilot-presentation}, crop-based inspection severely degrades performance on contextually demanding tasks. For instance, on the MME-RealWorld-RS benchmark, feeding the MLLM the isolated high-resolution crop actually reduces the answering accuracy by 9.2 percentage points (from 49.01$\%$ to 39.81$\%$) compared to simply reasoning over the blurry global overview. 
Furthermore, even when a global thumbnail is provided
alongside the isolated crop (Think with Image),
accuracy on MME-RealWorld-RS remains below Overview,
failing to fully recover the performance drop.

\paragraph{Implication.}
Figure~\ref{fig:gaze-tolerance-avg} summarizes gaze
localization across the three benchmarks.
Localization, however, determines where to inspect,
not how the selected evidence is presented.
While cropping magnifies local details, isolating a
region can remove the surrounding landmarks needed
for spatial reasoning.
Accurate localization, therefore, needs an observation
that exposes the selected evidence in context.
The visual input should allocate more pixels to the
target region while preserving its continuous spatial
connection to the periphery.
This motivates the topology-preserving foveated
framework of GazeEarth.

\section{GazeEarth}
\label{sec:method}

\subsection{The Perception-Action Loop}
\par We formulate UHR image understanding as a single-step perception-action loop driven entirely by a frozen MLLM, denoted as $F_\theta(\cdot)$. Given a UHR image $I \in \mathbb{R}^{H \times W \times 3}$ and a user query $q$, our objective is to dynamically reallocate a fixed canvas pixel budget $B = h \times w$ (determined by the maximum answer-stage canvas size) without severing the global topology of the scene.

\par
\begin{figure*}[!t]
  \centerline{\includegraphics[scale=0.52]{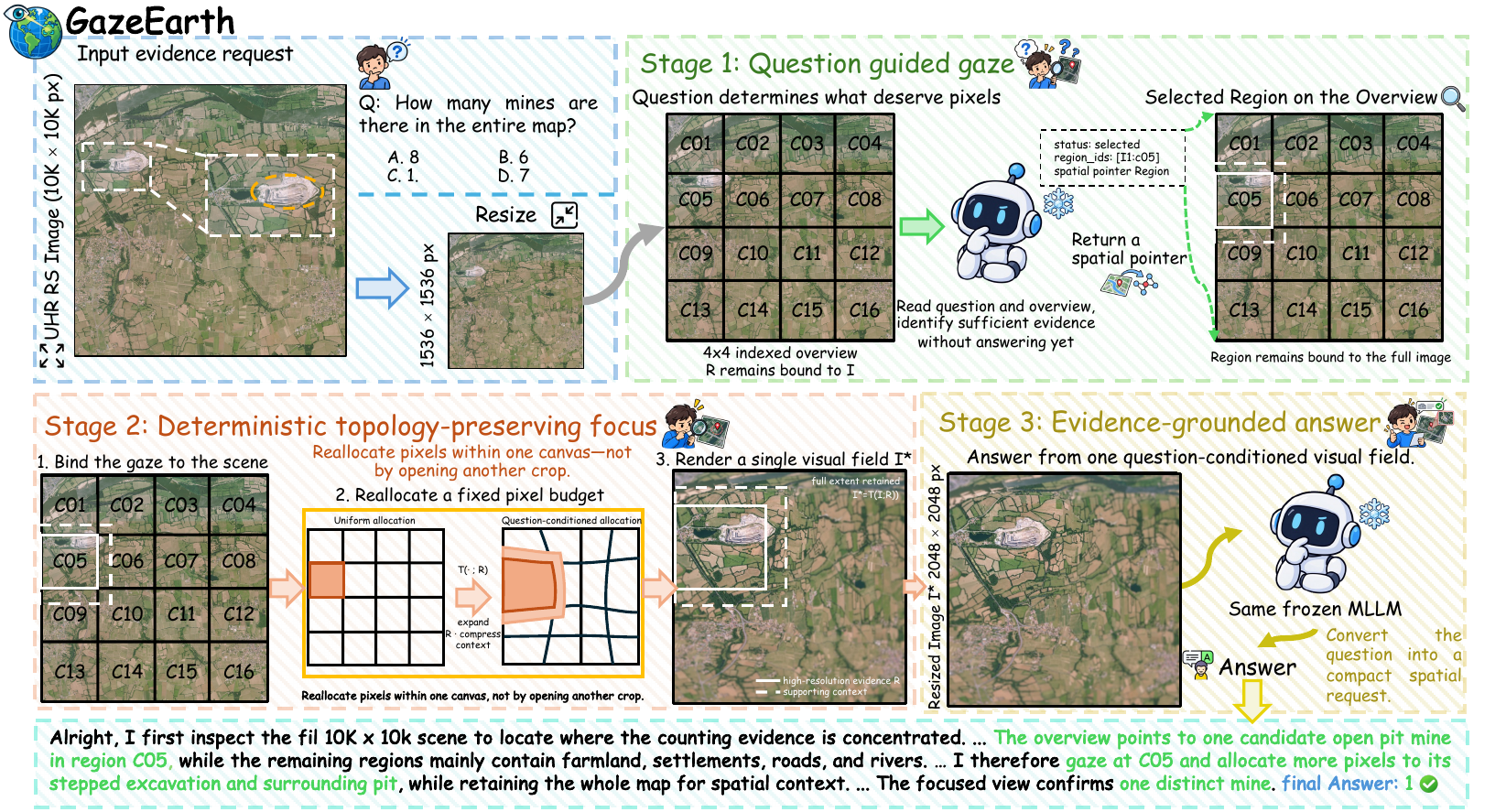}}
  \vspace{-0.1cm}
  \caption{Overview of GazeEarth. A frozen MLLM selects evidence regions from a question and an indexed overview. A deterministic, topology-preserving warp enlarges the selected neighborhood within a fixed-size, full-scene canvas. The same model then answers from this focused observation.}
      \label{fig-framework}
\end{figure*}

 As illustrated in Figure~\ref{fig-framework}, the GazeEarth framework execution unfolds in three consecutive phases: 
\begin{itemize}
    \item Question-Guided Gaze (Perception): The MLLM observes a severely downsampled overview $\tilde{I}_{idx}$ and the query $q$, acting as a zero-shot policy to output a spatial request $R$ indicating where the critical visual evidence is concentrated.
    \item Deterministic Foveation (Action): A non-parametric, topology-preserving operator $\mathcal{T}_B$ executes this spatial request, rendering a foveated observation $I^* = \mathcal{T}_B(I; R)$. This operator magnifies the requested region $R$ while compressing the global periphery, strictly adhering to the fixed budget $B$.  

    \item Evidence-Grounded Answering (Perception): The identical frozen MLLM interprets the foveated continuous canvas $I^*$ to generate the final semantic response $\hat{y}$.
\end{itemize}

\subsection{Question Guided Gaze}
\label{sec:gaze}

The pilot study shows that a frozen MLLM can already
produce useful question-guided spatial requests,
yet selecting a region alone does not ensure
successful answering.
We therefore elicit a spatial request before answer
generation and use it to guide the next observation.
An indexed overview provides the spatial references
through which the model specifies where to allocate
more of the answer stage canvas. We overlay an indexed $m\times n$ grid, with $m=n=4$,
on an overview whose long side is capped at 1536 pixels.
Each identifier maps to a source image region.
For zero-based row $r$ and column $c$, this region is
\begin{equation}
\mathcal C_{r,c}=
\left[\frac{cW}{n},\frac{(c+1)W}{n}\right)
\times
\left[\frac{rH}{m},\frac{(r+1)H}{m}\right),
\label{eq:gaze-cell}
\end{equation}
where $0\le r<m$ and $0\le c<n$, with coordinates
ordered as $(x,y)$ and raster boundaries rounded to pixels.
The prompt requests the smallest set of at most two cells
covering the evidence needed for $q$, without answering
the question.
This allows both regions in a comparison to be requested
within one selection round.
The valid identifiers are mapped to a set of grid
indices $\mathcal S$. $\operatorname{Parse}$ constructs the spatial request $R$
by padding the selected cells to include neighbouring
context and merging overlapping regions:
\begin{equation}
R=\operatorname{Merge}\!\left(
\left\{
\operatorname{Pad}_{\delta}(\mathcal C_{r,c})
\;\middle|\;
(r,c)\in\mathcal S
\right\}
\right).
\label{eq:gaze-request}
\end{equation}
Here, $\operatorname{Pad}_{\delta}$ extends each cell
by a fraction $\delta$ of its width and height on each
side and clips it to the source boundary;
$\operatorname{Merge}$ replaces overlapping groups
by their enclosing rectangles.
A single gaze denotes one selection round,
not one selected cell.
An empty request from \texttt{global\_only} or
\texttt{brief\_incomplete} retains the global view.
A successfully resolved box in $q$ directly defines
a contextual rectangle and bypasses model-guided selection.
Appendix~\ref{app:focus-details} gives the parsing
and coordinate rules.
The resulting request $R$ is passed to the focus operator,
which enlarges the requested neighborhood while retaining
the full scene.

\subsection{Topology-Preserving Foveated Mapping}
\label{focus}

Identifying a relevant region is only useful if the next
observation exposes its evidence in context.
We therefore translate $R$ into a continuous allocation
of the answer-stage canvas: the selected neighborhood
is enlarged while remaining spatially connected to
the full scene.

\paragraph{Piecewise-linear foveated mapping.}

We determine the output canvas dimensions $h\times w$ from
the source aspect ratio, with a 2048-pixel long-side cap. For nonempty $R$, the smallest enclosing rectangle defines
a shared focus neighborhood, including the space between
selected regions.
Its boundaries are projected onto the canvas and rounded
to pixels to determine the allocation, while the original
source coordinates are retained for resampling.
Along a canvas axis of length $L$, let $[a,b]$ denote the
projected focus interval, with $0\leq a<b\leq L$ and
$\ell=b-a$.
The enlargement factor $\alpha=1.5$ and focus-fraction
threshold $\beta=0.55$ determine the target span:
\begin{equation}
\ell^\star =
\min\!\left\{
\max\{\ell,\operatorname{round}(\alpha\ell)\},
\max\{\ell,\operatorname{round}(\beta L)\},
L-\epsilon_- -\epsilon_+
\right\}.
\label{eq:focus-span}
\end{equation}
Here, $\epsilon_-=1$ when $a>0$ and $\epsilon_+=1$ when
$b<L$, and each is zero otherwise.
These margins reserve one output pixel for each nonempty
peripheral interval.
We centre the target interval $[a^\star,b^\star]$,
where $b^\star=a^\star+\ell^\star$, on the projected interval
when feasible and shift it to respect these margins.
Appendix~\ref{app:focus-details} gives the exact placement rule.
The resulting canvas-coordinate mapping is
\begin{equation}
f(t)=
\begin{cases}
\dfrac{a^\star}{a}\,t,
& 0\leq t<a,\\[4pt]
a^\star+\dfrac{\ell^\star}{\ell}(t-a),
& a\leq t\leq b,\\[4pt]
b^\star+\dfrac{L-b^\star}{L-b}(t-b),
& b<t\leq L.
\end{cases}
\label{eq:focus-axis}
\end{equation}
Empty outer segments are omitted.
Applying this construction along both axes yields
$\Phi_R(x,y)=(f_x(x),f_y(y))$,
with $L=w$ horizontally and $L=h$ vertically.

\paragraph{Topology preservation.}
Each axis mapping is continuous and strictly increasing,
with positive slopes on all nonempty segments and fixed
canvas endpoints.
Their product $\Phi_R$ is therefore a homeomorphism
of the continuous canvas, preserving horizontal and
vertical coordinate order without folds or cuts.
The selected evidence and surrounding landmarks remain
connected within one view as their scales change.

\paragraph{Pixel-budget allocation.}
The local area multiplier quantifies output-pixel
allocation relative to the uniform canvas layout.
Away from segment boundaries,
\begin{equation}
\rho_R(x,y)
=\det D\Phi_R(x,y)
=f_x'(x)f_y'(y).
\label{eq:focus-density}
\end{equation}
On the focus rectangle, it equals
$(\ell_x^\star/\ell_x)(\ell_y^\star/\ell_y)\geq1$,
where the subscripts identify the two axis spans.
Whenever either span increases, the focus receives
more output area.
Because $\Phi_R$ covers the same canvas,
\begin{equation}
\int_0^h\!\int_0^w
\rho_R(x,y)\,\mathrm{d}x\,\mathrm{d}y
=hw=B.
\label{eq:focus-budget}
\end{equation}
Local expansion thus reduces the total area assigned
to the periphery within the fixed budget.
We render directly from the original image $I$, using
the source boundaries retained in $R$ and the target
boundaries specified by $\Phi_R$.
Bilinear resampling proceeds first horizontally,
retaining the original source height, and then vertically
to produce the final $h\times w$ canvas:
$I^\star=\operatorname{Render}_{\mathrm{bilinear}}
(I,R,\Phi_R)$.
An empty request returns an aspect-preserving bilinear
resize of $I$ to $h\times w$.
The resulting focused view is passed to the answering stage.

\subsection{Evidence-Grounded Answering}
\label{sec:answering}

\par Following the topology-preserving spatial reallocation, the identical frozen MLLM performs the final reasoning step using the foveated canvas $I^*$. Because the foveation operator intrinsically preserves the global scene topology, the target evidence and its surrounding periphery are presented coherently within a single, continuous visual field. The MLLM is prompted with the original query $q$ alongside the focused image $I^*$ (annotated with its source label) to generate the final prediction. The instruction explicitly requires the model to ground its reasoning in the visible evidence and encapsulate the final output within \texttt{<answer>} tags, which is subsequently parsed by a lightweight extraction function to yield $\hat{y}$. 

\par This final step closes a unified perception-action-perception sequence: the query directs the gaze, the gaze dynamically reconfigures the visual observation, and the foveated observation supports the final semantic reasoning. By design, this end-to-end framework requires at most two MLLM calls per question. Furthermore, this architectural separation establishes a common modular basis for evaluating the roles of the observer, gaze depth, and evidence composition, which we extensively analyse in our ablation studies.

\begin{figure*}[!t]
      \centerline{\includegraphics[scale=0.48]{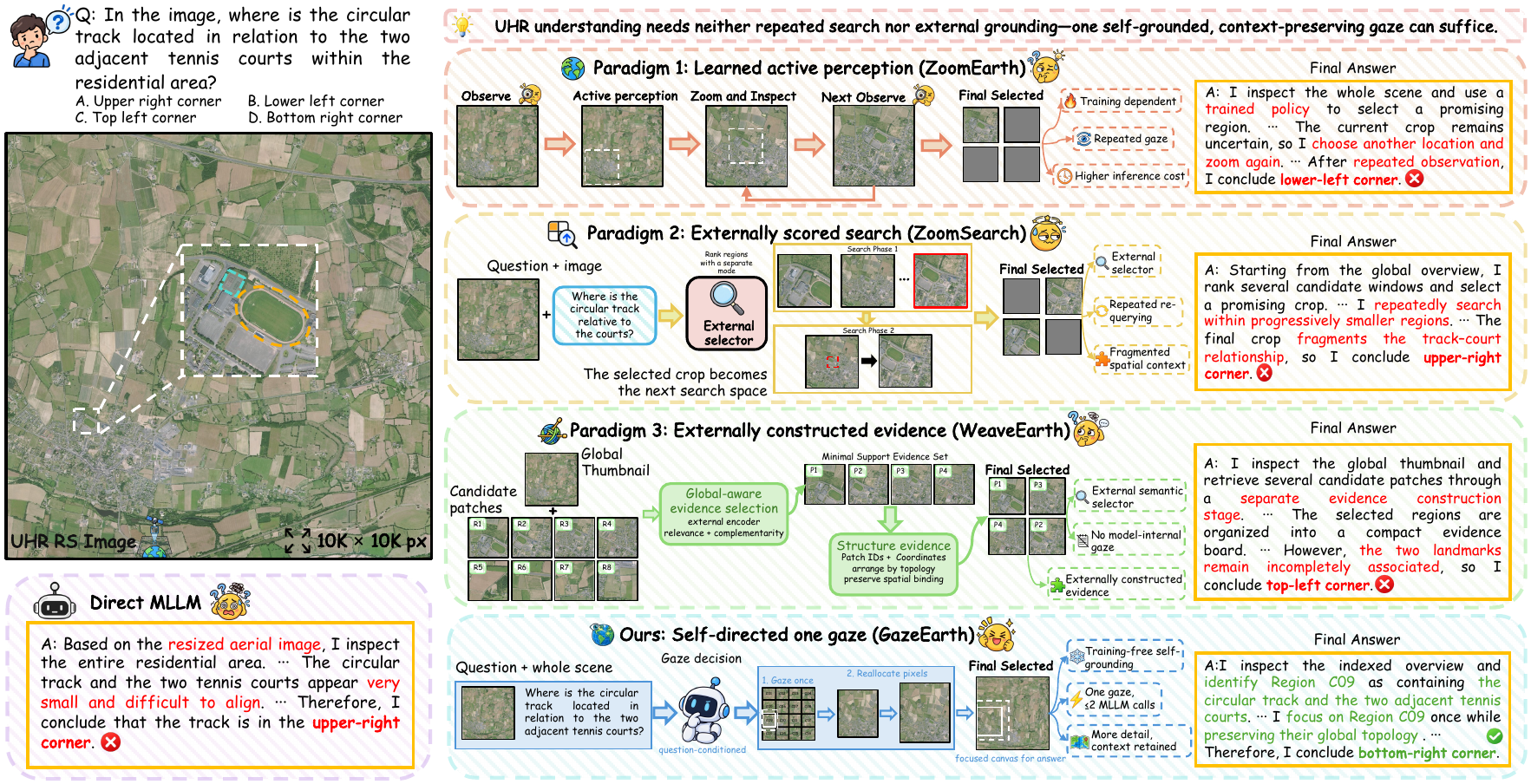}}
  \vspace{-0.1cm}
  \caption{Observation strategies on a spatial-relation question.
The example compares how different methods inspect a circular
track and adjacent tennis courts. GazeEarth enlarges both
landmarks within the full scene and correctly identifies
their relative position.}
  \label{fig-case-study}
\end{figure*}

\begin{table*}[t]
\centering
\caption{
Main results on LRS-GRO and MME-RealWorld-RS.
Scores (\%) follow the respective benchmark evaluation
protocols, with Avg.\ denoting the overall score
for each benchmark.
``--'' in the result columns indicates an unreported score.
}
\label{tab:lrs-gro-mme-main-comparison}
\vspace{-0.15cm}
\scriptsize
\renewcommand{\arraystretch}{0.88}
\setlength{\tabcolsep}{1.5pt}
\setlength{\aboverulesep}{0.25ex}
\setlength{\belowrulesep}{0.25ex}
\begin{tabularx}{\textwidth}{@{}lcc*{8}{Y}@{}}
\toprule
\multirow{2}{*}{\textbf{Method}}
& \multirow{2}{*}{\textbf{Param.}}
& \multirow{2}{*}{\textbf{TF}}
& \multicolumn{4}{c}{\textbf{LRS-GRO (\%) $\uparrow$}}
& \multicolumn{4}{c}{\textbf{MME-RealWorld-RS (\%) $\uparrow$}} \\
\cmidrule(lr){4-7}
\cmidrule(lr){8-11}
& & &
\textbf{Global}
& \textbf{Region}
& \textbf{Object}
& \textbf{Avg.}
& \textbf{Color}
& \textbf{Count}
& \textbf{Position}
& \textbf{Avg.} \\
\midrule
\multicolumn{11}{@{}l}{%
\textit{\textbf{Commercial MLLMs}}} \\[-0.25ex]
\textit{GPT-4o}~\citep{openai2024gpt4osystemcard}
& --
& \tfno
& 31.10
& 12.37
& 21.86
& 21.83
& 34.18
& 15.17
& 37.07
& 28.92 \\
\textit{Gemini-2.5-Flash}~\citep{geminiteam2025gemini}
& --
& \tfno
& 57.95
& 23.39
& 30.76
& 35.52
& \secondcell{55.86}
& \bestcell{33.28}
& 51.63
& 47.03\\
\midrule
\multicolumn{11}{@{}l}{%
\textit{\textbf{General-purpose MLLMs}}} \\[-0.25ex]
\textit{LLaVA-OV}~\citep{li2025llavaonevision}
& 7B
& \tfno
& 62.43
& 36.18
& 39.92
& 44.42
& 26.81
& 26.14
& 27.57
& 26.85 \\
\textit{LLaVA-v1.6}~\citep{liu2024llavanext}
& 7B
& \tfno
& 75.52
& 35.67
& 45.24
& 50.23
& 24.51
& 18.03
& 27.49
& 23.39 \\
\textit{Qwen2.5-VL}~\citep{bai2025qwen25vl}
& 7B
& \tfno
& 69.39
& 38.47
& 43.58
& 48.55
& 45.94
& 21.05
& 55.23
& 40.90 \\
\textit{Qwen3-VL}~\citep{bai2025qwen3vl}
& 8B
& \tfno
& 76.25
& 42.03
& 54.77
& 56.90
& 48.07
& 21.35
& 56.41
& 42.11 \\
\textit{Intern-S1-mini}~\cite{bai2025interns1}
& 8B
& \tfno
& 71.35
& 35.84
& 47.73
& 50.58
& 45.26
& 13.30
& \secondcell{66.11}
& 41.79 \\
\midrule
\multicolumn{11}{@{}l}{%
\textit{\textbf{Remote-sensing MLLMs}}} \\[-0.25ex]
\textit{RSUniVLM}~\citep{liu2026rsunivlm}
& 0.5B
& \tfno
& 63.02
& 34.10
& 37.55
& 42.83
& 28.97
& 19.00
& 28.33
& 25.48 \\
\textit{GeoChat}~\citep{kuckreja2024geochat}
& 7B
& \tfno
& 62.43
& 36.79
& 40.72
& 44.99
& 23.11
& 15.66
& 25.06
& 21.32 \\
\textit{VHM}~\citep{pang2025vhm}
& 7B
& \tfno
& 60.40
& 26.92
& 35.03
& 39.18
& 20.32
& 16.80
& 35.24
& 24.18 \\
\textit{GeoLLaVA-8K}~\citep{wang2025geollava8k}
& 7B
& \tfno
& 10.44
& 11.96
& 11.19
& 11.19
& 27.92
& 22.27
& 34.90
& 28.41 \\
\midrule
\multicolumn{11}{@{}l}{%
\textit{\textbf{UHR visual reasoning methods}}} \\[-0.25ex]
\textit{Coarse-to-Fine}~\citep{luo2025coarsetofine}
& 7B
& \tfno
& 71.35
& 35.32
& 36.52
& 44.57
& 44.70
& 31.00
& 49.72
& 41.89 \\
\textit{GeoLens}~\citep{wang2026beyondzooming}
& 7B
& \tfno
& \bestcell{79.00}
& 48.90
& \secondcell{57.60}
& \secondcell{60.67}
& 53.23
& 19.74
& 63.64
& 45.75 \\
\textit{ZoomEarth}~\citep{liu2026zoomearth}
& 3B
& \tfno
& 67.45
& 42.31
& 51.96
& 53.39
& 49.69
& \secondcell{31.66}
& 50.14
& 43.93 \\
\textit{ZoomSearch}~\citep{zhou2025zoomsearch}
& 8B
& \tfyes
& 67.14
& 46.78
& 52.04
& 54.41
& 41.67
& 17.86
& 54.97
& 38.34 \\
\textit{RADAR}~\citep{liu2026seeingclearly} $\dagger$
& 8B
& \tfyes
& --
& --
& --
& --
& 50.52
& 20.47
& 58.15
& 43.23 \\
\textit{WeaveEarth}~\citep{ma2026weaveearth}
& 8B
& \tfyes
& 75.34
& 47.74
& 56.50
& 58.94
& 51.39
& 27.65
& 62.60
& \secondcell{47.38} \\

\midrule
\multicolumn{11}{@{}l}{%
\textit{\textbf{General Focus Method for MLLMs}}} \\[-0.25ex]
\textit{AttWarp}~\citep{dalal2026constructive}
& 8B
& \tfyes
& 77.75 & 49.00 & 53.39 & 58.18 & 54.58 & 19.00 & 62.13 & 45.45 \\

\textit{ViCrop}~\citep{zhang2025mllmsknow}
& 8B
& \tfyes
& 77.45
& \bestcell{51.00}
& 53.82
& 58.80
& 51.87
& 17.54
& 62.13
& 44.06 \\

\midrule
\multicolumn{11}{@{}l}{%
\textit{\textbf{Ours}}} \\[-0.25ex]
\textit{\textbf{GazeEarth}}
& 8B
& \tfyes
& \secondcell{77.96}
& \secondcell{50.39}
& \bestcell{58.04}
& \bestcell{61.00}
& \bestcell{63.67}
& 21.78
& \bestcell{71.12}
& \bestcell{52.43} \\
\bottomrule
\end{tabularx}
\vspace{-0.1cm}
\end{table*}

\section{Experiments}

\subsection{Experimental Setup}
\label{sec:setup}

\paragraph{Benchmarks and Metrics.}
We evaluate GazeEarth on LRS-GRO~\citep{liu2026zoomearth}, MME-RealWorld-RS~\citep{zhang2025mmerealworld}, and XLRS-Bench~\citep{wang2025xlrsbench}, following the original evaluation protocols and metrics of each benchmark.

\paragraph{Backbones, Baselines, and Implementation.}
We use Qwen3-VL-8B as the primary backbone and assess cross-backbone generalization with LLaVA-v1.6-7B and Intern-S1-mini. We compare GazeEarth against commercial MLLMs, general-purpose MLLMs, remote-sensing MLLMs, and UHR visual reasoning methods. Unless otherwise specified, GazeEarth uses a $4\times4$ indexed overview with the longest side capped at 1,536 pixels and selects up to two regions in a single gaze round. We set the focus scale to 1.5 and cap the longest side of each answer-stage view at 2,048 pixels. The maximum generation length for the final answer is 1,024 tokens. 
The setting of Commercial MLLMs, prompts, and further implementation details are provided in the Appendix~\ref{app:all}.

\begin{table*}[!t]
\centering
\caption{
Performance (\%) on XLRS-Bench across eight categories:
Counting (Cnt), Scene Classification (SC),
Object Spatial Relationship (OSR), Object Properties (OP),
Planning (Plan), Anomaly Reasoning (AR),
Complex Reasoning (CR), and Spatiotemporal Reasoning (SR).
}
\label{tab:XLRS-Bench-comparison}
\vspace{-0.1cm}
\scriptsize
\renewcommand{\arraystretch}{0.92}
\setlength{\tabcolsep}{1.5pt}
\setlength{\aboverulesep}{0.25ex}
\setlength{\belowrulesep}{0.25ex}
\begin{tabularx}{\textwidth}{@{}lcc*{9}{Y}@{}}
\toprule
\multirow{2}{*}{\textbf{Method}}
& \multirow{2}{*}{\textbf{Param.}}
& \multirow{2}{*}{\textbf{TF}}
& \multicolumn{9}{c}{%
\textbf{XLRS-Bench (\%) $\uparrow$}} \\
\cmidrule(lr){4-12}
& & &
\textbf{Cnt}
& \textbf{SC}
& \textbf{OSR}
& \textbf{OP}
& \textbf{Plan}
& \textbf{AR}
& \textbf{CR}
& \textbf{SR}
& \textbf{Avg.} \\
\midrule
\multicolumn{12}{@{}l}{%
\textit{\textbf{Commercial MLLMs}}} \\
\textit{GPT-4o}~\citep{openai2024gpt4osystemcard}
& --
& \tfno
& 32.50 & 47.67 & 35.20 & 37.47 & 36.00 & 72.00 & 58.50 & 31.67 & 40.16 \\

\textit{Gemini-2.5-Flash}~\citep{geminiteam2025gemini}
& --
& \tfno
& \secondcell{38.75} & 51.67 & 30.40 & 39.10 & 42.00
& \secondcell{78.00} & 65.50 & 23.33 & 41.66 \\
\midrule
\multicolumn{12}{@{}l}{%
\textit{\textbf{General-purpose MLLMs}}} \\
\textit{LLaVA-OV}~\citep{li2025llavaonevision}
& 7B
& \tfno
& 31.25
& 35.67
& 25.20
& 45.42
& 24.00
& 76.00
& 59.50
& 43.33
& 41.62 \\
\textit{LLaVA-v1.6}~\citep{liu2024llavanext}
& 7B
& \tfno
& 24.38
& 21.33
& 23.80
& 21.27
& 40.00
& 35.00
& 34.50
& 33.33
& 24.00 \\
\textit{Qwen2.5-VL}~\citep{bai2025qwen25vl}
& 7B
& \tfno
& 30.63
& 40.33
& 30.80
& 43.31
& 32.00
& 67.00
& 62.50
& 43.33
& 41.98 \\
\textit{Qwen3-VL}~\citep{bai2025qwen3vl}
& 8B
& \tfno
& 32.50
& 41.67
& 31.40
& 43.92
& 34.00
& 70.00
& 63.50
& 46.33
& 42.92 \\
\textit{CogVLM2}~\citep{hong2024cogvlm2}
& 8B
& \tfno
& 36.88 & 46.33 & 36.20 & 36.87 & 34.00 & 69.00 & 60.50 & - & 40.20 \\
\textit{Intern-S1-mini}~\citep{bai2025interns1}
& 8B
& \tfno
& 35.63 & 58.00 & 32.80 & 38.80 & 51.00 & 69.00 & 66.00 & \bestcell{58.33} & 43.05 \\
\midrule
\multicolumn{12}{@{}l}{%
\textit{\textbf{Remote-sensing MLLMs}}} \\
\textit{RSUniVLM}~\citep{liu2026rsunivlm}
& 0.5B
& \tfno
& 21.88 & 24.67 & 34.40 & 30.18 & 20.00 & 28.00 & 28.00 & 25.00 & 29.25 \\
\textit{GeoChat}~\citep{kuckreja2024geochat}
& 7B
& \tfno
& 22.50
& 12.67
& 24.20
& 24.28
& 10.00
& 33.00
& 32.00
& --
& 23.35 \\
\textit{VHM}~\citep{pang2025vhm}
& 7B
& \tfno
& 32.50 & 33.33 & 31.40 & 26.69 & 33.00 & 58.00 & 49.50 & 31.67 & 31.20 \\
\textit{GeoLLaVA-8K}~\citep{wang2025geollava8k}
& 7B
& \tfno
& 35.00
& 62.00
& 35.80
& 37.71
& \bestcell{65.00}
& 68.00
& 63.50
& 51.67
& 43.44 \\
\midrule
\multicolumn{12}{@{}l}{%
\textit{\textbf{UHR visual reasoning methods}}} \\
\textit{Coarse-to-Fine}~\citep{luo2025coarsetofine}
& 7B
& \tfno
& 37.50 & 36.33 & 25.40 & 30.66 & 27.00 & 70.00 & 58.50 & 51.67 & 34.09\\
\textit{GeoLens}~\citep{wang2026beyondzooming}
& 7B
& \tfno
&36.25 & \secondcell{62.33} & 34.40 & 45.96 & \secondcell{58.00} & 66.00 & 65.00 & 48.33 & \secondcell{47.50} \\
\textit{ZoomEarth}~\citep{liu2026zoomearth}
& 3B
& \tfno
& 35.00
& 44.67
& 31.40
& 35.60
& 22.00
& 66.00
& 56.50
& 28.30
& 37.53 \\
\textit{ZoomSearch}~\citep{zhou2025zoomsearch}
& 8B
& \tfyes
& 37.50 & 59.33 & 25.80 & 41.99 & 34.00 & 69.00 & 68.00 & 43.33 & 43.15\\

\textit{WeaveEarth}~\citep{ma2026weaveearth}
& 8B
& \tfyes
& 37.50
& 50.33
& \secondcell{36.40}
& \secondcell{47.41}
& 36.00
& 73.00
& \secondcell{67.50}
& 46.33
& 47.14 \\
\midrule
\multicolumn{12}{@{}l}{%
\textit{\textbf{General Focus Method for MLLMs}}} \\
\textit{AttWarp}~\citep{dalal2026constructive}
& 8B
& \tfyes

& 33.75 & 58.67 & 31.00 & 44.94 & 39.00 & 74.00 & 62.50 & 46.67 & 45.36\\

\textit{ViCrop}~\citep{zhang2025mllmsknow}
& 8B
& \tfyes
& 31.88 & 59.00 & 29.40 & 44.82 & 37.00 & 74.00 & 63.00 & 45.00 & 44.90 \\
\midrule
\multicolumn{12}{@{}l}{%
\textit{\textbf{Ours}}} \\
\textit{\textbf{GazeEarth}}
& 8B
& \tfyes
& \bestcell{44.38}
& \bestcell{66.67}
& \bestcell{37.80}
& \bestcell{53.07}
& 49.00
& 76.00
& \bestcell{74.50}
& \secondcell{55.00}
& \bestcell{53.51} \\
\bottomrule
\end{tabularx}
\end{table*}

\subsection{Main Results}

\paragraph{Overall performance.}
As shown in Tables~\ref{tab:lrs-gro-mme-main-comparison} and~\ref{tab:XLRS-Bench-comparison}, GazeEarth achieves the best overall accuracy among the compared methods on all three benchmarks, reaching 61.00\% on LRS-GRO, 52.43\% on MME-RealWorld-RS, and 53.51\% on XLRS-Bench. Compared with direct inference using the same frozen Qwen3-VL-8B backbone, these results correspond to absolute gains of 4.10, 10.32, and 10.59 percentage points, respectively. The gains are not limited to aggregate performance:
relative to Overview, GazeEarth improves Object
accuracy on LRS-GRO by 1.82 percentage points
and Color accuracy on MME-RealWorld-RS by 8.13 points.
The additional changes in LRS-GRO Region and
MME-RealWorld-RS Position are +0.17 and -1.27 points,
respectively.
Appendix Figure~\ref{fig:category-gain} reports
category-level gains over Direct, while Tables~\ref{tab:category-level-controlled-analysis}
--~\ref{tab:category-level-controlled-analysis-mme-realworld-rs} distinguish these gains from the additional
changes over Overview.

\paragraph{Case study.}
Figure~\ref{fig-case-study} illustrates the change in observation
on a spatial-relation question.
GazeEarth selects the correct region and enlarges the circular track and
adjacent tennis courts within the full scene.
Both landmarks become easier to inspect while their relative
position is retained, supporting the correct answer,
``bottom-right corner.''

\begin{table}[t]
\centering
\begin{minipage}[t]{0.545\textwidth}
\centering
\caption{GazeEarth across four frozen backbones.
Gain denotes the percentage-point improvement over
the corresponding Direct baseline.
Mean is the unweighted average of the three benchmark scores.}
\label{tab:cross-backbone-gazeearth}
\scriptsize
\renewcommand{\arraystretch}{1.25}
\setlength{\tabcolsep}{1.45pt}
\setlength{\aboverulesep}{0.25ex}
\setlength{\belowrulesep}{0.25ex}
\begin{tabularx}{\linewidth}{
@{}
>{\itshape\raggedright\arraybackslash}p{0.28\linewidth}
>{\itshape\raggedright\arraybackslash}p{0.20\linewidth}
*{4}{Y}
@{}
}
\toprule

\raisebox{0.5\baselineskip}[0pt][0pt]{%
\normalfont\bfseries Backbone}
&
\raisebox{0.5\baselineskip}[0pt][0pt]{%
\normalfont\bfseries Method}
&
\shortstack[c]{\textbf{LRS-}\\\textbf{GRO}}
&
\shortstack[c]{\textbf{MME-}\\\textbf{RS}}
&
\shortstack[c]{\textbf{XLRS-}\\\textbf{Bench}}
&
\raisebox{0.5\baselineskip}[0pt][0pt]{\textbf{Mean}}
\\
\midrule

\rowcolor{directrow}
Qwen3-VL-8B & Direct
& 56.90 & 42.11 & 42.92 & 47.31 \\

\rowcolor{overviewrow}
& Overview
& 59.99 & 49.01 & 45.13 & 51.38 \\

\rowcolor{gazerow}
& \textbf{+ GazeEarth}
& \textbf{61.00} & \textbf{52.43}
& \textbf{53.51} & \textbf{55.65} \\

\rowcolor{relativerow}
& Gain
& +4.10 & +10.32 & +10.59 & +8.34 \\

\midrule

\rowcolor{directrow}
LLaVA-v1.6-7B & Direct
& 50.23 & 23.39 & 24.00 & 32.54 \\

\rowcolor{overviewrow}
& Overview
& 50.99 & 30.16 & 34.35 & 38.50 \\

\rowcolor{gazerow}
& \textbf{+ GazeEarth}
& \textbf{54.36} & \textbf{33.90}
& \textbf{37.47} & \textbf{41.91} \\

\rowcolor{relativerow}
& Gain
& +4.13 & +10.51 & +13.47 & +9.37 \\

\midrule

\rowcolor{directrow}
Intern-S1-mini & Direct
& 50.58 & 41.79 & 43.05 & 45.14 \\

\rowcolor{overviewrow}
& Overview
& 52.08 & 43.10 & 42.69 & 45.96 \\

\rowcolor{gazerow}
& \textbf{+ GazeEarth}
& \textbf{55.95} & \textbf{46.44}
& \textbf{46.82} & \textbf{49.74} \\

\rowcolor{relativerow}
& Gain
& +5.37 & +4.65 & +3.77 & +4.60 \\

\midrule

\rowcolor{directrow}
GPT-4o & Direct
& 21.83 & 28.92 & 40.16 & 30.30 \\

\rowcolor{overviewrow}
& Overview
& 23.20 & 32.27 & 41.03 & 32.17 \\

\rowcolor{gazerow}
& \textbf{+ GazeEarth}
& \textbf{26.79} & \textbf{36.18}
& \textbf{44.29} & \textbf{35.75} \\

\rowcolor{relativerow}
& Gain
& +4.96 & +7.26 & +4.13 & +5.45 \\

\bottomrule
\end{tabularx}
\end{minipage}\hfill
\begin{minipage}[t]{0.43\textwidth}
\centering
\caption{Observer substitution with Qwen3-VL-8B as the fixed
answerer.}
\label{tab:observer-substitution-qwen3vl8b}
\scriptsize
\renewcommand{\arraystretch}{0.92}
\setlength{\tabcolsep}{1.5pt}
\setlength{\aboverulesep}{0.25ex}
\setlength{\belowrulesep}{0.25ex}
\begin{tabularx}{\linewidth}{
@{}
>{\itshape\raggedright\arraybackslash}p{0.25\linewidth}
>{\hsize=0.95\hsize\linewidth=\hsize}Y
>{\hsize=0.85\hsize\linewidth=\hsize}Y
>{\hsize=1.30\hsize\linewidth=\hsize}Y
>{\hsize=0.80\hsize\linewidth=\hsize}Y
>{\hsize=1.10\hsize\linewidth=\hsize}Y
@{}
}
\toprule
\raisebox{1.5\baselineskip}[0pt][0pt]{%
\normalfont\bfseries Observer}
&
\raisebox{0.5\baselineskip}[0pt][0pt]{%
\shortstack[c]{%
\textbf{XLRS-}\\
\textbf{Bench}\\
\textbf{(\%)} $\uparrow$}}
&
\raisebox{0.5\baselineskip}[0pt][0pt]{%
\shortstack[c]{%
\textbf{LRS-}\\
\textbf{GRO}\\
\textbf{(\%)} $\uparrow$}}
&
\shortstack[c]{%
\textbf{MME-}\\
\textbf{Real-}\\
\textbf{World-RS}\\
\textbf{(\%)} $\uparrow$}
&
\raisebox{1.5\baselineskip}[0pt][0pt]{%
\textbf{Avg.}}
&
\raisebox{1.0\baselineskip}[0pt][0pt]{%
\shortstack[c]{%
\textbf{$\Delta$ vs.}\\
\textbf{Random}}}
\\
\midrule

Random
& 45.42 & 58.97 & 45.64 & 50.01 & -- \\

SigLIP2
& \secondcell{46.10}
& 59.54
& \secondcell{47.49}
& \secondcell{51.04}
& \secondcell{+1.03} \\

RemoteCLIP
& 45.94
& \secondcell{60.10}
& 47.08
& \secondcell{51.04}
& \secondcell{+1.03} \\

\midrule

\textbf{GazeEarth}
& \bestcell{53.51}
& \bestcell{61.00}
& \bestcell{52.43}
& \bestcell{55.65}
& \bestcell{+5.64} \\

\bottomrule
\end{tabularx}

\vspace{10pt}
\includegraphics[width=\linewidth]{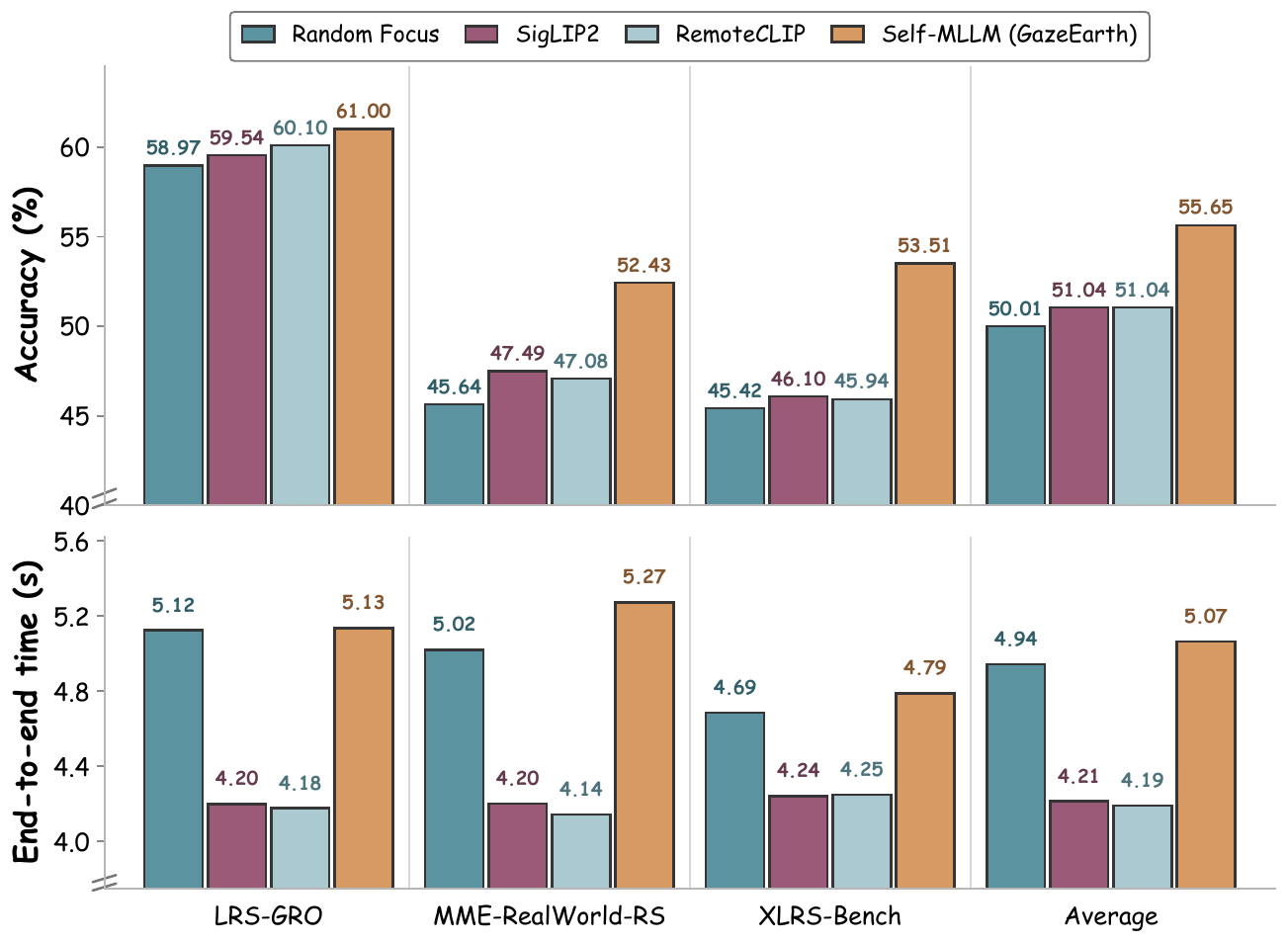}
\captionof{figure}{Accuracy and end-to-end inference time
under observer substitution.}
\label{fig:latency}
\end{minipage}
\end{table}

\paragraph{Cross-backbone performance.}
The gains extend beyond the primary Qwen backbone.
With each backbone generating its own spatial requests
and answering from the resulting focused views,
GazeEarth improves all three benchmark scores for
Qwen3-VL-8B, LLaVA-v1.6-7B, Intern-S1-mini and GPT-4o
(Table~\ref{tab:cross-backbone-gazeearth}).
Their benchmark-averaged gains over Direct are
8.34, 9.37, 4.60, and 5.45 percentage points,
respectively, with additional gains of
4.27, 3.41, 3.78, and 3.58 points over Overview.
The same observation procedure thus benefits all
four evaluated backbones without updating their
parameters.

\paragraph{Question-guided allocation.}
We next vary who selects the regions while keeping
Qwen3-VL-8B as the answerer and retaining the same
focused-view answering pipeline
(Table~\ref{tab:observer-substitution-qwen3vl8b}).
Random Focus achieves a benchmark-averaged score of 50.01\%;
SigLIP2 and RemoteCLIP each reach 51.04\%;
GazeEarth reaches 55.65\%.
Self-directed gaze improves over both external observers
on every benchmark, with a mean advantage of
4.61 percentage points.
Together with the localization probe, this shows that
the frozen MLLM's spatial requests are useful both
for locating evidence and for guiding observations
that support answering.
Appendix Figure~\ref{fig:hyper} reports sensitivity to
grid size, region budget, focus scale, and selector resolution.

\paragraph{Allocation matters more than input volume.}
The three answering conditions use the same frozen backbone but
differ in both the amount and spatial allocation of visual input.
Direct achieves 47.31\%
at 7.34 seconds per question.
Overview improves accuracy
to 51.38\% while reducing runtime to 4.02 seconds.
GazeEarth preserves the same canvas and pixel budget
(Eq.~\ref{eq:focus-budget}) but reallocates pixels according to
the question, reaching 55.65\% in 5.07 seconds
(Tables~\ref{tab:visual-evidence-presentation}
and~\ref{tab:qwen3-vl-inference-efficiency}).
Thus, more visual input is not necessarily better:
a smaller, well-structured input is both faster and more accurate,
and question-guided reallocation yields a further gain.
Of the 8.34-point improvement over Direct, 4.07 points come from
canvas normalization and the remaining 4.27 from directed focus.

\begin{table*}[htbp]
\centering

\caption{Ablation study on the three benchmarks.
Gaze denotes the number of region-selection rounds;
"Think with Image" indicates that both global and cropped views are provided to the model.}
\label{tab:visual-evidence-presentation}

\vspace{-0.15cm}

\scriptsize
\renewcommand{\arraystretch}{0.95}
\setlength{\tabcolsep}{1.5pt}
\setlength{\aboverulesep}{0.25ex}
\setlength{\belowrulesep}{0.25ex}

\begin{tabularx}{\textwidth}{@{}lccccc*{4}{Y}@{}}
\toprule

\multirow[c]{2}{*}{\textbf{Variant}}
& \multirow[c]{2}{*}{\textbf{Gaze}}
& \multirow[c]{2}{*}{%
    \shortstack[c]{%
      \textbf{Q-guided}\\
      \textbf{gaze}%
    }%
  }
& \multirow[c]{2}{*}{\textbf{Detail}}
& \multirow[c]{2}{*}{%
    \shortstack[c]{%
      \textbf{Full}\\
      \textbf{scene}%
    }%
  }
& \multirow[c]{2}{*}{%
    \shortstack[c]{%
      \textbf{Continuous}\\
      \textbf{Binding}%
    }%
  }
& \multicolumn{4}{c}{%
    \textbf{Performance (\%)} $\uparrow$%
  }
\\

\cmidrule(lr){7-10}

& & & & & &
\shortstack[c]{%
  \textbf{XLRS-}\\
  \textbf{Bench}%
}
&
\shortstack[c]{%
  \textbf{LRS-}\\
  \textbf{GRO}%
}
&
\shortstack[c]{%
  \textbf{MME-}\\
  \textbf{RealWorld-RS}%
}
&
\raisebox{0.5\baselineskip}[0pt][0pt]{%
  \textbf{Avg.}%
}
\\

\midrule

\textit{Direct}
& 0
& \tfno
& \tfno
& \tfyes
& --
& 42.92
& 56.90
& 42.11
& 47.31 \\

\textit{Overview}
& 0
& \tfno
& \tfno
& \tfyes
& --
& 45.13
& 59.99
& 49.01
& 51.38 \\

\textit{Gaze + Crop}
& 1
& \tfyes
& \tfyes
& \tfno
& \tfno
& 50.55
& 58.93
& 39.81
& 49.76 \\

\textit{Think with Image}
& 1
& \tfyes
& \tfyes
& \tfyes
& \tfno
& 51.43
& \secondcell{60.54}
& 42.24
& 51.40 \\

\textit{Linked-Global + Crop}
& 1
& \tfyes
& \tfyes
& \tfyes
& \tfyes
& \secondcell{52.56}
& 60.25
& 43.39
& 52.07 \\

\midrule

\textit{Two-Gaze}
& 2
& \tfyes
& \tfyes
& \tfyes
& \tfyes 
& 46.27
& 60.12
& \secondcell{52.33}
& \secondcell{52.91} \\

\textit{Three-Gaze}
& 3
& \tfyes
& \tfyes
& \tfyes
& \tfyes 
& 46.36
& 60.04
& 51.77
& 52.72 \\

\midrule

\textit{\textbf{GazeEarth}}
& 1
& \tfyes
& \tfyes
& \tfyes
& \tfyes 
& \bestcell{53.51}
& \bestcell{61.00}
& \bestcell{52.43}
& \bestcell{55.65} \\

\bottomrule
\end{tabularx}

\end{table*}

\subsection{Ablation Study}
\label{sec: ablation}

\paragraph{Evidence presentation.}
Table~\ref{tab:visual-evidence-presentation} compares
full-scene focus with an unfocused overview and
crop-based evidence presentation.
GazeEarth achieves a benchmark-averaged score of 55.65\%,
compared with 51.38\% for Overview, 49.76\% for
Gaze + Crop, and 51.40\% for Think with Image (a global overview and a separate crop without explicit spatial linkage).
The difference is largest on MME-RealWorld-RS:
continuous full-scene focus reaches 52.43\%,
versus 39.81\% and 42.24\% for the two crop-based variants.
On XLRS-Bench, GazeEarth reaches 53.51\%, compared with
50.55\% for Gaze + Crop, 51.43\% for Think with Image,
and 52.56\% for Linked Global + Crop.
Appendix \ref{app:extended} additionally evaluates AttWarp warping.
GazeEarth also achieves the highest three-benchmark mean
among the presentation variants evaluated on all three
benchmarks.
These results show that region selection alone does not
determine answering performance: how the selected evidence
is presented also matters. The comparison with Direct measures the overall gain
of the GazeEarth pipeline, whereas the comparison with
Overview measures its additional benefit over
full-scene answering.

\section{Related Work}
\label{sec:related-work}

\paragraph{Remote Sensing Multimodal Large  Language Models.}
Remote sensing MLLMs leverage descriptive and geographic
supervision~\citep{hu2025rsgpt,muhtar2024lhrsbot},
grounded dialogue~\citep{kuckreja2024geochat},
multisensor alignment~\citep{zhang2024earthgpt},
and geospatial reasoning~\citep{liu2026geocot}.
For UHR imagery, GeoLLaVA-8K retains informative
tokens~\citep{wang2025geollava8k},
ZoomSearch searches and reassembles
patches~\citep{zhou2025zoomsearch},
and WeaveEarth integrates local evidence, global context,
and spatial metadata~\citep{ma2026weaveearth}.
These passive token purification strategies rely heavily on how patches are selected and struggle to preserve spatial context.

\paragraph{Active Vision in General Domains and Remote Sensing.}
Active vision uses language-guided search~\citep{wu2024vstar}, cropping~\citep{zhang2023perceivingsmalldetails}, and learned inspection~\citep{su2025pixelreasoner,lee2026ergo,lin2026adaptvision}.
Internal attention also provides localization cues~\citep{zhang2025mllmsknow,shi2026catchingdetails}.
Training-free methods collect spatially arranged crops~\citep{wang2025rap} or hierarchical evidence~\citep{li2026deepscan}.
In remote sensing, learned zooming~\citep{liu2026zoomearth} and tool use~\citep{wang2026beyondzooming,wang2026textvisionstagedknowledge}
require post-training, while repeated cropping adds inference rounds and cross-view integration. Nonuniform sampling reallocates resolution~\citep{recasens2018learningtozoom,thavamani2021fovea};
AttWarp uses cross-modal attention to guide context-preserving image warping~\citep{dalal2026constructive}. GazeEarth instead elicits an explicit spatial request through the model's text-generation interface. This request specifies the focus neighborhood for a deterministic piecewise-linear mapping that samples the original UHR image into a fixed-size full-scene canvas.

\section{Conclusion}
In this work, we view UHR remote sensing understanding as a problem of active foveated representations. 
Our analysis shows that frozen MLLMs can already localize relevant evidence, but existing crop-based thinking-with-image methods do not consistently improve answering, which highlights the importance of how selected evidence is presented. 
Based on this insight, we introduced GazeEarth, a simple-yet-effective training-free framework that converts explicit spatial requests into topology-preserving, full-scene focused observations. 
Across three benchmarks and four frozen backbones,
GazeEarth consistently improves over direct and
overview answering, with benchmark-averaged gains
of 4.6--9.4 and 3.4--4.3 percentage points,
respectively.
These results suggest that effective observation construction, rather than repeated visual search alone, is a promising direction for improving MLLM reasoning over UHR imagery.

\bibliographystyle{iclr2026_conference}
\begingroup
\urlstyle{same}
\interlinepenalty=10000
\bibliography{iclr2026_conference}
\endgroup

\appendix

\section{Experimental and Implementation Details}
\label{app:all}
This section details the prompts, image preprocessing and decoding settings, benchmark evaluation protocols, and hardware and timing configurations used in our experiments. The supplementary analyses separate three questions: whether the model produces useful spatial requests (Appendix~\ref{app:pilot}), how these requests should be rendered for answering (Appendix~\ref{app:extended}), and what accuracy--cost trade-offs follow from the resulting observation pipeline (Appendices~\ref{app:Statistical}--\ref{app:sensitivity}). Throughout, we report where each comparison is controlled and where several factors change together, so that end-to-end gains are not attributed to a single component without a corresponding control.

\subsection{Prompts}
\label{app:prompts}

GazeEarth uses two stage-specific prompts for the same
model: one elicits a spatial request from the indexed
overview, and the other elicits an answer from the
resulting full-scene observation.
The request determines the visual input of the
answering call, which receives the question and
the rendered view as a fresh interaction.

\paragraph{Question-guided gaze.}
The selection prompt combines the original question
with the indexed source overview and requests
a compact JSON response.
The placeholder \texttt{\{question\}} contains the
question text, and \texttt{\{source\_ids\}} lists
the source identifiers, such as \texttt{I1}
or \texttt{I1, I2}.

\begin{Verbatim}[
fontsize=\small,
breaklines=true,
breakanywhere=true,
frame=single,
framesep=2mm
]
You are a visual relevance selector for a remote-sensing question, not an answerer.

Question: {question}
Indexed source overview(s): {source_ids}. Each source uses the same 4x4 neutral grid.

Choose the smallest visual brief that would let another expert answer:
- global_only: the overview already contains sufficient visual evidence;
- selected: magnify one or two cells needed for a local detail, local relation, or two-region comparison;
- brief_incomplete: the question requires evidence scattered over more than two regions, exhaustive counting/distribution, or cannot be represented by this brief.

Return JSON only. Never answer the question, describe objects, or provide reasoning.
Valid forms:
{"status":"global_only","region_ids":[]}
{"status":"selected","region_ids":["I1:C07"]}
{"status":"selected","region_ids":["I1:C04","I1:C12"]}
{"status":"brief_incomplete","region_ids":[]}
\end{Verbatim}

\paragraph{Evidence-grounded answering.}
The answering prompt accompanies one rendered full-scene view
per source image and requests a reasoning explanation followed
by the final answer enclosed in \texttt{<answer>} tags.
For multiple-choice questions,
\texttt{\{answer\_question\}} contains the question
followed by \texttt{Options:} and the letter-labelled
choices in their original order.
For open-ended questions, it contains the question alone.
GazeEarth and the Overview condition share this
answering template.

\begin{Verbatim}[
fontsize=\small,
breaklines=true,
breakanywhere=true,
frame=single,
framesep=2mm
]
The visual input contains one full-scene view for each remote-sensing source. Any labels are neutral correspondence cues only. 
Answer the visual question using only visible evidence. Explain your reasoning before giving the final answer.
Question: {answer_question}
Enclose only the final answer in <answer>...</answer> tags.
\end{Verbatim}

\subsection{Backbone Interfaces and Inference Protocol}
\label{app:backbones}

\paragraph{Visual interfaces.}
We use Qwen3-VL-8B, LLaVA-v1.6-Mistral-7B, and
Intern-S1-mini through their checkpoint-specific image
processors and chat templates.
Table~\ref{tab:backbone-inference-config} summarizes
their visual interfaces and generation settings.
Each source image is converted to RGB.
For question-guided gaze, we construct an indexed overview
with a long-side cap of 1536 pixels and a $4\times4$ grid.
The resulting overview and question form the selection input.
For evidence-grounded answering, the spatial request guides
resampling from the original image onto an aspect-preserving
canvas with a long-side cap of 2048 pixels.
The rendered view carries a neutral source identifier
without the selection grid.
These size limits apply to the scene content before
source-label headers and backbone-specific preprocessing.

\begin{table}[t]
\centering
\caption{Backbone interfaces and inference settings.
Local backbones follow their checkpoint-default visual processing,
whereas GPT-4o uses provider-managed visual preprocessing.
Scene-size caps apply before source-label headers and
backbone-specific or provider-side preprocessing.
Generation limits count new tokens.}
\label{tab:backbone-inference-config}

\scriptsize
\renewcommand{\arraystretch}{1.12}
\setlength{\tabcolsep}{2.8pt}

\begin{tabularx}{\linewidth}{
@{}>{\raggedright\arraybackslash}p{0.235\linewidth}
*{4}{>{\centering\arraybackslash}X}
@{}}
\toprule
Setting
& \shortstack{Qwen3-VL\\8B}
& \shortstack{LLaVA-v1.6\\Mistral-7B}
& \shortstack{Intern-S1\\mini}
& \shortstack{Commercial MLLMs\\API} \\
\midrule

Execution interface
& Local
& Local
& Local
& OpenRouter \\

Vision patch size
& $16\times16$
& $14\times14$
& $14\times14$
& Not disclosed \\

Vision input unit
& Variable-size image
& $336\times336$ tile
& $448\times448$ tile
& PNG/JPEG data URL \\

Resolution handling
& Dynamic resolution
& AnyRes tiling
& Dynamic tiling
& Provider-managed \\

Default processor limit
& 16,777,216 pixels
& 4 local tiles
& 12 local tiles
& Not disclosed \\

Additional global view
& --
& 1
& 1 when tiled
& Not disclosed \\

API image-detail mode
& --
& --
& --
& \texttt{high} \\

\midrule

Gaze overview long-side cap
& \multicolumn{4}{c}{1536 pixels} \\

Answer canvas long-side cap
& \multicolumn{4}{c}{2048 pixels} \\

Selection grid
& \multicolumn{4}{c}{$4\times4$} \\

Maximum selected cells
& \multicolumn{4}{c}{2} \\

Decoding
& Greedy
& Greedy
& Greedy
& Temperature 0 \\

Gaze \texttt{max\_new\_tokens}
& \multicolumn{4}{c}{64} \\

Answer \texttt{max\_new\_tokens}
& \multicolumn{4}{c}{1024} \\

\bottomrule
\end{tabularx}

\smallskip
\begin{minipage}{\linewidth}
\footnotesize
Qwen3-VL applies $2\times2$ spatial patch merging.
LLaVA's AnyRes candidate resolutions are
$336\times672$, $672\times336$, $672\times672$,
$1008\times336$, and $336\times1008$.
Intern-S1-mini adds a global thumbnail when more than
one local tile is produced.
GPT-4o is accessed through the OpenRouter Chat Completions
API using \texttt{openai/gpt-4o-2024-11-20} with
\texttt{detail=high} and temperature 0.
Visual inputs are transmitted as base64-encoded PNG images,
with JPEG quality 95 used only when an encoded image exceeds
29 MB. Provider-side patching and the effective encoder
resolution are not exposed by the API.
\end{minipage}
\end{table}

\paragraph{Model calls and decoding.}
The same backbone performs region selection and answering, with its parameters fixed throughout inference.
Each stage uses a separate interaction containing its visual input and stage-specific prompt.
The answering call receives the rendered observation and the original question, without the selector's conversation history.
The selector's effect is therefore mediated through the constructed visual observation, rather than through
forwarded reasoning text or conversation history.
Local backbones use greedy decoding with sampling disabled, so temperature and nucleus sampling do not affect token
selection.
closed source models use the API configuration listed in
Table~\ref{tab:backbone-inference-config}, including temperature 0.
All backbones use output limits of 64 new tokens for question-guided gaze and 1024 for evidence-grounded answering.

\paragraph{Hyperparameter configuration.}
The default configuration uses a $4\times4$ indexed grid,
a maximum of $K=2$ selected cells, a focus scale of
$\alpha=1.5$, and a selector overview with a
1536-pixel long-side cap.
The grid provides 16 spatial addresses for expressing
a compact observation request.
Allowing two cells accommodates both local inspection
and comparisons between two regions.
The focus scale controls the enlargement of the selected
neighbourhood relative to its uniformly resized extent,
while the selector resolution controls the image size
used to generate the request.
The same configuration is used across the three benchmarks. The sensitivity experiments evaluate
variations around this configuration.

\subsection{Benchmarks and Evaluation Protocols}
\label{app:benchmarks}

\paragraph{Benchmark composition.}
We evaluate on LRS-GRO, MME-RealWorld-RS, and XLRS-Bench,
covering open-ended and multiple-choice remote-sensing
question answering.
Table~\ref{tab:benchmark-protocol} summarizes the evaluation
sets and reporting categories.
LRS-GRO contains questions at global, region, and object
levels.
MME-RealWorld-RS evaluates color, counting, and position
recognition.
XLRS-Bench covers eight categories spanning visual
perception and reasoning.
The sample counts refer to question--answer instances.
The localization probe uses separately defined eligible
cohorts rather than the complete answer-evaluation sets.

\begin{table}[t]
\centering
\caption{Benchmark composition and evaluation scope.
Counts denote question--answer instances in the evaluation
sets used in this work.}
\label{tab:benchmark-protocol}
\small
\renewcommand{\arraystretch}{1.15}
\setlength{\tabcolsep}{4pt}

\begin{tabularx}{\linewidth}{
@{}l
r
>{\raggedright\arraybackslash}p{0.19\linewidth}
>{\raggedright\arraybackslash}X
@{}}
\toprule
Benchmark
& Questions
& Answer format
& Reporting categories \\
\midrule

LRS-GRO
& 9,734
& Open-ended
& Global, Region, Object \\

MME-RealWorld-RS
& 3,738
& Multiple-choice
& Color, Count, Position \\

XLRS-Bench
& 3,080
& Multiple-choice
& Cnt, SC, OSR, OP, Plan, AR, CR, SR \\

\bottomrule
\end{tabularx}
\end{table}

\paragraph{Scoring and aggregation.}
For LRS-GRO, we follow the benchmark's answer-matching
rule: a prediction is correct if it matches the reference
after lowercasing or satisfies the WordNet
path-similarity threshold of 0.8.
For MME-RealWorld-RS and XLRS-Bench, predictions are
matched to the reference multiple-choice answers.
Category scores are computed over the questions
belonging to each category.
Within each benchmark, Avg.\ denotes the overall
question-level accuracy, equivalently the category
accuracies weighted by their sample counts.
Across benchmarks, Avg.\ or Mean denotes the
unweighted arithmetic mean of the three overall scores.
Improvements are reported in absolute percentage points.
Cross-benchmark averages are computed only for methods
with results on all three benchmarks.

\paragraph{Baseline provenance.}
We evaluate GazeEarth, its controlled variants, and the
corresponding Direct baselines using the selected backbone
checkpoints.
The main comparison tables distinguish results evaluated
in this work from externally reported scores:
$\dagger$ indicates that the experimental results for this model are taken from the original paper.
When a method combines results from different sources,
the provenance is indicated at the individual-score level.
A dash denotes an unreported result. All metrics not reported in the original paper were reproduced using the official code and parameter configurations.

\subsection{Hardware and Timing Protocol}
\label{app:hardware}

\paragraph{Hardware.}
Timing experiments are conducted on a single NVIDIA A800
GPU with a batch size of one question.
The model is loaded once and reused across samples.
For GazeEarth, region selection and answering are executed
sequentially using the same model instance.

\paragraph{Timing protocol.}
We measure per-question wall-clock latency after the source
images have been loaded and converted to RGB, ending when
the extracted answer is available.
The timed interval includes overview construction,
question-guided gaze when invoked, focused-view rendering,
backbone-specific preprocessing, and answer generation.
Model initialization, source-image I/O, benchmark scoring,
and prediction-file writing are excluded.
Each evaluated question contributes one timing observation,
with no dedicated warm-up phase or repeated timing trials.
We report the arithmetic mean of per-question latencies
within each benchmark and the unweighted mean of the
three benchmark-specific means across benchmarks.
Selection and answering times are also recorded separately;
the total additionally includes visual construction and
inter-stage processing.

\FloatBarrier
\section{Foveated Mapping Details}
\label{app:focus-details}

This section details spatial-request parsing, target-interval
placement, and rendering from the original source images
for the topology-preserving foveated mapping operator.
Topology preservation is a property of the continuous
coordinate mapping: the mapping is monotonic along each axis and has no folds, so it preserves connectivity and horizontal and vertical order.

\subsection{Parsing and Coordinate Rules}
\label{app:focus-parsing}

The indexed grid links each selected cell to a rectangle
in the original image through Eq.~\ref{eq:gaze-cell}.
Cells are numbered in row-major order, and identifiers
such as \texttt{I1:C07} specify the source image and cell.
After normalisation and deduplication, a valid
\texttt{selected} response contains one or two identifiers. The statuses \texttt{global\_only} and \texttt{brief\_incomplete} require an empty identifier list.
Invalid responses produce an empty request.

Cell boundaries are rounded to source-image pixels.
Each selected cell is padded by a fraction $\delta$ of its width and height on each side, with $\delta=0.25$ by default.
The padded boundaries are rounded and clipped to
the image extent, and overlapping rectangles from the same source are merged.
The enclosing rectangle of the resulting request
defines the shared focus neighborhood.
Its original-image coordinates are retained for rendering. 

A labeled bounding box supplied in the question
is mapped directly to source coordinates.
Normalized coordinates are scaled by $W$ and $H$; absolute coordinates are interpreted at the declared image resolution.
The mapped boundaries are rounded and clipped.
For multiple images, an explicit source identifier determines which image the box refers to.

The resolved box is expanded into a contextual square.
Let $d$ denote its longer side and $M=\max(W,H)$.
The default side length is $4d$, clamped between
$u=\max\{d,\operatorname{round}(0.04M)\}$ and
$v=\max\{u,\operatorname{round}(0.30M)\}$,
then capped by $\min(W,H)$.
The square is centred on the box and shifted
inside the image when necessary.
It supplies the spatial request directly;
otherwise, the indexed-overview selector is used.

\subsection{Placement Rule for the Target Interval}
\label{app:focus-placement}

The target interval preserves the source interval's
centre whenever the available peripheral space permits.
For the span $\ell^\star$ defined in
Eq.~\ref{eq:focus-span}, its endpoints are
\begin{equation}
\begin{aligned}
a^\star
&=
\min\!\left\{
L-\epsilon_+-\ell^\star,\,
\max\!\left\{
\epsilon_-,\,
\operatorname{round}\!\left(
\frac{a+b-\ell^\star}{2}
\right)
\right\}
\right\},\\
b^\star &= a^\star+\ell^\star.
\end{aligned}
\label{eq:focus-placement}
\end{equation}
This places the enlarged interval within the canvas
while reserving one output pixel for each nonempty
peripheral segment.
A focus interval touching a canvas edge remains
attached to that edge.
The construction is applied independently along
both axes, with empty outer segments omitted.

\subsection{Rendering and Multi-Source Handling}
\label{app:focus-rendering}

Rendering pairs the retained original-image boundaries
with the target boundaries specified by $\Phi_R$.
The projection onto the $h\times w$ canvas determines
the target spans and positions, while the original
image supplies the pixels.
When a projected boundary rounds to a canvas edge,
its corresponding source boundary is set to the
original-image edge.

The separable mapping is rendered in two passes.
First, the original image is divided at the focus boundaries
along the horizontal axis into up to three vertical strips.
Each strip is bilinearly resampled to its assigned
target width while retaining the original height $H$.
Their ordered concatenation forms an $H\times w$
intermediate image.
The second pass uses the original row boundaries
to resample the corresponding horizontal strips
to their target heights, producing the final
$h\times w$ view.
Both passes use Pillow's bilinear resampling with
explicit source bounds.

For multiple source images, requests are grouped
by source identifier and rendered independently.
An empty request returns an aspect-preserving
bilinear resize under the same long-side cap;
images already within the cap retain their dimensions.
The resulting views are passed together to the
answerer in source order.
Each view carries the same neutral, 36-pixel-high
source header used by the Overview condition.
The header is appended above the scene canvas,
whose pixel budget is $B=hw$.

\FloatBarrier
\section{Pilot Study Details}
\label{app:pilot}

This section details the two pilot probes:
spatial-request alignment and evidence presentation.
We additionally validate target localization against
dataset-provided bounding boxes on LRS-GRO.

\subsection{Setup}
\label{app:polit-setup}

For the localisation probe, we overlay a $4\times4$
indexed grid on a full-scene overview with a
1536-pixel long-side cap.
The frozen MLLM receives the overview and question
and returns cell identifiers without answering.
We compare its requests with those of SigLIP2,
RemoteCLIP, Random Focus, and Question-only MLLM.
The latter receives the question and grid definition
without an image.

For the spatial-cue alignment analysis, the reference
region is constructed from explicit spatial expressions
in the question.
Ordered rules map corners, quadrants, directional bands,
and central regions to reference cells; the first
matching rule determines the reference region.
In this analysis, each observer is evaluated on questions
with both a matched reference and an extractable
selected cell.
Questions without a matched reference and requests without
an extractable selected cell are excluded from these statistics.
The resulting evaluation measures agreement with the
question-specified region.

For a nonempty selected-cell set and reference-cell set,
$d$ is the minimum Manhattan distance between their cells.
Thus, $d=0$ means that at least one selected cell belongs
to the reference region.
We report the alignment rate $P(d=0)$, mean grid distance,
and tolerance curves $P(d\leq r)$.
Table~\ref{tab:gaze-grounding} reports per-benchmark alignment
and its unweighted three-benchmark mean.
Figure~\ref{fig:gaze-tolerance} reports the tolerance curves
and eligible sample counts for the plotted observers.

For the evidence-presentation probe, we compare
\textit{Overview}, \textit{Gaze + Crop}, and
\textit{Think with Image} using the same answerer across
XLRS-Bench, LRS-GRO, and MME-RealWorld-RS.
\textit{Overview} uses a uniformly resized full-scene view.
\textit{Gaze + Crop} presents the question-selected crop
alone, whereas \textit{Think with Image} presents that crop
together with a separate overview without explicit spatial
linkage.

\subsection{Spatial-Request Alignment and Target Localization}
\label{app:tolerance}

Table~\ref{tab:gaze-grounding} expands the aggregate alignment
comparison in Table~\ref{tab:pilot-grounding}.
The image-conditioned MLLM achieves alignment rates of
73.89\%, 77.60\%, and 81.74\% on XLRS-Bench,
LRS-GRO, and MME-RealWorld-RS, respectively.
Its three-benchmark mean is 77.74\%, compared with
43.09\% for SigLIP2, 37.20\% for RemoteCLIP,
and 16.35\% for Random Focus.

Question-only MLLM achieves 41.50\%, 44.56\%, and
42.40\% on the three benchmarks, with a mean of 42.82\%.
The reported alignment rates of the image-conditioned
MLLM are higher by 32.39, 33.04, and 39.34 percentage
points, respectively, yielding a mean difference of
34.92 points.
Image-conditioned selection therefore exhibits stronger
agreement with the question-specified regions in this probe.

\begin{table*}[htbp]
\centering
\caption{Spatial-request alignment with question-derived reference
cells across benchmarks. We report $P(d=0)$ on the eligible
examples for each observer. Avg. is the unweighted mean of the
three benchmark scores; higher is better.}
\label{tab:gaze-grounding}

\vspace{0.1cm} 

\scriptsize
\renewcommand{\arraystretch}{0.85}
\setlength{\tabcolsep}{1.5pt} 
\setlength{\aboverulesep}{0.25ex}
\setlength{\belowrulesep}{0.25ex}

\begin{tabularx}{\textwidth}{@{} >{\raggedright\arraybackslash}X *{4}{Y} @{}}
\toprule
\textbf{Observer} 
& \textbf{XLRS-Bench (\%)} $\uparrow$ 
& \textbf{LRS-GRO (\%)} $\uparrow$ 
& \textbf{MME-RealWorld-RS (\%)} $\uparrow$ 
& \textbf{Avg. (\%)} $\uparrow$ \\
\midrule
Random Focus       & 14.65 & 21.83 & 12.56 & 16.35 \\
Question-only MLLM
& \secondcell{41.50} & 44.56
& \secondcell{42.40} & 42.82 \\
SigLIP2
& 39.49 & \secondcell{54.18}
& 35.59 & \secondcell{43.09} \\
RemoteCLIP         & 33.44 & 45.82 & 32.33 & 37.20 \\
\midrule
\textbf{Frozen MLLM} & \bestcell{73.89} & \bestcell{77.60} & \bestcell{81.74} & \bestcell{77.74} \\
\bottomrule
\end{tabularx}
\end{table*}

\begin{figure*}[htbp]
  \centerline{\includegraphics[width=0.83\linewidth]{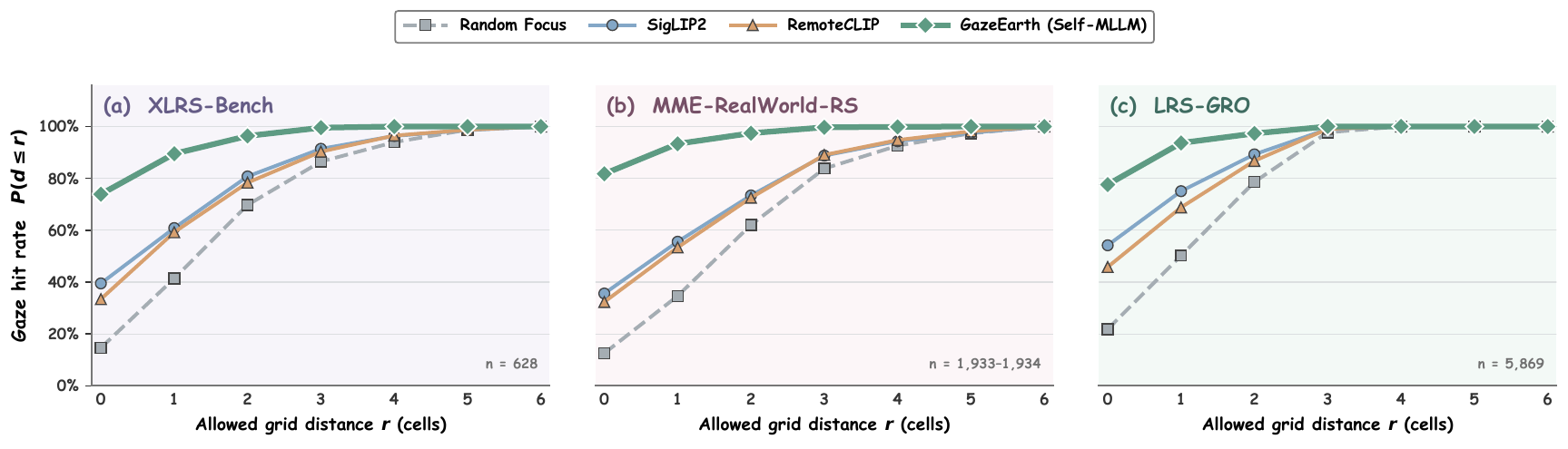}}
  \vspace{-0.1cm}
  \caption{Spatial-request alignment within an allowed grid
  distance $r$ on three benchmarks.
  Here, $d$ is the minimum Manhattan distance between
  selected cells and question-derived reference cells.
  Curves show $P(d\leq r)$ on the eligible examples for
  each observer; sample counts refer to these examples.}
  \label{fig:gaze-tolerance}
\end{figure*}

Figure~\ref{fig:gaze-tolerance} shows how alignment changes
with the allowed grid distance.
At $r=1$, the MLLM's hit rates reach 89.49\% on
XLRS-Bench, 93.65\% on LRS-GRO, and 93.33\% on
MME-RealWorld-RS; at $r=2$, all three exceed 96\%.
The increase from $r=0$ to $r=1$ accounts for more
than half of its off-target requests on each benchmark.
Its selected cells are thus concentrated within and
immediately around the question-specified regions.

\subsection{Localization against independent target annotations.}
To complement the spatial-cue analysis, we evaluate
the recorded selections against dataset-provided
target boxes on LRS-GRO.
The evaluation subset contains 537 object-attribute
questions from 159 images without explicit location
cues or coordinate references.
It covers object category, color, material, shape,
and state or activity.
Questions are selected by their content and annotation
validity, rather than model predictions or correctness.
We verify each question--target correspondence and
the mapping of annotation coordinates to the original
image.
The aspect-preserving, 1024-pixel long-side annotation
coordinates are converted to source-image coordinates
by multiplying both axes by $\max(W,H)/1024$. 
Reference boxes are used only for evaluation.
The external observers retain GazeEarth's request
status and per-question cell budget:
493 questions have one selected cell, 29 have two,
and 15 have no selected cell. Center hit measures whether the selected cells contain
the annotated target center.
Box coverage measures the fraction of the target-box
area covered by the union of selected cells,
averaged over questions.
Both metrics use the original $4\times4$ grid cells
before padding or region merging.
All 537 questions remain in the denominator,
with empty or invalid requests receiving zero.

\begin{table}[t]
\centering

\caption{Target localization and oracle-gaze answering on
537 LRS-GRO questions from 159 images without explicit
location cues.
The three observers share the same per-question request
status and cell budget for localization.
Higher is better for all percentage metrics.}
\label{tab:independent-target-localization}
\scriptsize
\renewcommand{\arraystretch}{0.95}
\setlength{\tabcolsep}{5pt}

\begin{tabular}{@{}lccc@{}}
\toprule
Observer
& Center hit (\%) $\uparrow$
& Box coverage (\%) $\uparrow$
& Answer accuracy (\%) $\uparrow$ \\
\midrule
SigLIP2
& \secondcell{52.33} & \secondcell{47.56} & 64.43\\
RemoteCLIP
& 37.99 & 34.33 & 62.94\\
Self-MLLM
& \bestcell{64.99} & \bestcell{55.65} & \secondcell{66.29} \\
Oracle Gaze
& -- & -- & \bestcell{70.20} \\
\bottomrule
\end{tabular}
\end{table}

Table~\ref{tab:independent-target-localization}
shows that GazeEarth achieves a center-hit rate
of 64.99\%, exceeding SigLIP2 and RemoteCLIP
by 12.66 and 27.00 percentage points, respectively.
Its mean target-box coverage reaches 55.65\%,
compared with 47.56\% and 34.33\%.
These results extend the spatial-cue analysis
to independently annotated targets:
the answerer's own requests also localize
question-relevant objects when their positions
are not explicitly stated. We further evaluate answering with oracle gaze on the
same annotated subset.
Dataset-provided target boxes replace predicted regions
in constructing the full-scene focused view, which is
then passed to the frozen Qwen3-VL-8B answerer using
the CoT2 prompt.
Oracle-bbox Gaze achieves 70.20\% answering accuracy, complementing the localization results with
an assessment of answering under annotation-guided focus.

\subsection{Aligned versus Off-target Requests}
\label{app:aligned}

For the question-derived reference-cell analysis,
Acc.@Aligned measures answer accuracy on requests
with $d=0$, and Acc.@Off-target measures accuracy
on requests with $d>0$, separately for each observer.
Spatial Acc. is the answer accuracy over that observer's
evaluated spatial subset.
Table~\ref{tab:pilot-grounding} reports the unweighted
three-benchmark mean of each metric.
Tables~\ref{tab:gaze-grounding-XLRS-Bench},
\ref{tab:gaze-grounding-mme-realworld-rs}, and
\ref{tab:gaze-grounding-lrs-gro} provide the corresponding
per-benchmark results.

For the MLLM's own requests, aligned examples achieve
50.49\% benchmark-averaged answer accuracy, compared with
42.00\% for off-target examples.
The same association holds within each benchmark:
44.40\% versus 37.20\% on XLRS-Bench,
57.65\% versus 50.00\% on LRS-GRO, and
49.43\% versus 38.81\% on MME-RealWorld-RS.
Aligned requests are more often answered correctly,
but alignment alone does not ensure a correct answer.
These conditional accuracies reflect the combined outcome
of evidence perception and answer reasoning.
They motivate examining how the selected region is
subsequently presented to the answerer.

\begin{table*}[htbp]
\centering

\caption{Spatial-request alignment and conditional answering
accuracy on XLRS-Bench.}
\label{tab:gaze-grounding-XLRS-Bench}

\vspace{-0.15cm}

\scriptsize
\renewcommand{\arraystretch}{0.85}
\setlength{\tabcolsep}{1.5pt}
\setlength{\aboverulesep}{0.25ex}
\setlength{\belowrulesep}{0.25ex}

\begin{tabularx}{\textwidth}{
  @{}
  >{\itshape\raggedright\arraybackslash}p{0.28\textwidth}
  *{5}{Y}
  @{}
}
\toprule

\raisebox{1.0\baselineskip}[0pt][0pt]{%
  \normalfont\bfseries Observer%
}
& \raisebox{0.5\baselineskip}[0pt][0pt]{%
    \shortstack[c]{%
      \textbf{Align.}\\
      \textbf{(\%)} $\uparrow$%
    }%
  }
& \raisebox{0.5\baselineskip}[0pt][0pt]{%
    \shortstack[c]{%
      \textbf{Grid}\\
      \textbf{Dist.} $\downarrow$%
    }%
  }
& \shortstack[c]{%
    \textbf{Acc.@}\\
    \textbf{Aligned}\\
    \textbf{(\%)} $\uparrow$%
  }
& \shortstack[c]{%
    \textbf{Acc.@}\\
    \textbf{Off-target}\\
    \textbf{(\%)} $\uparrow$%
  }
& \shortstack[c]{%
    \textbf{Spatial}\\
    \textbf{Acc.}\\
    \textbf{(\%)} $\uparrow$%
  }
\\

\midrule

Random Focus
& 14.65
& 1.951
& 39.13
& \bestcell{38.99}
& \secondcell{39.01}
\\

SigLIP2
& \secondcell{39.49}
& \secondcell{1.326}
& 40.32
& \secondcell{37.37}
& 38.54
\\

RemoteCLIP
& 33.44
& 1.435
& \secondcell{41.90}
& 35.89
& 37.90
\\

\midrule

\shortstack[l]{\textbf{Frozen MLLM}}
& \bestcell{73.89}
& \bestcell{0.408}
& \bestcell{44.40}
& 37.20
& \bestcell{42.52}
\\

\bottomrule
\end{tabularx}

\end{table*}

\begin{table*}[htbp]
\centering

\caption{Spatial-request alignment and conditional answering
accuracy on MME-RealWorld-RS.}
\label{tab:gaze-grounding-mme-realworld-rs}

\scriptsize
\renewcommand{\arraystretch}{0.92}
\setlength{\tabcolsep}{1.5pt}
\setlength{\aboverulesep}{0.25ex}
\setlength{\belowrulesep}{0.25ex}

\begin{tabularx}{\textwidth}{
  @{}
  >{\itshape\raggedright\arraybackslash}p{0.28\textwidth}
  *{5}{Y}
  @{}
}
\toprule

\raisebox{1.0\baselineskip}[0pt][0pt]{%
  \normalfont\bfseries Observer%
}
& \raisebox{0.5\baselineskip}[0pt][0pt]{%
    \shortstack[c]{%
      \textbf{Align.}\\
      \textbf{(\%)} $\uparrow$%
    }%
  }
& \raisebox{0.5\baselineskip}[0pt][0pt]{%
    \shortstack[c]{%
      \textbf{Grid}\\
      \textbf{Dist.} $\downarrow$%
    }%
  }
& \shortstack[c]{%
    \textbf{Acc.@}\\
    \textbf{Aligned}\\
    \textbf{(\%)} $\uparrow$%
  }
& \shortstack[c]{%
    \textbf{Acc.@}\\
    \textbf{Off-target}\\
    \textbf{(\%)} $\uparrow$%
  }
& \shortstack[c]{%
    \textbf{Spatial}\\
    \textbf{Acc.}\\
    \textbf{(\%)} $\uparrow$%
  }
\\

\midrule

Random Focus
& 12.56
& 2.170
& 41.98
& \secondcell{36.72}
& 37.38
\\

SigLIP2
& \secondcell{35.59}
& \secondcell{1.548}
& \secondcell{49.27}
& 36.22
& \secondcell{40.87}
\\

RemoteCLIP
& 32.33
& 1.599
& 48.64
& 36.16
& 40.20
\\

\midrule

\shortstack[l]{\textbf{Frozen MLLM}}
& \bestcell{81.74}
& \bestcell{0.280}
& \bestcell{49.43}
& \bestcell{38.81}
& \bestcell{47.49}
\\

\bottomrule
\end{tabularx}

\end{table*}

\subsection{Evidence Presentation}
\label{app:pilot-presentation}

Table~\ref{tab:pilot-presentation} compares answering
from the full-scene overview, the selected crop alone,
and the crop accompanied by a separate overview.
This probe examines how the answerer uses
question-selected evidence under different presentations.

The effects differ across benchmarks.
On XLRS-Bench, presenting the selected crop raises accuracy
from 45.13\% to 50.55\%, and adding the overview increases
it to 51.43\%.
On MME-RealWorld-RS, isolated cropping reduces accuracy
from 49.01\% to 39.81\%; adding the overview recovers
part of this decrease, reaching 42.24\%.
On LRS-GRO, the corresponding scores are 59.99\%,
58.93\%, and 60.54\%.

Across the three benchmarks, \textit{Overview},
\textit{Gaze + Crop}, and \textit{Unlinked Global + Crop}
achieve mean accuracies of 51.38\%, 49.76\%, and
51.40\%, respectively.
Presenting the requested region as a crop therefore helps on XLRS-Bench but hurts on MME-RealWorld-RS, and does not improve the three-benchmark mean.
With the selection procedure held fixed, answering can become better or worse depending on how the requested region is rendered.
Together, the two probes separate spatial-request alignment from the usefulness of the resulting observation, and motivate treating the conversion of a request into the next observation as a design problem in its own right.
GazeEarth is one such conversion: it enlarges the selected neighbourhood within a continuous full-scene view.

\begin{table*}[htbp]
\centering

\caption{Spatial-request alignment and conditional answering
accuracy on LRS-GRO.}
\label{tab:gaze-grounding-lrs-gro}

\scriptsize
\renewcommand{\arraystretch}{0.92}
\setlength{\tabcolsep}{1.5pt}
\setlength{\aboverulesep}{0.25ex}
\setlength{\belowrulesep}{0.25ex}

\begin{tabularx}{\textwidth}{
  @{}
  >{\itshape\raggedright\arraybackslash}p{0.28\textwidth}
  *{5}{Y}
  @{}
}
\toprule

\raisebox{1.0\baselineskip}[0pt][0pt]{%
  \normalfont\bfseries Observer%
}
& \raisebox{0.5\baselineskip}[0pt][0pt]{%
    \shortstack[c]{%
      \textbf{Align.}\\
      \textbf{(\%)} $\uparrow$%
    }%
  }
& \raisebox{0.5\baselineskip}[0pt][0pt]{%
    \shortstack[c]{%
      \textbf{Grid}\\
      \textbf{Dist.} $\downarrow$%
    }%
  }
& \shortstack[c]{%
    \textbf{Acc.@}\\
    \textbf{Aligned}\\
    \textbf{(\%)} $\uparrow$%
  }
& \shortstack[c]{%
    \textbf{Acc.@}\\
    \textbf{Off-target}\\
    \textbf{(\%)} $\uparrow$%
  }
& \shortstack[c]{%
    \textbf{Spatial}\\
    \textbf{Acc.}\\
    \textbf{(\%)} $\uparrow$%
  }
\\

\midrule

Random Focus
& 21.83
& 1.516
& 53.63
& \bestcell{53.07}
& 53.19
\\

SigLIP2
& \secondcell{54.18}
& \secondcell{0.825}
& \secondcell{57.26}
& 51.39
& \secondcell{54.57}
\\

RemoteCLIP
& 45.82
& 0.996
& 55.86
& \secondcell{52.36}
& 53.96
\\

\midrule

\shortstack[l]{\textbf{Frozen MLLM}}
& \bestcell{77.60}
& \bestcell{0.314}
& \bestcell{57.65}
& 50.00
& \bestcell{55.93}
\\

\bottomrule
\end{tabularx}

\end{table*}

\FloatBarrier
\section{Extended Quantitative Results}
\label{app:extended}

This section expands the main quantitative results
through category-level comparisons, observer substitution,
gaze-depth analysis, and the comparison between
random focus and the full-scene overview.

\begin{figure*}[htbp]
  \centerline{\includegraphics[scale=0.75]{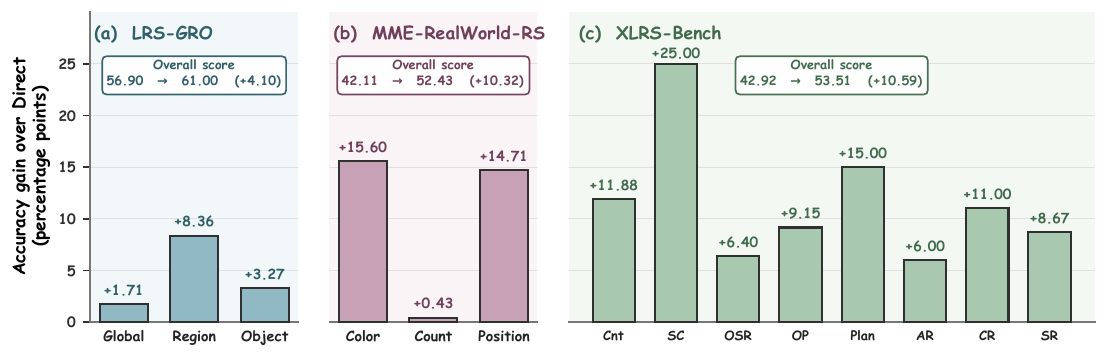}}
  \vspace{-0.1cm}
  \caption{Category-wise accuracy gains of GazeEarth over
    Direct using the same Qwen3-VL-8B backbone on
    (a) LRS-GRO, (b) MME-RealWorld-RS, and (c) XLRS-Bench.
    Bars show absolute improvements in percentage points.
    Insets report the overall benchmark scores
    (Direct $\rightarrow$ GazeEarth) and their differences.}
  \label{fig:category-gain}
\end{figure*}

\subsection{Category-Level Breakdown}
\label{app:category}

Figure~\ref{fig:category-gain} reports category-wise gains
over Direct using Qwen3-VL-8B.
GazeEarth improves all reported categories across the three
benchmarks. On XLRS-Bench, the largest gains occur in
Scene Classification and Planning, at 25.00 and 15.00
percentage points, followed by Counting and Complex
Reasoning at 11.88 and 11.00 points.
Counting improvements differ across benchmarks:
MME-RealWorld-RS increases from 21.35\% to 21.78\%,
whereas XLRS-Bench increases from 32.50\% to 44.38\%.

\begin{table*}[htbp]
\centering

\caption{Category-level controlled analysis on XLRS-Bench.}
\label{tab:category-level-controlled-analysis}

\vspace{-0.15cm}

\scriptsize
\renewcommand{\arraystretch}{0.92}
\setlength{\tabcolsep}{1.5pt}
\setlength{\aboverulesep}{0.25ex}
\setlength{\belowrulesep}{0.25ex}

\begin{tabularx}{\textwidth}{
  @{}
  lccccc
  *{9}{Y}
  @{}
}
\toprule

\multirow[c]{2}{*}{\textbf{Variant}}
& \multirow[c]{2}{*}{\textbf{Gaze}}
& \multirow[c]{2}{*}{%
    \shortstack[c]{%
      \textbf{Q-guided}\\
      \textbf{gaze}%
    }%
  }
& \multirow[c]{2}{*}{\textbf{Detail}}
& \multirow[c]{2}{*}{%
    \shortstack[c]{%
      \textbf{Full}\\
      \textbf{scene}%
    }%
  }
& \multirow[c]{2}{*}{\textbf{Binding}}
& \multicolumn{9}{c}{%
    \textbf{XLRS-Bench (\%)} $\uparrow$%
  }
\\

\cmidrule(lr){7-15}

& & & & & &
\textbf{Cnt}
& \textbf{SC}
& \textbf{OSR}
& \textbf{OP}
& \textbf{Plan}
& \textbf{AR}
& \textbf{CR}
& \textbf{SR}
& \textbf{Avg.}
\\

\midrule

\multicolumn{15}{@{}l}{%
  \textit{\textbf{Visual Evidence Composition}}%
}
\\[-0.25ex]

\textit{Direct}
& 0
& \tfno
& \tfno
& \tfyes
& --
& 32.50
& 41.67
& \secondcell{31.40}
& 43.92
& 34.00
& 70.00
& 63.50
& \secondcell{46.33}
& 42.92
\\
\textit{Overview}
& 0
& \tfno
& \tfno
& \tfyes
& --
& 35.00
& 60.67
& 28.80
& 44.52
& 42.00
& 71.00
& 64.50
& 45.00
& 45.13
\\
\textit{Random Focus}
& 1
& \tfno
& \tfyes
& \tfyes
& \tfyes implicit
& 35.00
& 61.33
& 30.60
& 44.10
& 39.00
& \bestcell{78.00}
& 65.50
& 43.33
& 45.42
\\
\textit{Gaze + Crop}
& 1
& \tfyes
& \tfyes
& \tfno
& \tfno
& \secondcell{41.88}
& \secondcell{66.33}
& 29.20
& 52.29
& 41.00
& 72.00
& \secondcell{69.50}
& 41.67
& 50.55
\\
\textit{Think with Image}
& 1
& \tfyes
& \tfyes
& \tfyes
& \tfno
& 36.88
& 63.67
& 29.20
& \secondcell{55.66}
& 40.00
& 71.00
& 64.50
& 40.00
& 51.43
\\
\textit{Linked Global + Crop}
& 1
& \tfyes
& \tfyes
& \tfyes
& \tfyes explicit
& 33.75
& 62.33
& 30.00
& \bestcell{57.83}
& 39.00
& 73.00
& 66.00
& 40.00
& \secondcell{52.56}
\\
\textit{AttWarp}
& 1
& \tfyes
& \tfyes
& \tfyes
& \tfyes implicit
& 33.75 & 58.67 & 31.00 & 44.94 & 39.00 & 74.00 & 62.50 & 46.67 & 45.36
\\
\midrule
\multicolumn{15}{@{}l}{%
\textit{\textbf{Gaze Depth and Iterative Baselines}}%
}
\\[-0.25ex]

\textit{Two-Gaze}
& 2
& \tfyes
& \tfyes
& \tfyes
& \tfyes implicit
& 38.12
& 64.33
& 31.20
& 44.82
& 44.00
& 72.00
& 65.50
& 40.00
& 46.27
\\
\textit{Three-Gaze}
& 3
& \tfyes
& \tfyes
& \tfyes
& \tfyes implicit
& 40.62
& 65.00
& 31.00
& 44.16
& \secondcell{45.00}
& 74.00
& 68.00
& 41.67
& 46.36
\\
\textit{\textbf{GazeEarth}}
& 1
& \tfyes
& \tfyes
& \tfyes
& \tfyes implicit
& \bestcell{44.38}
& \bestcell{66.67}
& \bestcell{37.80}
& 53.07
& \bestcell{49.00}
& \secondcell{76.00}
& \bestcell{74.50}
& \bestcell{55.00}
& \bestcell{53.51} \\

\bottomrule
\end{tabularx}

\end{table*}

Tables~\ref{tab:category-level-controlled-analysis},
\ref{tab:category-level-controlled-analysis-lrs-gro},
and~\ref{tab:category-level-controlled-analysis-mme-realworld-rs}
provide category-level comparisons of the
evidence-presentation variants.
\textit{Overview} presents the uniformly resized full
scene, while \textit{Gaze + Crop} presents only the
question-selected region.
\textit{Think with Image} (\textit{Unlinked Global + Crop})
provides both views without explicit spatial linkage;
\textit{Linked Global + Crop} additionally identifies
the correspondence between the crop and its source
region in the overview.
We additionally evaluate AttWarp with Qwen3-VL-8B
as the backbone.
AttWarp guides image warping through cross-modal attention,
whereas GazeEarth uses an explicit textual spatial request
to guide deterministic piecewise-linear full-scene resampling.
GazeEarth instead uses the model's explicit spatial
request to guide piecewise-linear full-scene warping.
On XLRS-Bench, GazeEarth achieves the highest overall
accuracy and leads on Counting, Scene Classification,
OSR, Planning, Complex Reasoning, and SR among the
listed variants. Linked Global + Crop leads on Object
Properties, while Random Focus leads on Anomaly Reasoning.
On LRS-GRO, GazeEarth achieves the highest Global,
Object, and overall scores, while Linked-Global + Crop
leads on Region questions.
The comparison with Overview separates these gains
from those obtained by changing the full-scene input.
On LRS-GRO, Region accuracy increases from 42.03\%
with Direct to 50.22\% with Overview and 50.39\%
with GazeEarth.
Object accuracy increases from 54.77\% to 56.22\%
and 58.04\%, respectively.
Thus, Region shows the largest gain over Direct,
whereas Object shows the largest additional gain
over Overview.
On MME-RealWorld-RS, Gaze + Crop improves Color
accuracy over Overview from 55.54\% to 62.47\%,
but reduces Position accuracy from 72.39\% to 34.45\%.
The effect of evidence presentation, therefore, varies
across question categories. GazeEarth focuses on details while preserving global topology, thereby offering an overall advantage.

\begin{table*}[htbp]
\centering

\caption{Category-level controlled analysis on LRS-GRO.}
\label{tab:category-level-controlled-analysis-lrs-gro}

\vspace{-0.15cm}

\scriptsize
\renewcommand{\arraystretch}{0.92}
\setlength{\tabcolsep}{1.5pt}
\setlength{\aboverulesep}{0.25ex}
\setlength{\belowrulesep}{0.25ex}

\begin{tabularx}{\textwidth}{
  @{}
  lccccc
  *{4}{Y}
  @{}
}
\toprule

\multirow[c]{2}{*}{\textbf{Variant}}
& \multirow[c]{2}{*}{\textbf{Gaze}}
& \multirow[c]{2}{*}{%
    \shortstack[c]{%
      \textbf{Q-guided}\\
      \textbf{gaze}%
    }%
  }
& \multirow[c]{2}{*}{\textbf{Detail}}
& \multirow[c]{2}{*}{%
    \shortstack[c]{%
      \textbf{Full}\\
      \textbf{scene}%
    }%
  }
& \multirow[c]{2}{*}{\textbf{Binding}}
& \multicolumn{4}{c}{%
    \textbf{LRS-GRO (\%)} $\uparrow$%
  }
\\

\cmidrule(lr){7-10}

& & & & & &
\textbf{Global}
& \textbf{Region}
& \textbf{Object}
& \textbf{Avg.}
\\

\midrule

\multicolumn{10}{@{}l}{%
  \textit{\textbf{Visual Evidence Composition}}%
}
\\[-0.25ex]

\textit{Direct}
& 0
& \tfno
& \tfno
& \tfyes
& --
& 76.25
& 42.03
& 54.77
& 56.90
\\

\textit{Overview}
& 0
& \tfno
& \tfno
& \tfyes
& --
& \secondcell{77.88}
& 50.22
& 56.22
& 59.99
\\

\textit{Random Focus}
& 1
& \tfno
& \tfyes
& \tfyes
& \tfyes implicit
& 77.84
& 48.78
& 54.95
& 58.97
\\

\textit{Gaze + Crop}
& 1
& \tfyes
& \tfyes
& \tfno
& \tfno
& 74.66
& 47.39
& 56.95
& 58.93
\\

\textit{Think with Image}
& 1
& \tfyes
& \tfyes
& \tfyes 
& \tfno
& 76.80
& \secondcell{51.09}
& \secondcell{57.38}
& \secondcell{60.54}
\\

\textit{Linked Global + Crop}
& 1
& \tfyes
& \tfyes
& \tfyes 
& \tfyes explicit
& 75.34
& \bestcell{51.52}
& 57.30
& 60.25
\\

\textit{AttWarp}
& 1
& \tfyes
& \tfyes
& \tfyes 
& \tfyes implicit
& 77.75 & 49.00 & 53.39 & 58.18 \\

\midrule

\multicolumn{10}{@{}l}{%
  \textit{\textbf{Gaze Depth and Iterative Baselines}}%
}
\\[-0.25ex]

\textit{Two-Gaze}
& 2
& \tfyes
& \tfyes
& \tfyes
& \tfyes implicit
& 76.93
& 50.00
& 57.01
& 60.12
\\

\textit{Three-Gaze}
& 3
& \tfyes
& \tfyes
& \tfyes
& \tfyes implicit
& 76.68
& 49.17
& 57.34
& 60.04
\\

\textit{\textbf{GazeEarth}}
& \textbf{1}
& \tfyes
& \tfyes
& \tfyes
& \tfyes implicit
& \bestcell{77.96}
& 50.39
& \bestcell{58.04}
& \bestcell{61.00}
\\

\bottomrule
\end{tabularx}

\vspace{-0.3cm}
\end{table*}

\begin{table*}[htbp]
\centering

\caption{Category-level controlled analysis on MME-RealWorld-RS.}
\label{tab:category-level-controlled-analysis-mme-realworld-rs}

\vspace{-0.15cm}

\scriptsize
\renewcommand{\arraystretch}{0.92}
\setlength{\tabcolsep}{1.5pt}
\setlength{\aboverulesep}{0.25ex}
\setlength{\belowrulesep}{0.25ex}

\begin{tabularx}{\textwidth}{
@{}
lccccc
*{4}{Y}
@{}
}
\toprule

\multirow[c]{2}{*}{\textbf{Variant}}
& \multirow[c]{2}{*}{\textbf{Gaze}}
& \multirow[c]{2}{*}{%
\shortstack[c]{%
\textbf{Q-guided}\\
\textbf{gaze}%
}%
}
& \multirow[c]{2}{*}{\textbf{Detail}}
& \multirow[c]{2}{*}{%
\shortstack[c]{%
\textbf{Full}\\
\textbf{scene}%
}%
}
& \multirow[c]{2}{*}{\textbf{Binding}}
& \multicolumn{4}{c}{%
\textbf{MME-RealWorld-RS (\%)} $\uparrow$%
}
\\

\cmidrule(lr){7-10}

& & & & & &
\textbf{Color}
& \textbf{Count}
& \textbf{Position}
& \textbf{Avg.}
\\

\midrule

\multicolumn{10}{@{}l}{%
\textit{\textbf{Visual Evidence Composition}}%
}
\\[-0.25ex]

\textit{Direct}
& 0
& \tfno
& \tfno
& \tfyes
& --
& 48.07
& 21.35
& 56.41
& 42.11
\\

\textit{Overview}
& 0
& \tfno
& \tfno
& \tfyes
& --
& 55.54
& 18.35
& \bestcell{72.39}
& 49.01
\\

\textit{Random Focus}
& 1
& \tfno
& \tfyes
& \tfyes
& \tfyes implicit
& 52.43
& 16.80
& 66.98
& 45.64
\\

\textit{Gaze + Crop}
& 1
& \tfyes
& \tfyes
& \tfno
& \tfno
& 62.47
& 22.10
& 34.45
& 39.81
\\

\textit{Unlinked Global + Crop}
& 1
& \tfyes
& \tfyes
& \tfyes
& \tfno
& 63.35
& 20.64
& 42.24
& 42.24
\\

\textit{Linked Global + Crop}
& 1
& \tfyes
& \tfyes
& \tfyes
& \tfyes explicit
& 62.79
& 19.74
& 47.10
& 43.39
\\

\textit{AttWarp}
& 1
& \tfyes
& \tfyes
& \tfyes
& \tfyes implicit
& 54.58 & 19.00 & 62.13 & 45.45
\\

\midrule

\multicolumn{10}{@{}l}{%
\textit{\textbf{Gaze Depth and Iterative Baselines}}%
}
\\[-0.25ex]

\textit{\textbf{GazeEarth}}
& \textbf{1}
& \tfyes
& \tfyes
& \tfyes
& \tfyes implicit
& 63.67
& 21.78
& \secondcell{71.12}
& \bestcell{52.43}
\\

\textit{Two-Gaze}
& 2
& \tfyes
& \tfyes
& \tfyes
& \tfyes implicit
& \bestcell{65.74}
& \secondcell{23.98}
& 66.59
& \secondcell{52.33}
\\

\textit{Three-Gaze}
& 3
& \tfyes
& \tfyes
& \tfyes
& \tfyes implicit
& \secondcell{64.78}
& \bestcell{24.71}
& 65.16
& 51.77
\\

\bottomrule
\end{tabularx}

\end{table*}

\subsection{Observer Substitution per Benchmark}
\label{app:observer}

We vary the observer while keeping Qwen3-VL-8B as the
answerer and using the same focused-view construction
and answering pipeline.
Table~\ref{tab:observer-substitution-mme-realworld-rs} reports the
category-level results on MME-RealWorld-RS.
Self-directed gaze achieves the highest accuracy in
all three categories.
Compared with the strongest external observer in each
category, GazeEarth improves Color by 7.57 percentage
points, Count by 2.45 points, and Position by 4.14 points.
Overall accuracy reaches 52.43\%, exceeding SigLIP2
by 4.94 points.

On LRS-GRO
(Table~\ref{tab:observer-substitution-lrs-gro}),
GazeEarth achieves the highest Global and Object
accuracies, at 77.96\% and 58.04\%, respectively.
Its Object score exceeds RemoteCLIP by 1.80 points
and SigLIP2 by 2.21 points.
RemoteCLIP achieves the highest Region accuracy
at 50.78\%, followed by GazeEarth at 50.39\%.
Across the three categories, self-directed gaze yields
the highest overall score of 61.00\%, compared with
60.10\% for RemoteCLIP and 59.54\% for SigLIP2.
On XLRS-Bench (Table~\ref{tab:observer-substitution}), GazeEarth reaches 53.51\%,
exceeding SigLIP2 and RemoteCLIP by 7.41 and 7.57
percentage points, respectively.

\begin{table*}[htbp]
\centering

\caption{Observer substitution on MME-RealWorld-RS with
Qwen3-VL-8B as the fixed answerer. Best and second-best marks are assigned per column over the whole table; tied values receive the same mark.}
\label{tab:observer-substitution-mme-realworld-rs}

\vspace{-0.15cm}

\scriptsize
\renewcommand{\arraystretch}{0.92}
\setlength{\tabcolsep}{1.5pt}
\setlength{\aboverulesep}{0.25ex}
\setlength{\belowrulesep}{0.25ex}

\begin{tabularx}{\textwidth}{
  @{}
  >{\itshape\raggedright\arraybackslash}p{0.28\textwidth}
  *{5}{Y}
  @{}
}
\toprule

\raisebox{0.5\baselineskip}[0pt][0pt]{%
  \normalfont\bfseries Observer%
}
& \raisebox{0.5\baselineskip}[0pt][0pt]{%
    \textbf{Color}%
  }
& \raisebox{0.5\baselineskip}[0pt][0pt]{%
    \textbf{Count}%
  }
& \raisebox{0.5\baselineskip}[0pt][0pt]{%
    \textbf{Position}%
  }
& \raisebox{0.5\baselineskip}[0pt][0pt]{%
    \textbf{Avg.}%
  }
& \shortstack[c]{%
    \textbf{$\Delta$ vs.}\\
    \textbf{Random}%
  }
\\

\midrule

Random
& 52.43
& 16.80
& \secondcell{66.98}
& 45.64
& 0.00
\\

SigLIP2
& \secondcell{56.10}
& \secondcell{19.33}
& 66.35
& \secondcell{47.49}
& \secondcell{+1.85}
\\

RemoteCLIP
& 55.46
& 18.11
& \secondcell{66.98}
& 47.08
& +1.44
\\

\midrule

\shortstack[l]{%
  \textbf{GazeEarth}%
}
& \bestcell{63.67}
& \bestcell{21.78}
& \bestcell{71.12}
& \bestcell{52.43}
& \bestcell{+6.79}
\\

\bottomrule
\end{tabularx}

\end{table*}

\begin{table*}[htbp]
\centering

\caption{Observer substitution on LRS-GRO with Qwen3-VL-8B as
the fixed answerer.}
\label{tab:observer-substitution-lrs-gro}

\vspace{-0.15cm}

\scriptsize
\renewcommand{\arraystretch}{0.92}
\setlength{\tabcolsep}{1.5pt}
\setlength{\aboverulesep}{0.25ex}
\setlength{\belowrulesep}{0.25ex}

\begin{tabularx}{\textwidth}{
  @{}
  >{\itshape\raggedright\arraybackslash}p{0.28\textwidth}
  *{5}{Y}
  @{}
}
\toprule

\raisebox{0.5\baselineskip}[0pt][0pt]{%
  \normalfont\bfseries Observer%
}
& \raisebox{0.5\baselineskip}[0pt][0pt]{%
    \textbf{Global}%
  }
& \raisebox{0.5\baselineskip}[0pt][0pt]{%
    \textbf{Region}%
  }
& \raisebox{0.5\baselineskip}[0pt][0pt]{%
    \textbf{Object}%
  }
& \raisebox{0.5\baselineskip}[0pt][0pt]{%
    \textbf{Avg.}%
  }
& \shortstack[c]{%
    \textbf{$\Delta$ vs.}\\
    \textbf{Random}%
  }
\\

\midrule

Random
& \secondcell{77.84}
& 48.78
& 54.95
& 58.97
& --
\\

SigLIP2
& 77.58
& 49.52
& 55.83
& 59.54
& +0.58
\\

RemoteCLIP
& 77.75
& \bestcell{50.78}
& \secondcell{56.24}
& \secondcell{60.10}
& \secondcell{+1.13}
\\

\midrule

\shortstack[l]{%
  \textbf{GazeEarth}%
}
& \bestcell{77.96}
& \secondcell{50.39}
& \bestcell{58.04}
& \bestcell{61.00}
& \bestcell{+2.03}
\\

\bottomrule
\end{tabularx}

\end{table*}

\begin{table*}[htbp]
\centering

\caption{Observer substitution with
Qwen3-VL-8B as the fixed answerer on XLRS-Bench. Ties are broken using the higher overall average.}
\label{tab:observer-substitution}

\vspace{-0.15cm}

\scriptsize
\renewcommand{\arraystretch}{0.92}
\setlength{\tabcolsep}{1.5pt}
\setlength{\aboverulesep}{0.25ex}
\setlength{\belowrulesep}{0.25ex}

\begin{tabularx}{\textwidth}{
@{}
>{\itshape\raggedright\arraybackslash}p{0.28\textwidth}
*{10}{Y}
@{}
}
\toprule

\raisebox{0.5\baselineskip}[0pt][0pt]{%
\normalfont\bfseries Observer%
}
& \raisebox{0.5\baselineskip}[0pt][0pt]{%
\textbf{Cnt}%
}
& \raisebox{0.5\baselineskip}[0pt][0pt]{%
\textbf{SC}%
}
& \raisebox{0.5\baselineskip}[0pt][0pt]{%
\textbf{OSR}%
}
& \raisebox{0.5\baselineskip}[0pt][0pt]{%
\textbf{OP}%
}
& \raisebox{0.5\baselineskip}[0pt][0pt]{%
\textbf{Plan}%
}
& \raisebox{0.5\baselineskip}[0pt][0pt]{%
\textbf{AR}%
}
& \raisebox{0.5\baselineskip}[0pt][0pt]{%
\textbf{CR}%
}
& \raisebox{0.5\baselineskip}[0pt][0pt]{%
\textbf{SR}%
}
& \raisebox{0.5\baselineskip}[0pt][0pt]{%
\textbf{Avg.}%
}
& \shortstack[c]{%
\textbf{$\Delta$ vs.}\\
\textbf{Random}%
}
\\

\midrule

Random
& 35.00
& 61.33
& 30.60
& 44.10
& 39.00
& \bestcell{78.00}
& 65.50
& 43.33
& 45.42
& 0.00
\\

SigLIP2
& 36.25
& \secondcell{62.67}
& \secondcell{31.40}
& \secondcell{45.12}
& \secondcell{39.00}
& 73.00
& \secondcell{65.50}
& 41.67
& \secondcell{46.10}
& \secondcell{+0.68}
\\

RemoteCLIP
& \secondcell{37.50}
& 61.67
& 29.80
& 45.12
& 38.00
& \secondcell{77.00}
& 65.50
& \secondcell{43.33}
& 45.94
& +0.52
\\

\midrule

\shortstack[l]{\textbf{GazeEarth}
}
& \bestcell{44.38}
& \bestcell{66.67}
& \bestcell{37.80}
& \bestcell{53.07}
& \bestcell{49.00}
& 76.00
& \bestcell{74.50}
& \bestcell{55.00}
& \bestcell{53.51}
& \bestcell{+8.09}
\\

\bottomrule
\end{tabularx}

\end{table*}

\subsection{Effect of Gaze Depth}
\label{app:gaze-depth}
We evaluate gaze depth through recursive reselection
and refocusing of the current view.
Among these configurations, a single gaze achieves
the highest benchmark-averaged accuracy with
the lowest runtime.
In Table~\ref{tab:visual-evidence-presentation},
GazeEarth reaches 55.65\% benchmark-averaged accuracy,
compared with 52.91\% for Two-Gaze and 52.72\%
for Three-Gaze.
Meanwhile, average inference time increases from
5.07 seconds to 5.95 and 7.02 seconds, respectively
(Table~\ref{tab:qwen3-vl-inference-efficiency}).
The single-gaze configuration also achieves the
highest overall accuracy on each of the three
benchmarks among these depth variants.

In the repeated-selection variants, each subsequent
gaze selects regions from an indexed overview of
the already warped scene and applies another focus
transformation to that view.
The transformations therefore compose: each new request
reallocates the current canvas, and a shift in focus
can compress regions enlarged by an earlier round.
The answerer receives the final transformed view
rather than an accumulated collection of observations.
Repeated selection thus revises the spatial allocation
within the same canvas.
The single-gaze configuration instead uses one spatial
request to construct the full-scene observation directly
from the original image, achieving the strongest
accuracy--runtime trade-off among the evaluated
gaze depths.

\subsection{Random Focus versus Overview}
\label{app:random-focus}

Random Focus applies the same full-scene focus operator
to randomly selected regions.
Its benchmark-averaged accuracy is 50.01\%, compared
with 51.38\% for Overview
(Tables~\ref{tab:observer-substitution-qwen3vl8b}
and~\ref{tab:visual-evidence-presentation}).
On LRS-GRO, accuracy decreases from 59.99\% to 58.97\%;
on MME-RealWorld-RS, it decreases from 49.01\% to 45.64\%.
XLRS-Bench increases from 45.13\% to 45.42\%.
The LRS-GRO breakdown
(Table~\ref{tab:category-level-controlled-analysis-lrs-gro})
places the largest reductions in Region and Object
questions, which fall from 50.22\% to 48.78\%
and from 56.22\% to 54.95\%, respectively.
Under a fixed canvas budget, enlarging a region
compresses the periphery, making the location of
the allocated pixels consequential for answering.

With the focus operator and answerer held fixed,
SigLIP2 and RemoteCLIP each achieve a three-benchmark
mean accuracy of 51.04\%, compared with 50.01\%
for random selection.
Self-directed gaze reaches 55.65\% and is the only
evaluated observer configuration to exceed
Overview's mean accuracy of 51.38\%, by
4.27 percentage points.

\FloatBarrier
\section{Statistical Analysis}
\label{app:Statistical}

To rigorously evaluate the performance improvements, we conducted paired image-cluster label-swap permutation tests across the three benchmarks to calculate $p$-values, while 95\% confidence intervals (CIs) were estimated via paired bootstrap. To control the family-wise error rate across the nine primary comparisons, all reported $p$-values are adjusted using the Holm-Bonferroni method ($p_{\text{Holm}}$). As shown in Table \ref{tab:statistical-significance}, GazeEarth yields highly significant accuracy gains over the Direct baseline across all datasets ($p_{\text{Holm}} < 0.001$). When compared to the full-scene Overview, GazeEarth demonstrates strongly significant improvements on contextually demanding datasets, achieving a gain of +3.42 percentage points on MME-RealWorld-RS ($p_{\text{Holm}} < 0.001$) and a substantial +8.38 points on XLRS-Bench ($p_{\text{Holm}} < 0.001$). On LRS-GRO, where the uniform Overview strategy already reaches a high performance threshold (59.99\%), the additional gain from GazeEarth (+1.01 points) is comparatively marginal and not statistically significant after adjustment ($p_{\text{Holm}} = 0.078$). These results indicate that GazeEarth reliably extracts critical local evidence to overcome resolution bottlenecks, while functioning as a stable, non-degrading observation mechanism when global views are already sufficient.

\begin{table}[htbp]
\centering
\caption{\textbf{Statistical significance of paired comparisons.} $\Delta$ indicates the accuracy difference (percentage points) between the first and second method. The 95\% confidence intervals (CIs) are derived from paired image-cluster bootstrap. $p$-values are computed via paired image-cluster label-swap tests and adjusted across the nine comparisons using the Holm-Bonferroni method ($p_{\text{Holm}}$).}
\label{tab:statistical-significance}
\vspace{0.1cm}
\scriptsize
\renewcommand{\arraystretch}{0.95}
\setlength{\tabcolsep}{6pt}
\begin{tabular}{llrcr}
\toprule
\textbf{Benchmark} & \textbf{Comparison} & \textbf{$\Delta$} & \textbf{95\% CI} & \textbf{$p_{\text{Holm}}$} \\
\midrule
XLRS-Bench & Overview $-$ Direct & $+2.21$ & $[+0.85, +3.54]$ & $0.004$ \\
           & GazeEarth $-$ Direct & $+10.59$ & $[+9.05, +12.13]$ & $< 0.001$ \\
           & GazeEarth $-$ Overview & $+8.38$ & $[+6.85, +9.91]$ & $< 0.001$ \\
\midrule
LRS-GRO    & Overview $-$ Direct & $+3.09$ & $[+2.15, +4.03]$ & $< 0.001$ \\
           & GazeEarth $-$ Direct & $+4.10$ & $[+3.08, +5.12]$ & $< 0.001$ \\
           & GazeEarth $-$ Overview & $+1.01$ & $[-0.12, +2.14]$ & $0.078$ \\
\midrule
MME-RealWorld-RS & Overview $-$ Direct & $+6.90$ & $[+5.11, +8.69]$ & $< 0.001$ \\
                 & GazeEarth $-$ Direct & $+10.32$ & $[+8.45, +12.19]$ & $< 0.001$ \\
                 & GazeEarth $-$ Overview & $+3.42$ & $[+1.85, +4.99]$ & $< 0.001$ \\
\bottomrule
\end{tabular}
\end{table}

\FloatBarrier
\section{Efficiency Analysis}
\label{app:efficiency}

Table~\ref{tab:qwen3-vl-inference-efficiency} reports
end-to-end runtime across the three benchmarks.
Tables~\ref{tab:qwen3-vl-efficiency-mme-realworld-rs}--~\ref{tab:qwen3-vl-efficiency-xlrs} decompose this runtime
into selection, visual construction, and answering,
using the protocol in Appendix~\ref{app:hardware}.

\begin{table*}[htbp]
\centering

\caption{Qwen3-VL-8B inference efficiency. Runtime is measured
in seconds; lower is better. Speedup is relative to Direct;
higher is better.}
\label{tab:qwen3-vl-inference-efficiency}

\vspace{-0.15cm}

\scriptsize
\renewcommand{\arraystretch}{0.92}
\setlength{\tabcolsep}{1.5pt}
\setlength{\aboverulesep}{0.25ex}
\setlength{\belowrulesep}{0.25ex}

\begin{tabularx}{\textwidth}{
  @{}
  >{\itshape\raggedright\arraybackslash}p{0.28\linewidth}
  *{5}{Y}
  @{}
}
\toprule

\raisebox{1.5\baselineskip}[0pt][0pt]{%
  \normalfont\bfseries Variant%
}
& \raisebox{0.5\baselineskip}[0pt][0pt]{%
    \shortstack[c]{%
      \textbf{XLRS-}\\
      \textbf{Bench}\\
      \textbf{(s)} $\downarrow$%
    }%
  }
& \raisebox{0.5\baselineskip}[0pt][0pt]{%
    \shortstack[c]{%
      \textbf{LRS-}\\
      \textbf{GRO}\\
      \textbf{(s)} $\downarrow$%
    }%
  }
& \shortstack[c]{%
    \textbf{MME-}\\
    \textbf{RealWorld-}\\
    \textbf{RS}\\
    \textbf{(s)} $\downarrow$%
  }
& \raisebox{1.0\baselineskip}[0pt][0pt]{%
    \shortstack[c]{%
      \textbf{Avg.}\\
      \textbf{(s)} $\downarrow$%
    }%
  }
& \raisebox{1.0\baselineskip}[0pt][0pt]{%
    \shortstack[c]{%
      \textbf{Speedup}\\
      $\boldsymbol{\uparrow}$%
    }%
  }
\\

\midrule

\multicolumn{6}{@{}l}{%
  \textit{\textbf{Visual Evidence Composition}}} \\[-0.25ex]

Direct
& 8.051
& 6.417
& 7.563
& 7.344
& -- \\

Overview
& \secondcell{4.096}
& \bestcell{3.967}
& \bestcell{3.998}
& \bestcell{4.020}
& \bestcell{1.827$\times$} \\

Random Focus
& 4.687
& 5.124
& 5.019
& 4.943
& 1.486$\times$ \\

Gaze + Crop
& \bestcell{3.945}
& \secondcell{4.689}
& 4.661
& \secondcell{4.432}
& \secondcell{1.657$\times$} \\

Think with Image
& 4.690
& 5.323
& 5.088
& 5.034
& 1.459$\times$ \\

Linked Global + Crop
& 4.702
& 5.307
& 5.365
& 5.125
& 1.433$\times$
\\

AttWarp
& 4.306
& 4.786
& \secondcell{4.209}
& 4.434
& 1.656$\times$
\\

\midrule

\multicolumn{6}{@{}l}{%
  \textit{\textbf{Gaze Depth and Iterative Baselines}}} \\[-0.25ex]

Two-Gaze
& 5.753
& 6.170
& 5.936
& 5.953
& 1.234$\times$ \\

Three-Gaze
& 6.835
& 7.298
& 6.931
& 7.021
& 1.046$\times$ \\

\textbf{GazeEarth}
& 4.789
& 5.135
& 5.272
& 5.065
& 1.450$\times$ \\

\bottomrule
\end{tabularx}

\end{table*}

\begin{table*}[htbp]
\centering

\caption{Qwen3-VL-8B inference efficiency on
MME-RealWorld-RS. Lower is better for the number of MLLM calls
and runtime components, while higher is better for speedup. Ties are
broken using the lower total runtime.}
\label{tab:qwen3-vl-efficiency-mme-realworld-rs}

\vspace{-0.15cm}

\scriptsize
\renewcommand{\arraystretch}{0.92}
\setlength{\tabcolsep}{1.5pt}
\setlength{\aboverulesep}{0.25ex}
\setlength{\belowrulesep}{0.25ex}

\begin{tabularx}{\textwidth}{
  @{}
  >{\itshape\raggedright\arraybackslash}p{0.28\textwidth}
  *{6}{Y}
  @{}
}
\toprule

\raisebox{1.0\baselineskip}[0pt][0pt]{%
  \normalfont\bfseries Variant%
}
& \raisebox{0.5\baselineskip}[0pt][0pt]{%
    \shortstack[c]{%
      \textbf{MLLM}\\
      \textbf{calls} $\downarrow$%
    }%
  }
& \raisebox{0.5\baselineskip}[0pt][0pt]{%
    \shortstack[c]{%
      \textbf{Selection}\\
      \textbf{(s)} $\downarrow$%
    }%
  }
& \shortstack[c]{%
    \textbf{Visual}\\
    \textbf{construction}\\
    \textbf{(s)} $\downarrow$%
  }
& \raisebox{0.5\baselineskip}[0pt][0pt]{%
    \shortstack[c]{%
      \textbf{Answer}\\
      \textbf{(s)} $\downarrow$%
    }%
  }
& \raisebox{0.5\baselineskip}[0pt][0pt]{%
    \shortstack[c]{%
      \textbf{Total}\\
      \textbf{(s)} $\downarrow$%
    }%
  }
& \shortstack[c]{%
    \textbf{Speedup}\\
    \textbf{vs. Direct}\\
    $\boldsymbol{\uparrow}$%
  }
\\

\midrule

\multicolumn{7}{@{}l}{%
  \textit{\textbf{Visual Evidence Composition}}%
}
\\[-0.25ex]

Direct
& 1
& --
& --
& 7.563
& 7.563
& --
\\

Overview
& 1
& --
& \bestcell{0.147}
& 3.851
& \bestcell{3.998}
& \bestcell{1.892$\times$}
\\

Random Focus
& 2
& \secondcell{0.889}
& 0.325
& 3.805
& 5.019
& 1.507$\times$
\\

Gaze + Crop
& 2
& 0.905
& \secondcell{0.176}
& \bestcell{3.580}
& 4.661
& 1.623$\times$
\\

Think with Image
& 2
& \bestcell{0.884}
& 0.252
& 3.952
& 5.088
& 1.486$\times$
\\

Linked Global + Crop
& 2
& 0.986
& 0.305
& 4.075
& 5.365
& 1.410$\times$
\\

AttWarp
& 2
& --
& --
& --
& \secondcell{4.209}
& \secondcell{1.797$\times$}
\\

\midrule

\multicolumn{7}{@{}l}{%
  \textit{\textbf{Gaze Depth and Iterative Baselines}}%
}
\\[-0.25ex]

Two-Gaze
& 3
& 1.763
& 0.370
& \secondcell{3.803}
& 5.936
& 1.274$\times$
\\

Three-Gaze
& 4
& 2.683
& 0.445
& \secondcell{3.803}
& 6.931
& 1.091$\times$
\\

\textbf{GazeEarth}
& \textbf{$\leq 2$}
& 1.011
& 0.355
& 3.907
& 5.272
& 1.435$\times$
\\

\bottomrule
\end{tabularx}

\end{table*}

\begin{table*}[htbp]
\centering

\caption{Qwen3-VL-8B inference efficiency on LRS-GRO.
Lower is better for the number of MLLM calls and runtime
components, while higher is better for speedup. Ties are broken using the lower
total runtime.}
\label{tab:qwen3-vl-efficiency-lrs-gro}

\vspace{-0.15cm}

\scriptsize
\renewcommand{\arraystretch}{0.92}
\setlength{\tabcolsep}{1.5pt}
\setlength{\aboverulesep}{0.25ex}
\setlength{\belowrulesep}{0.25ex}

\begin{tabularx}{\textwidth}{
  @{}
  >{\itshape\raggedright\arraybackslash}p{0.28\textwidth}
  *{6}{Y}
  @{}
}
\toprule

\raisebox{1.0\baselineskip}[0pt][0pt]{%
  \normalfont\bfseries Variant%
}
& \raisebox{0.5\baselineskip}[0pt][0pt]{%
    \shortstack[c]{%
      \textbf{MLLM}\\
      \textbf{calls} $\downarrow$%
    }%
  }
& \raisebox{0.5\baselineskip}[0pt][0pt]{%
    \shortstack[c]{%
      \textbf{Selection}\\
      \textbf{(s)} $\downarrow$%
    }%
  }
& \shortstack[c]{%
    \textbf{Visual}\\
    \textbf{construction}\\
    \textbf{(s)} $\downarrow$%
  }
& \raisebox{0.5\baselineskip}[0pt][0pt]{%
    \shortstack[c]{%
      \textbf{Answer}\\
      \textbf{(s)} $\downarrow$%
    }%
  }
& \raisebox{0.5\baselineskip}[0pt][0pt]{%
    \shortstack[c]{%
      \textbf{Total}\\
      \textbf{(s)} $\downarrow$%
    }%
  }
& \shortstack[c]{%
    \textbf{Speedup}\\
    \textbf{vs. Direct}\\
    $\boldsymbol{\uparrow}$%
  }
\\

\midrule

\multicolumn{7}{@{}l}{%
  \textit{\textbf{Visual Evidence Composition}}%
}
\\[-0.25ex]

Direct
& 1
& --
& --
& 6.417
& 6.417
& --
\\

Overview
& 1
& --
& \secondcell{0.169}
& 3.798
& \bestcell{3.967}
& \bestcell{1.618$\times$}
\\

Random Focus
& 2
& \secondcell{1.118}
& 0.212
& \secondcell{3.794}
& 5.124
& 1.252$\times$
\\

Gaze + Crop
& 2
& 1.141
& \bestcell{0.060}
& \bestcell{3.488}
& \secondcell{4.689}
& \secondcell{1.369$\times$}
\\

Think with Image
& 2
& 1.155
& 0.192
& 3.976
& 5.323
& 1.206$\times$
\\

Linked Global + Crop
& 2
& \bestcell{0.970}
& 0.364
& 3.973
& 5.307
& 1.209$\times$
\\

AttWarp
& 2
& --
& --
& --
& 4.786
& 1.341$\times$
\\

\midrule

\multicolumn{7}{@{}l}{%
  \textit{\textbf{Gaze Depth and Iterative Baselines}}%
}
\\[-0.25ex]

Two-Gaze
& 3
& 2.117
& 0.249
& 3.804
& 6.170
& 1.040$\times$
\\

Three-Gaze
& 4
& 3.195
& 0.288
& 3.815
& 7.298
& 0.879$\times$
\\

\shortstack[l]{\textbf{GazeEarth}}
& \textbf{$\leq 2$}
& 1.125
& 0.210
& 3.800
& 5.135
& 1.250$\times$
\\

\bottomrule
\end{tabularx}

\vspace{-0.3cm}
\end{table*}

\begin{table*}[htbp]
\centering

\caption{Qwen3-VL-8B inference efficiency on XLRS-Bench.
Lower is better for the number of MLLM calls and runtime
components, while higher is better for speedup. Ties are broken using the lower
total runtime.}
\label{tab:qwen3-vl-efficiency-xlrs}

\vspace{-0.15cm}

\scriptsize
\renewcommand{\arraystretch}{0.92}
\setlength{\tabcolsep}{1.5pt}
\setlength{\aboverulesep}{0.25ex}
\setlength{\belowrulesep}{0.25ex}

\begin{tabularx}{\textwidth}{
@{}
>{\itshape\raggedright\arraybackslash}p{0.28\textwidth}
*{6}{Y}
@{}
}
\toprule

\raisebox{1.0\baselineskip}[0pt][0pt]{%
\normalfont\bfseries Variant%
}
& \raisebox{0.5\baselineskip}[0pt][0pt]{%
\shortstack[c]{%
\textbf{MLLM}\\
\textbf{calls} $\downarrow$%
}%
}
& \raisebox{0.5\baselineskip}[0pt][0pt]{%
\shortstack[c]{%
\textbf{Selection}\\
\textbf{(s)} $\downarrow$%
}%
}
& \shortstack[c]{%
\textbf{Visual}\\
\textbf{construction}\\
\textbf{(s)} $\downarrow$%
}
& \raisebox{0.5\baselineskip}[0pt][0pt]{%
\shortstack[c]{%
\textbf{Answer}\\
\textbf{(s)} $\downarrow$%
}%
}
& \raisebox{0.5\baselineskip}[0pt][0pt]{%
\shortstack[c]{%
\textbf{Total}\\
\textbf{(s)} $\downarrow$%
}%
}
& \shortstack[c]{%
\textbf{Speedup}\\
\textbf{vs. Direct}\\
$\boldsymbol{\uparrow}$%
}
\\

\midrule

\multicolumn{7}{@{}l}{%
\textit{\textbf{Visual Evidence Composition}}%
}
\\[-0.25ex]

Direct
& 1
& --
& --
& 8.051
& 8.051
& --
\\

Overview
& 1
& --
& \secondcell{0.331}
& 3.765
& \secondcell{4.096}
& \secondcell{1.966$\times$}
\\

Random Focus
& 2
& \bestcell{0.459}
& 0.481
& \secondcell{3.747}
& 4.687
& 1.718$\times$
\\

Gaze + Crop
& 2
& \secondcell{0.465}
& \bestcell{0.193}
& \bestcell{3.287}
& \bestcell{3.945}
& \bestcell{2.041$\times$}
\\

Unlinked Global + Crop
& 2
& 0.474
& 0.422
& 3.794
& 4.690
& 1.717$\times$
\\

Linked Global + Crop
& 2
& 0.483
& 0.421
& 3.798
& 4.702
& 1.712$\times$
\\

AttWarp
& 2
& --
& --
& --
& 4.306
& 1.870$\times$
\\

\midrule

\multicolumn{7}{@{}l}{%
\textit{\textbf{Gaze Depth and Iterative Baselines}}%
}
\\[-0.25ex]

Two-Gaze
& 3
& 1.427
& 0.570
& 3.756
& 5.753
& 1.399$\times$
\\

Three-Gaze
& 4
& 2.430
& 0.654
& 3.751
& 6.835
& 1.178$\times$
\\

\shortstack[l]{\textbf{GazeEarth}}
& \textbf{$\leq 2$}
& 0.474
& 0.532
& 3.783
& 4.789
& 1.681$\times$
\\

\bottomrule
\end{tabularx}

\vspace{-0.3cm}
\end{table*}

\subsection{Model calls and observation construction.}
GazeEarth uses one question-guided selection call
followed by one answering call; a directly resolved
spatial request bypasses selection.
The focus operator constructs the observation between
these calls through deterministic resampling.
Selection takes 1.011 seconds on MME-RealWorld-RS
and 1.125 seconds on LRS-GRO.
Visual construction, including focused-view rendering,
takes 0.355 and 0.210 seconds, respectively,
accounting for 6.7\% and 4.1\% of total runtime.
The selection call therefore contributes most of
the additional cost over Overview.

\subsection{Answering cost.}
Answer generation accounts for approximately
three quarters of GazeEarth's total runtime.
It takes 3.907 seconds on MME-RealWorld-RS
and 3.800 seconds on LRS-GRO,
close to Overview's 3.851 and 3.798 seconds.
Compared with Direct, the answering stage is shorter
by 3.656 and 2.617 seconds, respectively.
These reductions exceed the combined selection
and visual-construction costs of 1.366 and
1.335 seconds, producing the overall runtime
advantage reported in the main text.

\begin{figure*}[htbp]
  \centering
  \vspace{-0.1cm}

  \includegraphics[
    width=0.55\linewidth,
    keepaspectratio
  ]{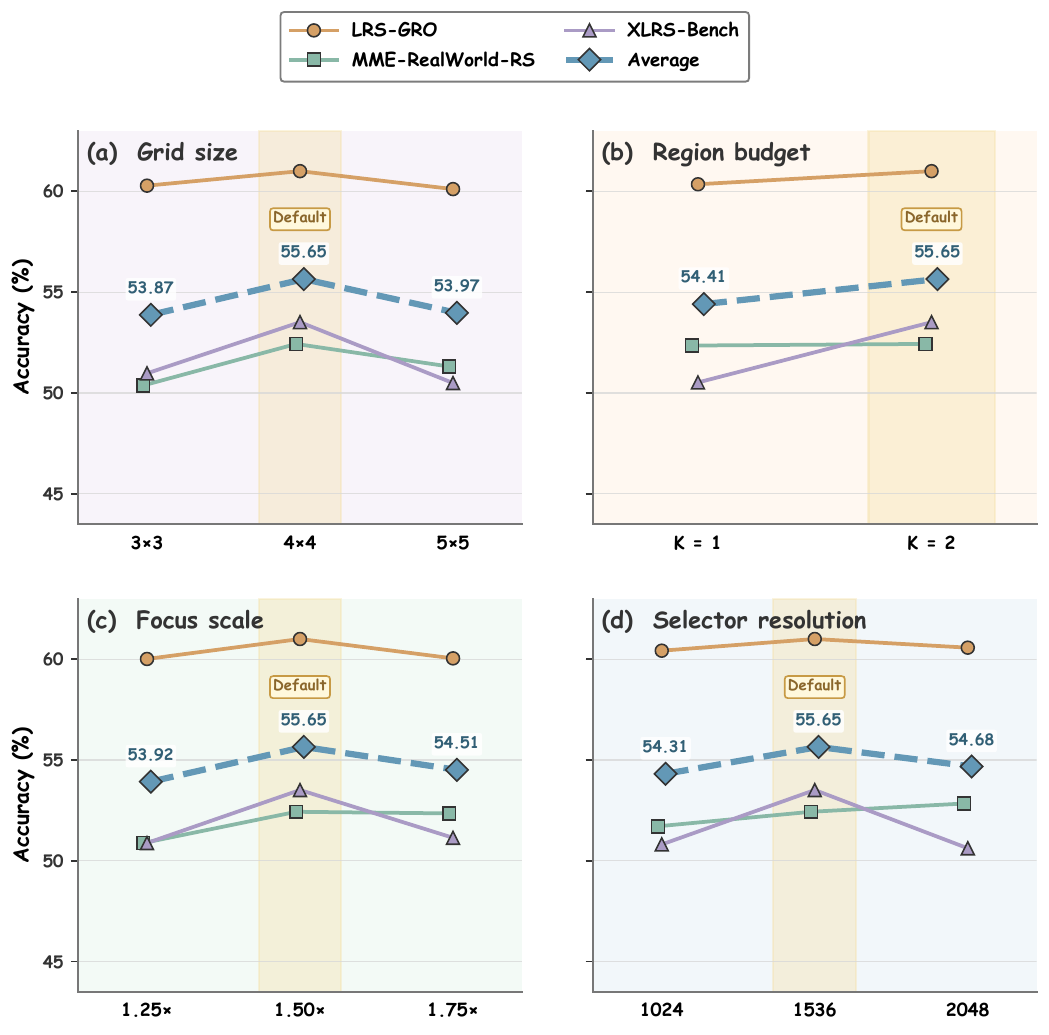}

  \vspace{-0.15cm}
  \caption{Hyperparameter sensitivity of GazeEarth with
    Qwen3-VL-8B: (a) selection-grid size,
    (b) maximum number of selected cells,
    (c) focus enlargement factor, and
    (d) selector-overview long-side cap in pixels.
    Solid curves show individual benchmark scores;
    the dashed curve shows their unweighted mean.
    Shaded bands mark the default settings.}
  \label{fig:hyper}
  \vspace{-0.2cm}
\end{figure*}

\FloatBarrier
\section{Hyperparameter Sensitivity}
\label{app:sensitivity}

Figure~\ref{fig:hyper} examines four settings that control
spatial selection and pixel allocation.
The default configuration uses a $4\times4$ grid,
a maximum of two selected cells, a focus enlargement
factor of $\alpha=1.5$, and a selector-overview
long-side cap of 1536 pixels.
Tables~\ref{tab:parameter-sensitivity-mme-realworld-rs}
--~\ref{tab:parameter-sensitivity-xlrs} provide category-level
results on MME-RealWorld-RS, LRS-GRO, and XLRS-Bench.

\subsection{Selection-grid size.}
The $4\times4$ grid achieves a benchmark-averaged
accuracy of 55.65\%, compared with 53.87\% for
$3\times3$ and 53.97\% for $5\times5$.
It also gives the highest overall score on each
benchmark among the three grid settings.
Increasing selection granularity beyond $4\times4$ therefore does not improve answering performance.

\subsection{Region budget.}
Increasing the maximum number of selected cells
from one to two raises mean accuracy from
54.41\% to 55.65\%.
On LRS-GRO, Object accuracy increases from
56.69\% to 58.04\%, while Global accuracy increases
from 77.84\% to 77.96\%.
A two-cell budget allows a spatial request to cover
separated evidence within a single selection round.

\subsection{Focus scale.}
The enlargement factor controls how much output area
is assigned to the selected neighbourhood.
Mean accuracy reaches 55.65\% at $\alpha=1.5$,
compared with 53.92\% at 1.25 and 54.51\% at 1.75.
The category-level results reveal different responses
to stronger magnification.
On MME-RealWorld-RS, increasing $\alpha$ from
1.5 to 1.75 improves Color and Count accuracy
but reduces Position accuracy from 71.12\% to 68.10\%.
The intermediate scale yields the highest
cross-benchmark mean.

\subsection{Selector resolution.}
A 1536-pixel overview achieves a mean accuracy
of 55.65\%, compared with 54.31\% at 1024 pixels
and 54.68\% at 2048 pixels.
The higher-resolution overview improves
MME-RealWorld-RS accuracy from 52.43\% to 52.84\%,
whereas LRS-GRO decreases from 61.00\% to 60.57\%.
The default resolution thus provides the highest
mean score across the three benchmarks.

\begin{table*}[htbp]
\centering

\caption{Category-level parameter sensitivity on
MME-RealWorld-RS. Bold setting names mark the default configuration. Best and second-best marks are assigned within each sweep and column; tied values receive the same mark.}
\label{tab:parameter-sensitivity-mme-realworld-rs}

\vspace{-0.15cm}

\scriptsize
\renewcommand{\arraystretch}{0.92}
\setlength{\tabcolsep}{1.5pt}
\setlength{\aboverulesep}{0.25ex}
\setlength{\belowrulesep}{0.25ex}

\begin{tabularx}{\textwidth}{
  @{}
  >{\itshape\raggedright\arraybackslash}p{0.28\textwidth}
  *{4}{Y}
  @{}
}
\toprule

{\normalfont\bfseries Setting}
& \textbf{Color}
& \textbf{Count}
& \textbf{Position}
& \textbf{Avg.}
\\

\midrule

\multicolumn{5}{@{}l}{%
  \textit{\textbf{Grid-size Sensitivity}}%
}
\\[-0.25ex]

Grid $3{\times}3$
& 61.75
& 20.88
& 67.78
& 50.37
\\

\textbf{Grid $4{\times}4$}
& \bestcell{63.67}
& \bestcell{21.78}
& \bestcell{71.12}
& \bestcell{52.43}
\\

Grid $5{\times}5$
& \secondcell{62.15}
& \bestcell{21.78}
& \secondcell{69.29}
& \secondcell{51.31}
\\

\midrule

\multicolumn{5}{@{}l}{%
  \textit{\textbf{Region-budget Sensitivity}}%
}
\\[-0.25ex]

$K=1$
& \bestcell{63.75}
& \bestcell{22.02}
& \secondcell{70.56}
& \secondcell{52.35}
\\

\textbf{$K=2$}
& \secondcell{63.67}
& \secondcell{21.78}
& \bestcell{71.12}
& \bestcell{52.43}
\\

\midrule

\multicolumn{5}{@{}l}{%
  \textit{\textbf{Focus-scale Sensitivity}}%
}
\\[-0.25ex]

Focus 1.25
& 59.36
& 20.55
& \bestcell{72.00}
& 50.88
\\

\textbf{Focus 1.50}
& \secondcell{63.67}
& \secondcell{21.78}
& \secondcell{71.12}
& \bestcell{52.43}
\\

Focus 1.75
& \bestcell{64.62}
& \bestcell{23.65}
& 68.10
& \secondcell{52.35}
\\

\midrule

\multicolumn{5}{@{}l}{%
  \textit{\textbf{Selector-resolution Sensitivity}}%
}
\\[-0.25ex]

Selector 1024
& 62.71
& \secondcell{21.94}
& 69.77
& 51.71
\\

\textbf{Selector 1536}
& \secondcell{63.67}
& 21.78
& \secondcell{71.12}
& \secondcell{52.43}
\\

Selector 2048
& \bestcell{64.06}
& \bestcell{22.43}
& \bestcell{71.28}
& \bestcell{52.84}
\\

\bottomrule
\end{tabularx}

\vspace{-0.3cm}
\end{table*}

\begin{table*}[htbp]
\centering

\caption{Category-level parameter sensitivity on LRS-GRO.}
\label{tab:parameter-sensitivity-lrs-gro}

\vspace{-0.15cm}

\scriptsize
\renewcommand{\arraystretch}{0.92}
\setlength{\tabcolsep}{1.5pt}
\setlength{\aboverulesep}{0.25ex}
\setlength{\belowrulesep}{0.25ex}

\begin{tabularx}{\textwidth}{
  @{}
  >{\itshape\raggedright\arraybackslash}p{0.28\textwidth}
  *{4}{Y}
  @{}
}
\toprule

{\normalfont\bfseries Setting}
& \textbf{Global}
& \textbf{Region}
& \textbf{Object}
& \textbf{Avg.}
\\

\midrule

\multicolumn{5}{@{}l}{%
  \textit{\textbf{Grid-size Sensitivity}}%
}
\\[-0.25ex]

Grid $3{\times}3$
& 77.23
& \bestcell{50.44}
& \secondcell{56.99}
& \secondcell{60.28}
\\

\textbf{Grid $4{\times}4$}
& \bestcell{77.96}
& \secondcell{50.39}
& \bestcell{58.04}
& \bestcell{61.00}
\\

Grid $5{\times}5$
& \secondcell{77.45}
& 49.78
& 56.85
& 60.11
\\

\midrule

\multicolumn{5}{@{}l}{%
  \textit{\textbf{Region-budget Sensitivity}}%
}
\\[-0.25ex]

$K=1$
& \secondcell{77.84}
& \bestcell{50.74}
& \secondcell{56.69}
& \secondcell{60.35}
\\

\textbf{$K=2$}
& \bestcell{77.96}
& \secondcell{50.39}
& \bestcell{58.04}
& \bestcell{61.00}
\\

\midrule

\multicolumn{5}{@{}l}{%
  \textit{\textbf{Focus-scale Sensitivity}}%
}
\\[-0.25ex]

Focus 1.25
& \secondcell{77.75}
& 50.17
& 56.34
& 60.01
\\

\textbf{Focus 1.50}
& \bestcell{77.96}
& \bestcell{50.39}
& \bestcell{58.04}
& \bestcell{61.00}
\\

Focus 1.75
& 77.32
& \secondcell{50.22}
& \secondcell{56.58}
& \secondcell{60.04}
\\

\midrule

\multicolumn{5}{@{}l}{%
  \textit{\textbf{Selector-resolution Sensitivity}}%
}
\\[-0.25ex]

Selector 1024
& 77.66
& \secondcell{50.83}
& 56.87
& 60.42
\\

\textbf{Selector 1536}
& \bestcell{77.96}
& 50.39
& \bestcell{58.04}
& \bestcell{61.00}
\\

Selector 2048
& \secondcell{77.79}
& \bestcell{51.18}
& \secondcell{56.95}
& \secondcell{60.57}
\\

\bottomrule
\end{tabularx}

\vspace{-0.3cm}
\end{table*}

\begin{table*}[htbp]
\centering

\caption{Category-level parameter sensitivity on XLRS-Bench.}
\label{tab:parameter-sensitivity-xlrs}

\vspace{-0.15cm}

\scriptsize
\renewcommand{\arraystretch}{0.92}
\setlength{\tabcolsep}{1.5pt}
\setlength{\aboverulesep}{0.25ex}
\setlength{\belowrulesep}{0.25ex}

\begin{tabularx}{\textwidth}{
@{}
>{\itshape\raggedright\arraybackslash}p{0.28\textwidth}
*{9}{Y}
@{}
}
\toprule

{\normalfont\bfseries Setting}
& \textbf{Cnt}
& \textbf{SC}
& \textbf{OSR}
& \textbf{OP}
& \textbf{Plan}
& \textbf{AR}
& \textbf{CR}
& \textbf{SR}
& \textbf{Avg.}
\\

\midrule

\multicolumn{10}{@{}l}{\textit{\textbf{Grid-size Sensitivity}}}\\[-0.25ex]
Grid $3{\times}3$
& \secondcell{43.13}
& \bestcell{67.67}
& 34.00
& \secondcell{50.42}
& 45.00
& \secondcell{77.00}
& \secondcell{70.50}
& \secondcell{46.67}
& \secondcell{50.97} \\

\textbf{Grid $4{\times}4$}
& \bestcell{44.38}
& \secondcell{66.67}
& \bestcell{37.80}
& \bestcell{53.07}
& \bestcell{49.00}
& 76.00
& \bestcell{74.50}
& \bestcell{55.00}
& \bestcell{53.51} \\

Grid $5{\times}5$
& 41.25
& 65.33
& \secondcell{34.20}
& 49.88
& \secondcell{47.00}
& \bestcell{78.00}
& \secondcell{70.50}
& \secondcell{46.67}
& 50.49 \\

\midrule
\multicolumn{10}{@{}l}{\textit{\textbf{Region-budget Sensitivity}}}\\[-0.25ex]
$K=1$
& \secondcell{40.63}
& \bestcell{67.00}
& \secondcell{32.80}
& \secondcell{50.06}
& \secondcell{45.00}
& \bestcell{79.00}
& \secondcell{71.00}
& \secondcell{48.33}
& \secondcell{50.52} \\

\textbf{$K=2$}
& \bestcell{44.38}
& \secondcell{66.67}
& \bestcell{37.80}
& \bestcell{53.07}
& \bestcell{49.00}
& \secondcell{76.00}
& \bestcell{74.50}
& \bestcell{55.00}
& \bestcell{53.51} \\

\midrule
\multicolumn{10}{@{}l}{\textit{\textbf{Focus-scale Sensitivity}}}\\[-0.25ex]
Focus 1.25
& \secondcell{41.88}
& \secondcell{67.67}
& 33.60
& 50.06
& \secondcell{47.00}
& \bestcell{78.00}
& \secondcell{72.50}
& \secondcell{46.67}
& 50.88 \\

\textbf{Focus 1.50}
& \bestcell{44.38}
& 66.67
& \bestcell{37.80}
& \bestcell{53.07}
& \bestcell{49.00}
& 76.00
& \bestcell{74.50}
& \bestcell{55.00}
& \bestcell{53.51} \\

Focus 1.75
& 41.25
& \bestcell{68.33}
& \secondcell{34.20}
& \secondcell{50.84}
& \secondcell{47.00}
& \secondcell{77.00}
& 69.00
& 45.00
& \secondcell{51.14} \\

\midrule
\multicolumn{10}{@{}l}{\textit{\textbf{Selector-resolution Sensitivity}}}\\[-0.25ex]
Selector 1024
& 40.00
& \bestcell{67.33}
& \secondcell{33.80}
& \secondcell{50.36}
& 45.00
& \secondcell{76.00}
& \secondcell{73.00}
& \secondcell{45.00}
& \secondcell{50.81} \\

\textbf{Selector 1536}
& \bestcell{44.38}
& \secondcell{66.67}
& \bestcell{37.80}
& \bestcell{53.07}
& \bestcell{49.00}
& \secondcell{76.00}
& \bestcell{74.50}
& \bestcell{55.00}
& \bestcell{53.51} \\

Selector 2048
& \secondcell{41.25}
& \secondcell{66.67}
& 32.00
& 50.30
& \secondcell{47.00}
& \bestcell{80.00}
& 72.00
& \secondcell{45.00}
& 50.62 \\
\bottomrule
\end{tabularx}

\vspace{-0.3cm}
\end{table*}

\FloatBarrier
\section{Qualitative Examples}
\label{app:qualitative}
This section illustrates how spatial requests guide
observation construction and subsequent answering.
The examples cover spatial responses, region selection,
and the resulting visual observations across methods.

\subsection{Spatial Response Visualization}
\label{app:spatial-response-visualization}

\begin{figure*}[htbp]
  \centerline{\includegraphics[width=\linewidth]{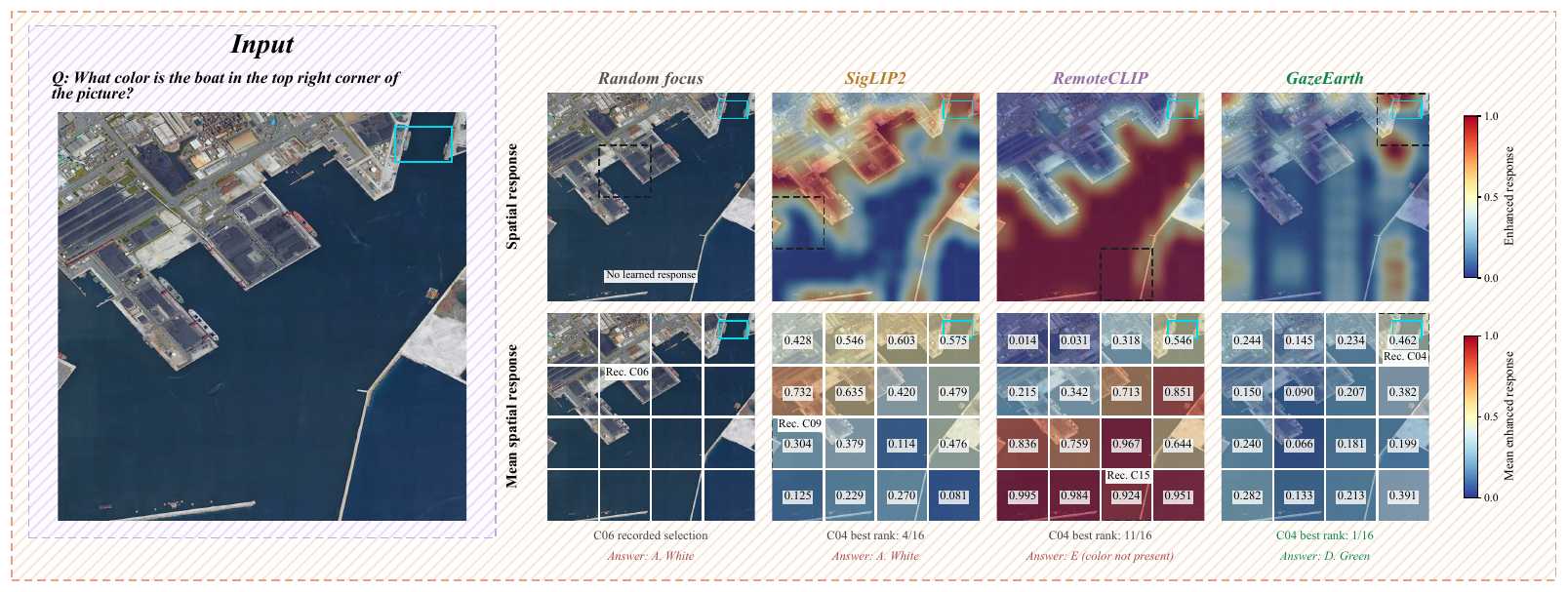}}
  \vspace{-0.01cm}
  \caption{Spatial responses and region selections
under observer substitution (Part 1).
Each example compares Random Focus, SigLIP2, RemoteCLIP,
and GazeEarth for the same image and question.
Spatial-response maps and cell-level summaries are shown
alongside the recorded selections and final answers.
Random Focus selects a cell without a learned response.}
  \vspace{-0.1cm}
  \label{figcase-new-1}
\end{figure*}

\begin{figure*}[htbp]
  \centerline{\includegraphics[width=\linewidth]{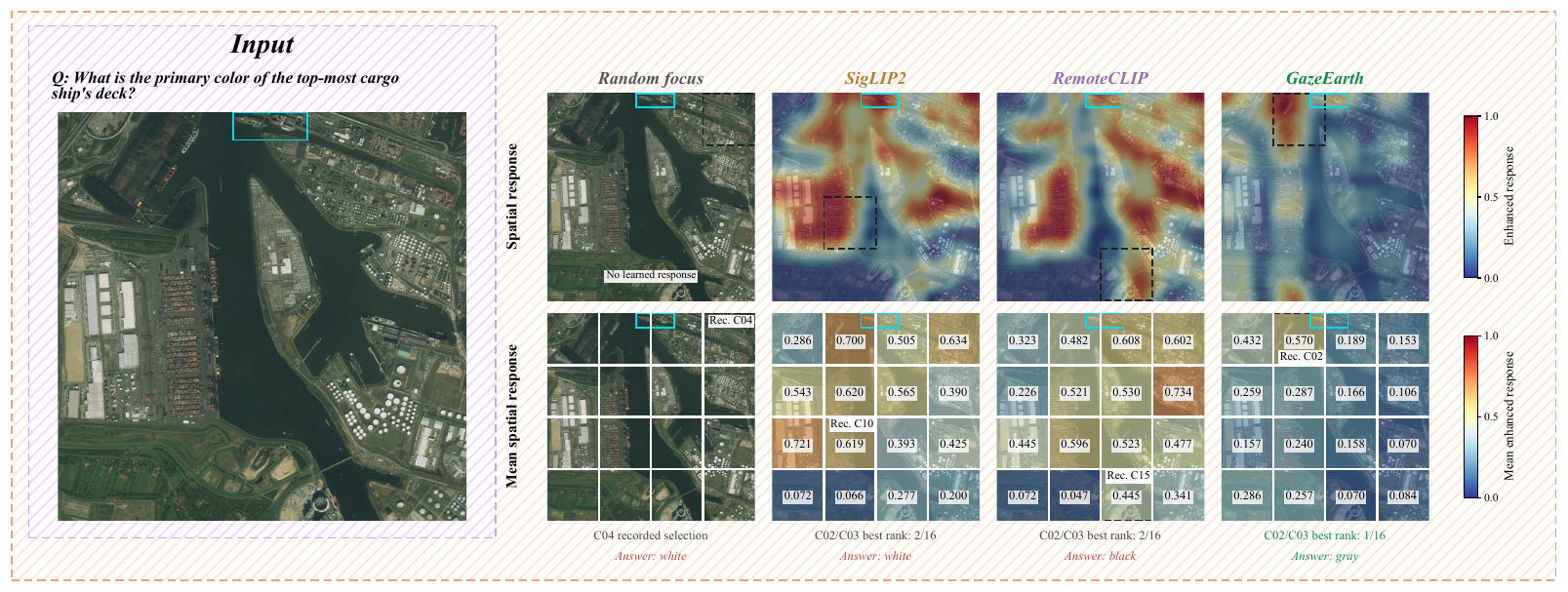}}
  \vspace{-0.01cm}
  \caption{Spatial responses and region selections
under observer substitution (Part 2).
Each example compares Random Focus, SigLIP2, RemoteCLIP,
and GazeEarth for the same image and question.
Spatial-response maps and cell-level summaries are shown
alongside the recorded selections and final answers.
Random Focus selects a cell without a learned response.}
  \vspace{-0.1cm}
  \label{figcase-new-2}
\end{figure*}

\begin{figure*}[htbp]
  \centerline{\includegraphics[width=\linewidth]{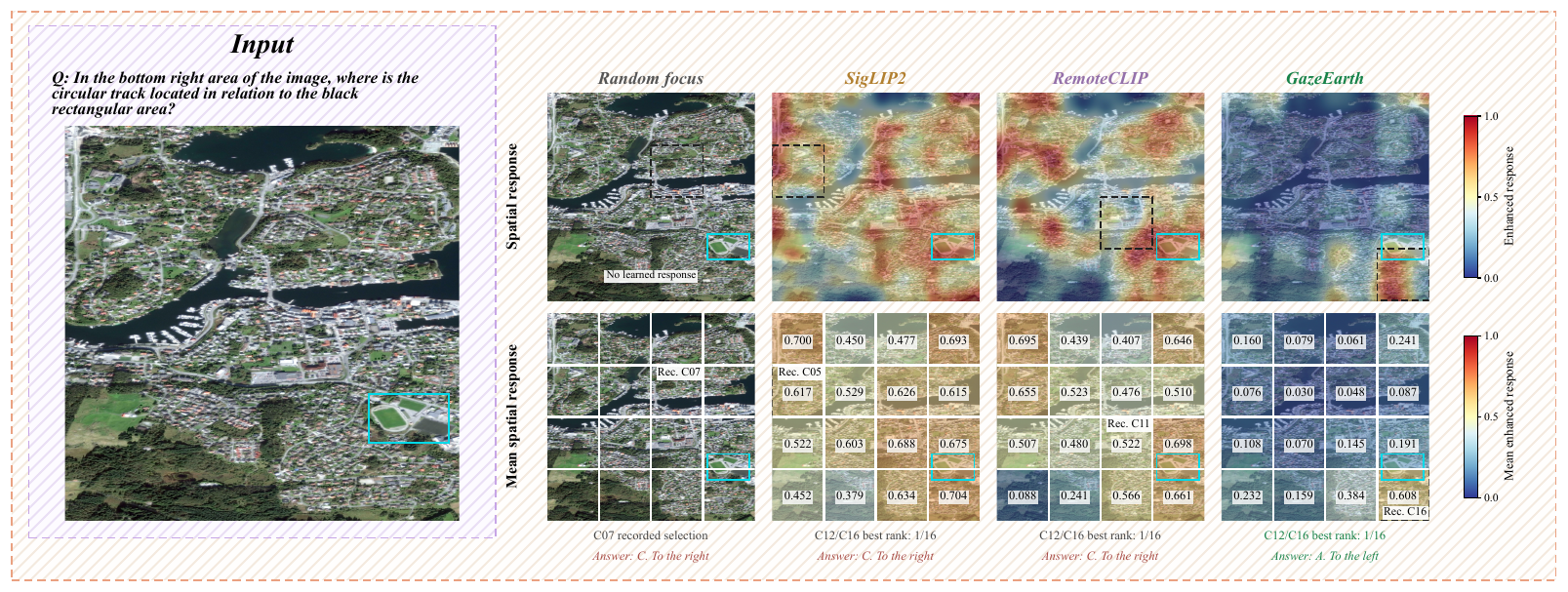}}
  \vspace{-0.01cm}
  \caption{Spatial responses and region selections
under observer substitution (Part 3).
Each example compares Random Focus, SigLIP2, RemoteCLIP,
and GazeEarth for the same image and question.
Spatial-response maps and cell-level summaries are shown
alongside the recorded selections and final answers.
Random Focus selects a cell without a learned response.}
  \vspace{-0.1cm}
  \label{figcase-new-4}
\end{figure*}

\begin{figure*}[htbp]
  \centerline{\includegraphics[width=\linewidth]{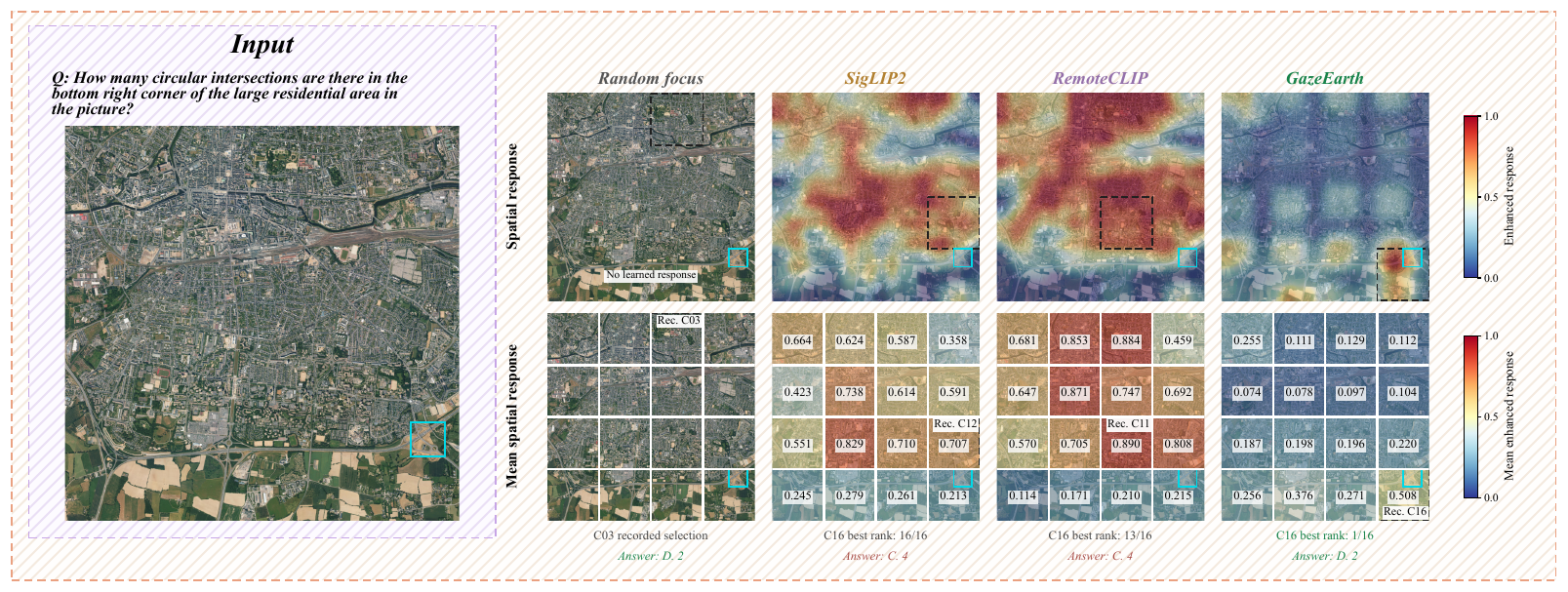}}
  \vspace{-0.01cm}
  \caption{Spatial responses and region selections
under observer substitution (Part 4).
Each example compares Random Focus, SigLIP2, RemoteCLIP,
and GazeEarth for the same image and question.
Spatial-response maps and cell-level summaries are shown
alongside the recorded selections and final answers.
Random Focus selects a cell without a learned response.}
  \vspace{-0.1cm}
  \label{figcase-new-5}
\end{figure*}

\begin{figure*}[htbp]
  \centerline{\includegraphics[width=\linewidth]{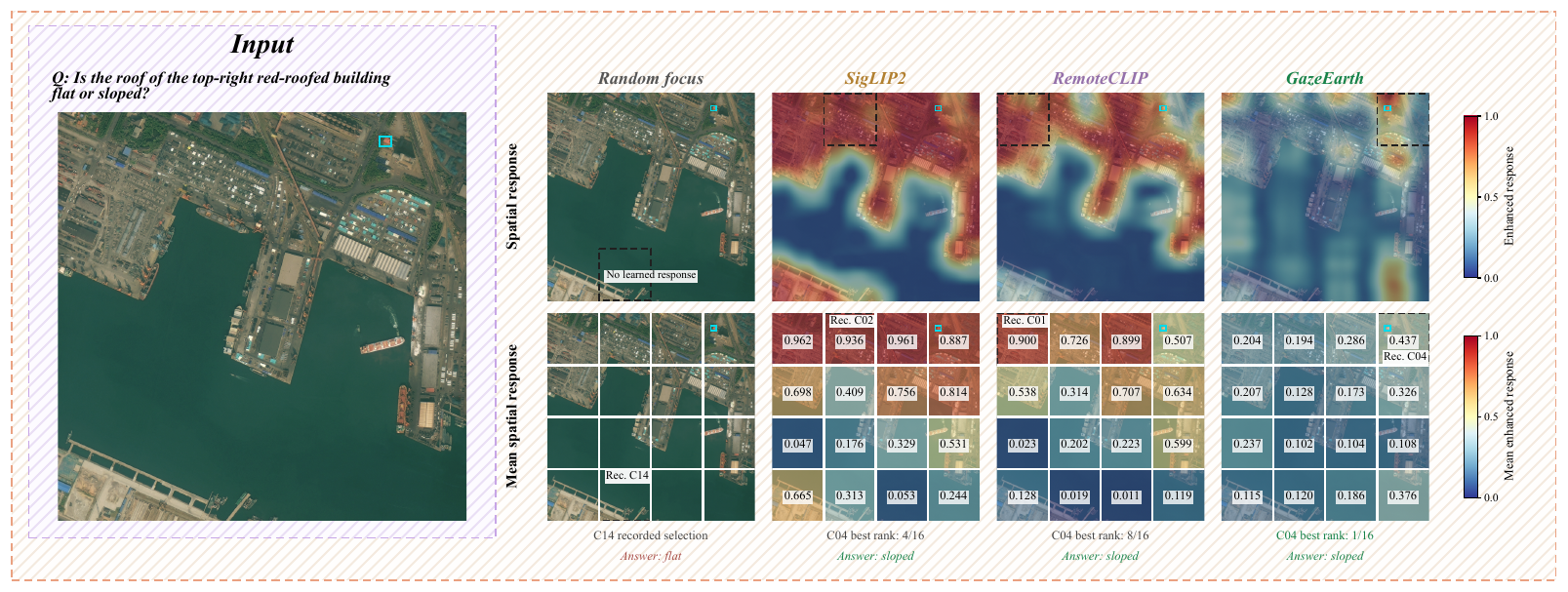}}
  \vspace{-0.01cm}
  \caption{Spatial responses and region selections
under observer substitution (Part 5).
Each example compares Random Focus, SigLIP2, RemoteCLIP,
and GazeEarth for the same image and question.
Spatial-response maps and cell-level summaries are shown
alongside the recorded selections and final answers.
Random Focus selects a cell without a learned response.}
  \vspace{-0.1cm}
  \label{figcase-new-6}
\end{figure*}

\begin{figure*}[htbp]
  \centerline{\includegraphics[width=\linewidth]{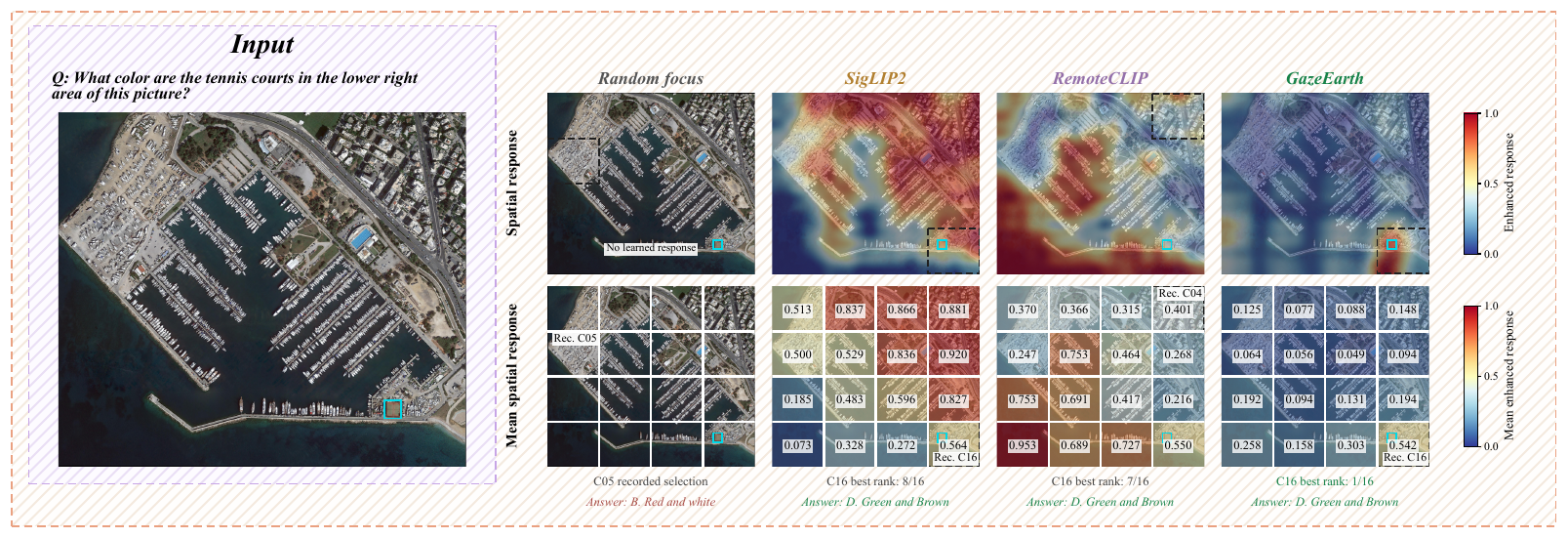}}
  \vspace{-0.01cm}
  \caption{Spatial responses and region selections
under observer substitution (Part 6).
Each example compares Random Focus, SigLIP2, RemoteCLIP,
and GazeEarth for the same image and question.
Spatial-response maps and cell-level summaries are shown
alongside the recorded selections and final answers.
Random Focus selects a cell without a learned response.}
  \vspace{-0.1cm}
  \label{figcase-new-9}
\end{figure*}

Figures~\ref{figcase-new-1}--\ref{figcase-new-9} compare spatial responses, recorded region
selections, and final answers for the same image--question
pairs.
The examples show how different observers distribute
their responses across the scene and which regions they
ultimately select.
Displaying the answers alongside these selections connects
where each observer looks with the resulting prediction. In each visualization, \textit{Rec.} identifies the
recorded region selection, while the cell values and
rank labels describe the displayed spatial responses.
The recorded selection and the response distribution
are shown separately to distinguish the region used
for observation construction from the spatial pattern
visualized over the scene.

\FloatBarrier

\subsection{Observation and Answering Examples}
\label{app:overvation-examples}

\begin{figure*}[htbp]
  \centerline{\includegraphics[width=\linewidth]{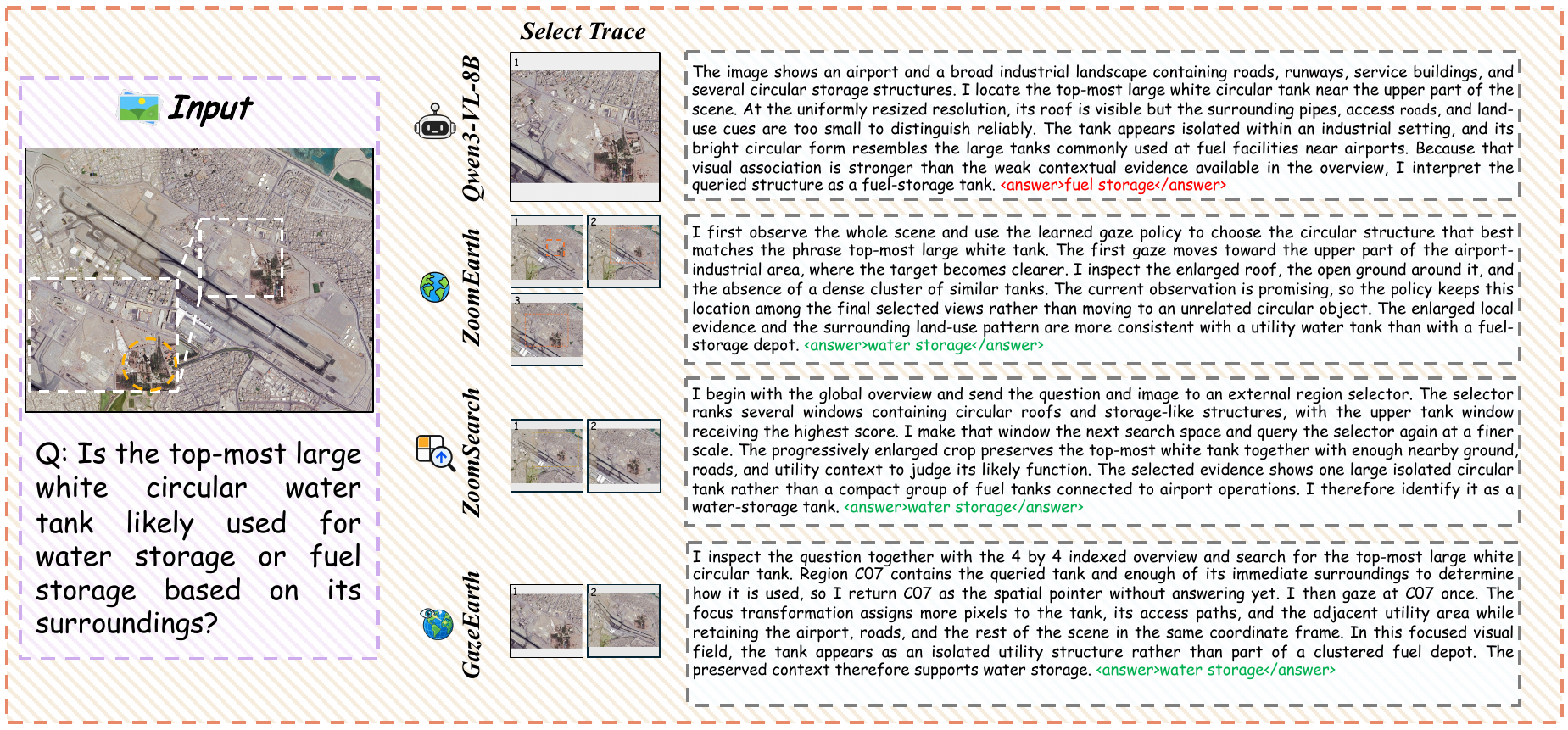}}
  \vspace{-0.01cm}
  \caption{Qualitative comparison of observation and answering
across methods (Part 1).
Each example presents the input image and question,
the visual observations used by the compared methods,
and their accompanying explanations and final answers.
GazeEarth enlarges the question-selected neighbourhood
within a continuous full-scene view, retaining the
surrounding spatial context for answering.}
  \vspace{-0.1cm}
  \label{figcase-1}
\end{figure*}

\begin{figure*}[htbp]
  \centerline{\includegraphics[width=\linewidth]{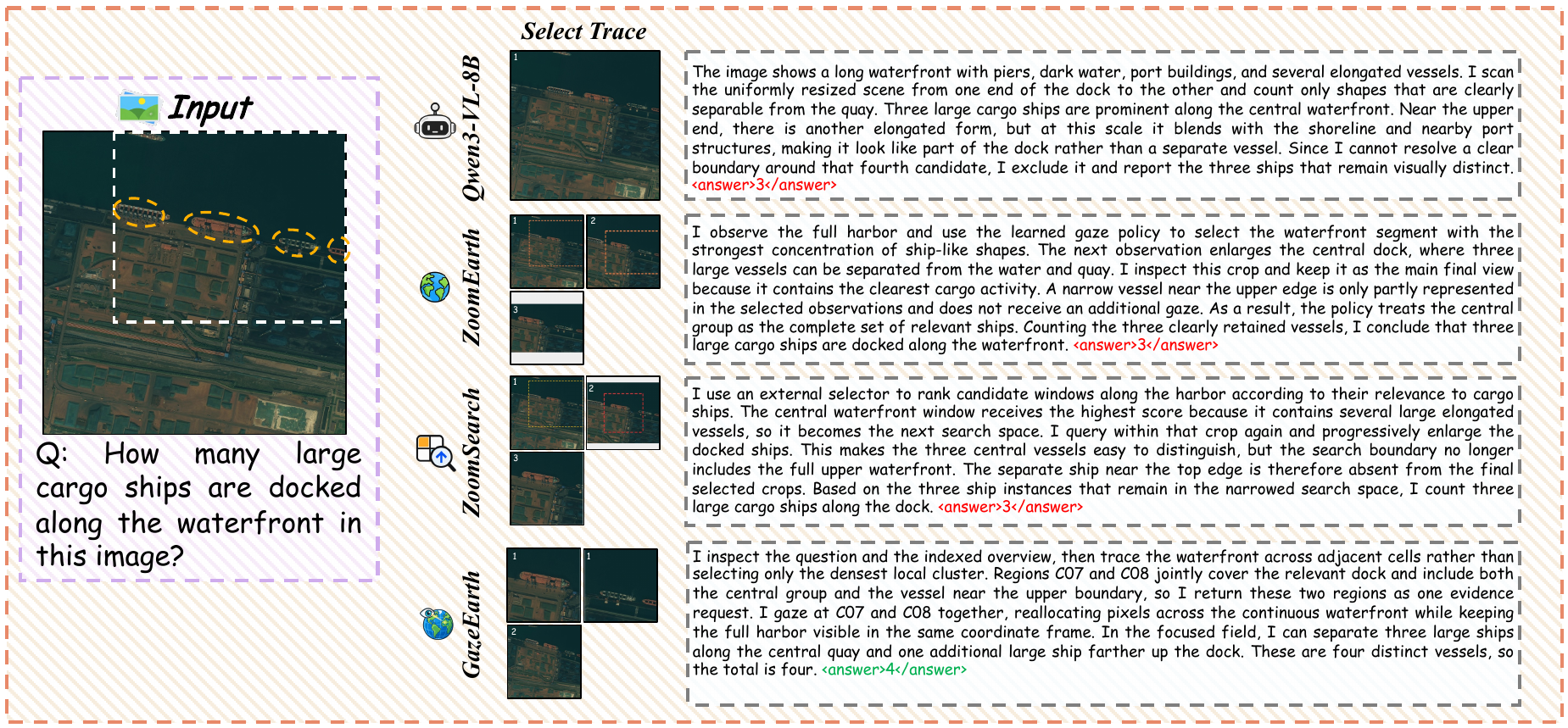}}
  \vspace{-0.01cm}
  \caption{Qualitative comparison of observation and answering
across methods (Part 2).
Each example presents the input image and question,
the visual observations used by the compared methods,
and their accompanying explanations and final answers.
GazeEarth enlarges the question-selected neighbourhood
within a continuous full-scene view, retaining the
surrounding spatial context for answering.}
  \vspace{-0.1cm}
  \label{figcase-2}
\end{figure*}

\begin{figure*}[htbp]
  \centerline{\includegraphics[width=\linewidth]{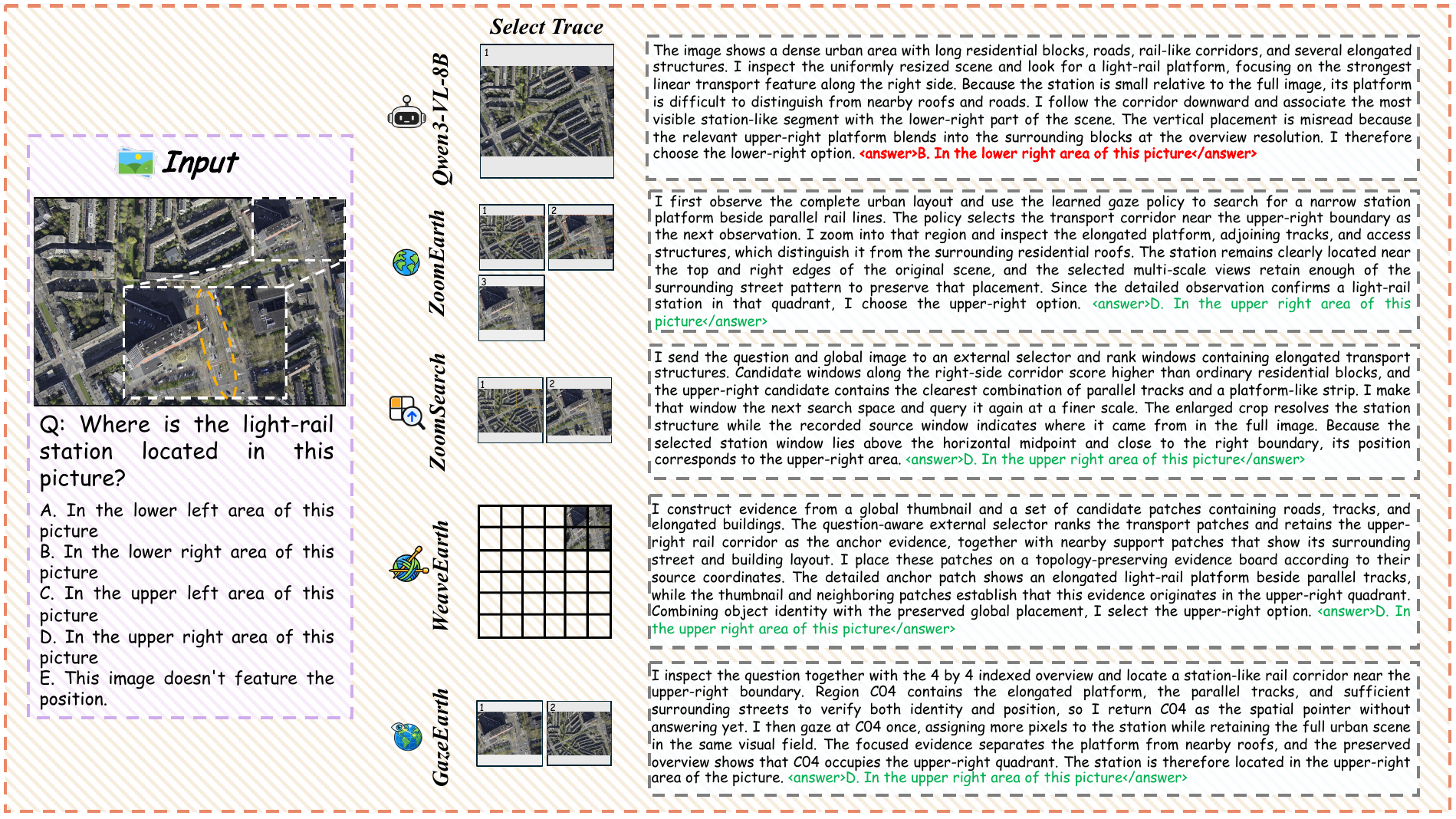}}
  \vspace{-0.01cm}
  \caption{Qualitative comparison of observation and answering
across methods (Part 3).
Each example presents the input image and question,
the visual observations used by the compared methods,
and their accompanying explanations and final answers.
GazeEarth enlarges the question-selected neighbourhood
within a continuous full-scene view, retaining the
surrounding spatial context for answering.}
  \vspace{-0.1cm}
  \label{figcase-3}
\end{figure*}

\begin{figure*}[htbp]
  \centerline{\includegraphics[width=\linewidth]{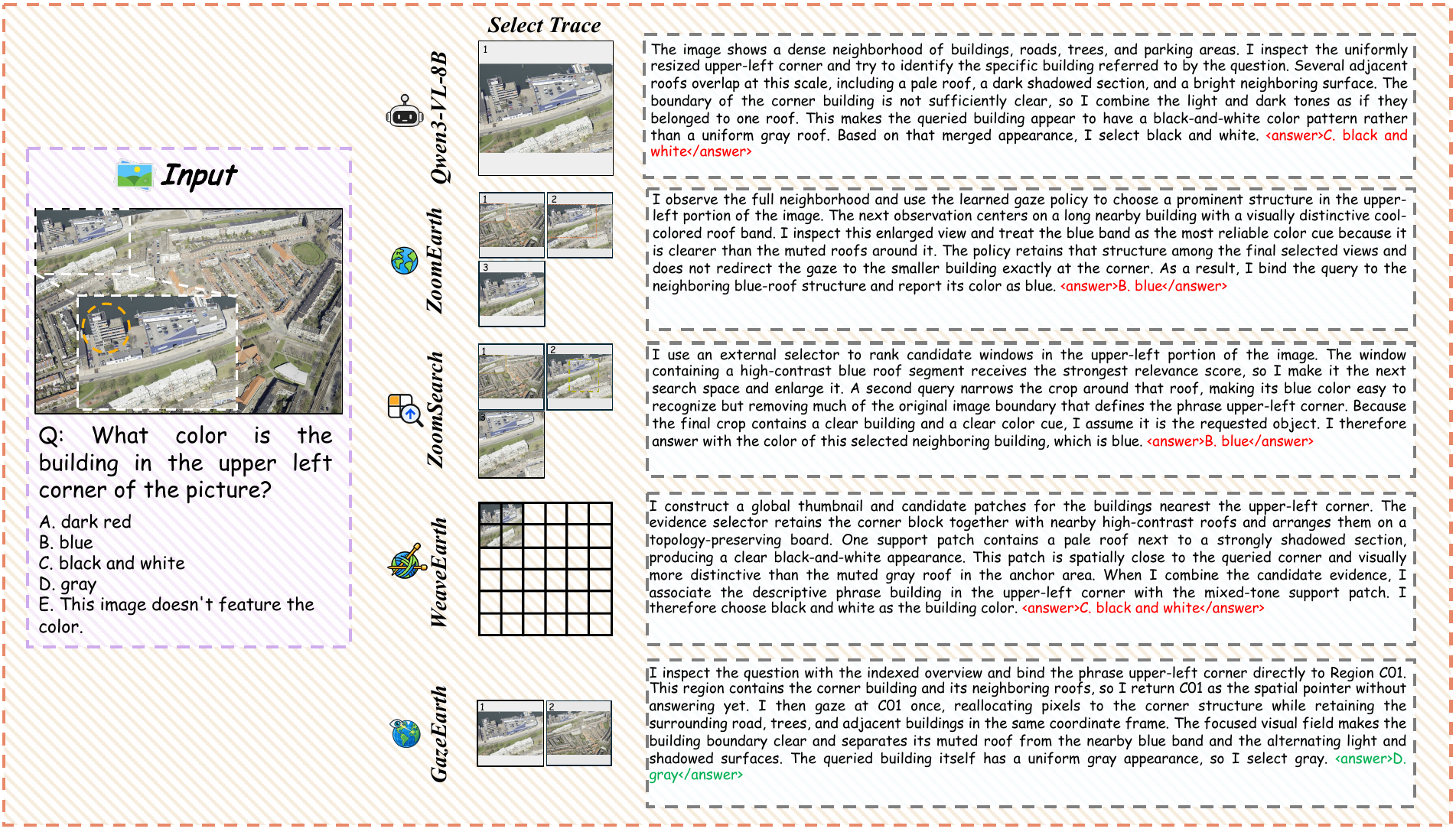}}
  \vspace{-0.01cm}
  \caption{Qualitative comparison of observation and answering
across methods (Part 4).
Each example presents the input image and question,
the visual observations used by the compared methods,
and their accompanying explanations and final answers.
GazeEarth enlarges the question-selected neighbourhood
within a continuous full-scene view, retaining the
surrounding spatial context for answering.}
  \vspace{-0.1cm}
  \label{figcase-4}
\end{figure*}

\begin{figure*}[htbp]
  \centerline{\includegraphics[width=\linewidth]{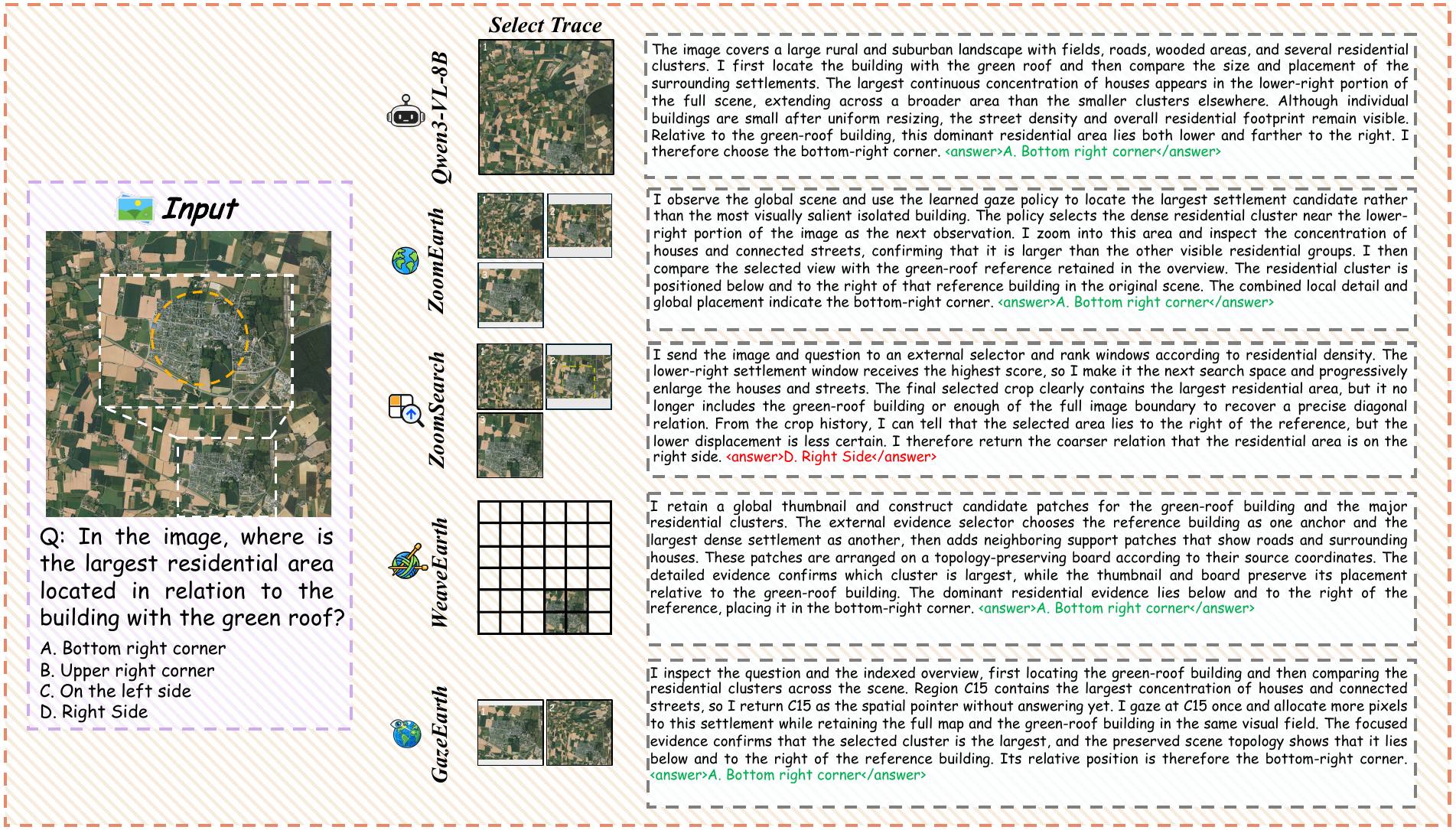}}
  \vspace{-0.01cm}
  \caption{Qualitative comparison of observation and answering
across methods (Part 5).
Each example presents the input image and question,
the visual observations used by the compared methods,
and their accompanying explanations and final answers.
GazeEarth enlarges the question-selected neighbourhood
within a continuous full-scene view, retaining the
surrounding spatial context for answering.}
  \vspace{-0.1cm}
  \label{figcase-5}
\end{figure*}

\begin{figure*}[htbp]
  \centerline{\includegraphics[width=\linewidth]{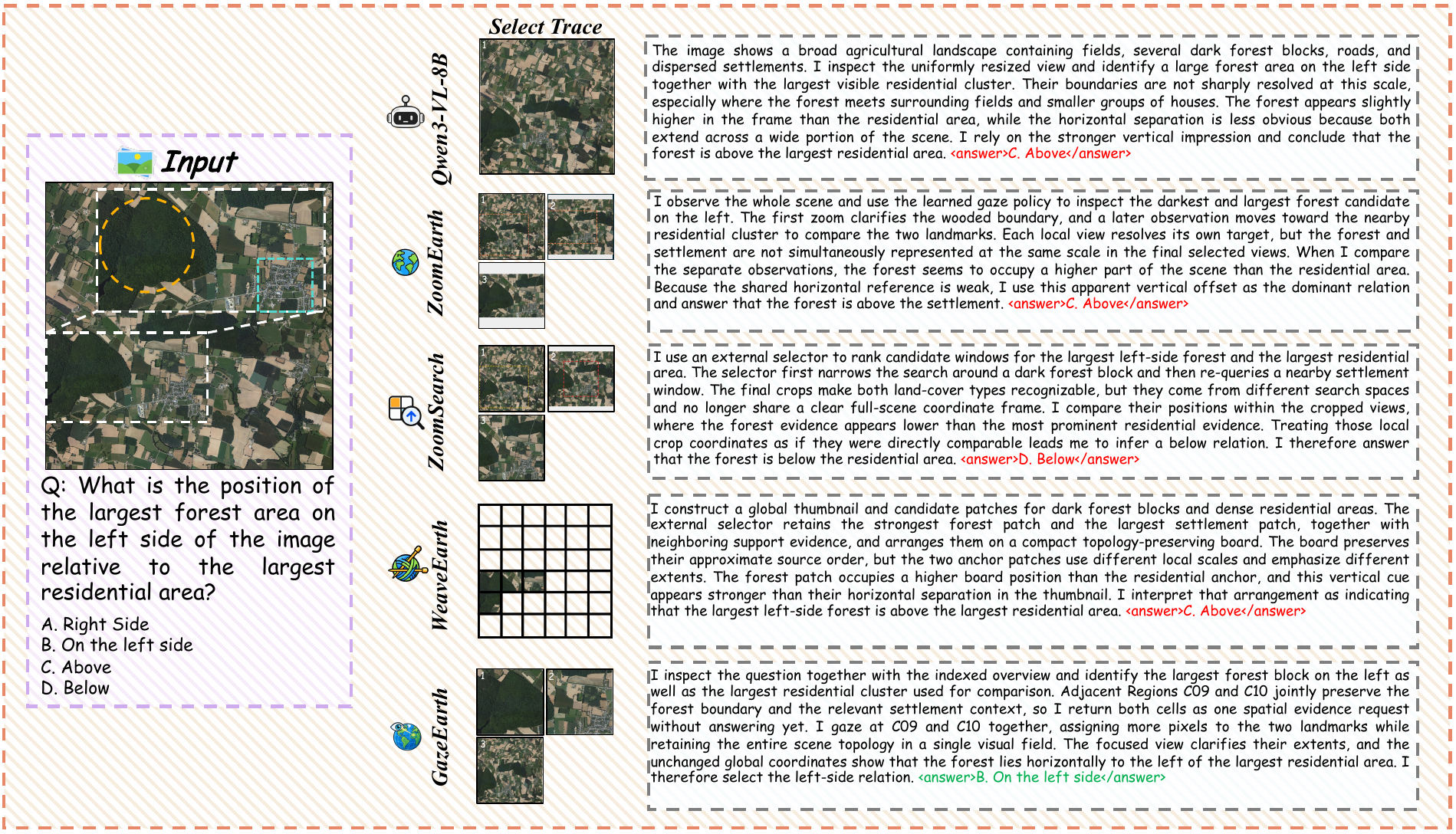}}
  \vspace{-0.01cm}
  \caption{Qualitative comparison of observation and answering
across methods (Part 6).
Each example presents the input image and question,
the visual observations used by the compared methods,
and their accompanying explanations and final answers.
GazeEarth enlarges the question-selected neighbourhood
within a continuous full-scene view, retaining the
surrounding spatial context for answering.}
  \vspace{-0.1cm}
  \label{figcase-6}
\end{figure*}

Figures~\ref{figcase-1}--\ref{figcase-6} compare the visual observations and answers
across methods on questions involving object attributes,
counting, and spatial relations.
GazeEarth enlarges the selected neighbourhood while
retaining surrounding landmarks within the same full-scene
view.
These examples illustrate how local evidence and its
spatial context are presented together for answering.
The accompanying explanations summarize the observation
and answering workflows rather than a single continuous
model response.
For GazeEarth, region selection produces a spatial request,
and answering is performed in a separate interaction
using the rendered view and the original question.
Spatial preservation refers to the ordering and
connectivity of scene regions under the warp;
their pixel coordinates and local scales change.
The examples include both shared successes across methods and differing answers to the same
question, illustrating how the
observation strategies behave across individual scenes.

\FloatBarrier
\section{Broader Impact}
\label{app:limitations}

UHR remote sensing supports disaster response, environmental
and agricultural monitoring, infrastructure inspection, and
urban planning, where analysts must relate small objects to
their surrounding scene.
GazeEarth improves such analysis without training or
fine-tuning: it reuses frozen, publicly available MLLMs and
adds only a deterministic resampling step.
This lowers the barrier for groups that cannot collect
task-specific annotations or train large models, such as
public agencies, non-governmental organizations, and
researchers in low-resource settings.
Because the answer-stage canvas is bounded, GazeEarth also
answers 1.45$\times$ faster than native-resolution Direct
inference in our Qwen3-VL-8B setting
(Appendix~\ref{app:efficiency}), reducing the compute and
energy cost per query.

Unlike methods that rely on internal attention or learned
zooming policies, GazeEarth exposes its spatial decision as
an explicit list of grid cells, and the rendered view that
the answerer receives can be inspected directly.
Users can therefore check where the model looked before
trusting an answer, which supports human oversight and
error auditing in applied settings.

Beyond remote sensing, our results point to observation construction as a design axis for multimodal models. Current MLLMs mostly receive a fixed, uniformly resized view, and progress on high-resolution inputs is usually sought through larger encoders, longer visual contexts, or additional training. Our findings suggest a complementary route: a frozen model can already express where it needs more detail, and how that request is turned into the next observation affects what the model can answer, sometimes as much as the choice of region itself.

\end{document}